\documentclass{article}

\PassOptionsToPackage{numbers, compress}{natbib}
\usepackage{times}
\usepackage{epsfig}
\usepackage{graphicx}
\usepackage{amsmath}
\usepackage{amssymb}
\usepackage{tabularx}
\usepackage{mathtools}
\usepackage{algorithm}
\usepackage{algpseudocode}
\usepackage{bm}
\usepackage{wrapfig}

\usepackage[dvipsnames]{xcolor}

\usepackage{graphicx}
\graphicspath{{figures/}{figures/qualitative/}}
\usepackage{booktabs}
\usepackage[lofdepth,lotdepth]{subfig}
\usepackage{appendix}
\usepackage{multirow}

\usepackage{makecell, tabularx}
\newcolumntype{Y}{>{\centering\arraybackslash}X}
\usepackage{rotating}
\usepackage[export]{adjustbox}
\usepackage{pgfplots}
\pgfplotsset{compat=1.18}

\usepackage[main, final]{neurips_2026}

\usepackage[utf8]{inputenc} %
\usepackage[T1]{fontenc}    %
\usepackage{hyperref}       %
\usepackage[capitalize]{cleveref} %
\usepackage{url}            %
\usepackage{booktabs}       %
\usepackage{amsfonts}       %
\usepackage{nicefrac}       %
\usepackage{microtype}      %
\usepackage{xcolor}         %

\renewcommand{\citet}[2][]{\cite[#1]{#2}}
\newcommand{\citenamed}[2]{#1~\cite{#2}}

\title{RigPAPR: Rig-Based Animation of Static Neural Point Clouds from a Fixed-Viewpoint Video}

\author{Shichong Peng$^{1}$, Yanshu Zhang$^{1}$, Ke Li$^{1,2,3}$ \\
$^{1}$APEX Lab, School of Computing Science, Simon Fraser University, Canada\\
$^{2}$Alberta Machine Intelligence Institute (Amii), Canada\\
$^{3}$Canadian Institute for Advanced Research (CIFAR), Canada\\
{\tt\small \{shichong\_peng,yanshu\_zhang,keli\}@sfu.ca}}

\begin{document}

\maketitle

\begin{abstract}
Static neural point reconstructions capture a subject at high fidelity from posed images. Given such a reconstruction, we aim to animate it to follow a monocular fixed-viewpoint driving video of the subject, whether captured or produced by image-to-video (I2V) generation, and to recover a rigged, re-posable 3D asset.
Existing methods deform Gaussian splats through direct linear blend skinning (LBS) or mesh proxies, both of which are prone to joint-boundary artifacts under articulation, even with per-primitive corrections.
We trace the artifact to the representation: each splat carries an individual shape calibrated in the canonical pose to tile with its neighbours. Under LBS each splat can be moved, rotated, or rescaled but never bent, so the canonical tiling comes apart at joint boundaries.
Proximity attention point rendering (PAPR) instead carries no per-primitive shape; each pixel is recomposed at render time from the deformed primitives' positions, so the surface re-forms naturally with the articulation.
We present RigPAPR, which auto-rigs a static PAPR cloud and drives it under direct LBS from a single fixed-viewpoint video, without a category template or a pose-dependent correction network.
RigPAPR is competitive with the strongest baseline at the supervised view and outperforms splatting-based and mesh-based baselines at novel views, by 3~dB PSNR on synthetic subjects and about 2~dB on real-world subjects, with cleaner joint-boundary renderings. Project~page: \mbox{\url{https://niopeng.github.io/RigPAPR/}}
\end{abstract}
    
\section{Introduction}
\label{sec:intro}

Triangle meshes are the default asset for 3D animation, yet recovering a high-fidelity mesh from real-world capture remains difficult, requiring both accurate surface reconstruction and view-consistent texture and material recovery. Neural rendering sidesteps this by reconstructing scenes at photoreal fidelity from posed images alone, beginning with NeRF's volumetric formulation~\citep{mildenhall2020nerf}. Explicit primitives have since become the more popular formulation, led by 3D Gaussian splatting~\citep{kerbl3Dgaussians} alongside other point-based alternatives~\citep{zhang2023papr, Lassner2021PulsarES, Rckert2021ADOPAD}. Because each learned primitive is editable after reconstruction, these representations become viable animation assets, not only novel-view renderers.

The most useful animation asset on top of such a static reconstruction is a skeletal rig: a sparse, human-interpretable, re-posable control interface that decades of skeletal-animation practice and automatic rigging are built around~\citep{magnenat1989joint, lewis2000pose, baran2007automatic}. Driving such a rig requires motion; image-to-video (I2V) generation~\citep{Chen2025KlingOmniTR, blattmann2023stable, brooks2024video} is a natural source, animating any subject from a single canonical frame. Existing rig-based work for neural point representations~\citep{uzolas2023template, wan2024template, animatablegaussians, qian20243dgs} requires multi-camera or orbiting-camera motion capture, while I2V most directly produces a single fixed-viewpoint video. We focus on this fixed-viewpoint setting: given such a monocular driving video and a static multi-view reconstruction of the subject, we aim to recover a rigged 3D asset whose motion can be re-rendered from novel viewpoints and whose skeleton can be re-posed to unseen configurations.

\begin{figure}[t]
  \vspace{-0.25em}
  \centering
  \includegraphics[width=\linewidth]{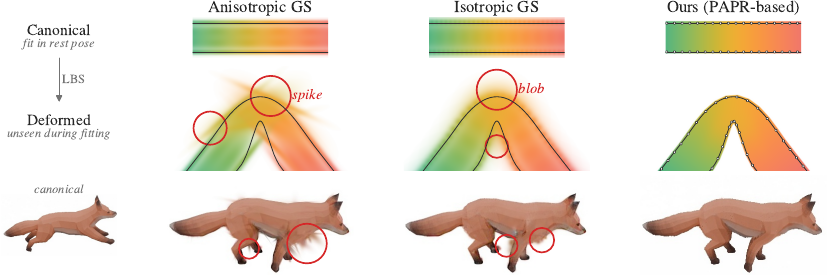}
  \caption{\textbf{Top:} a cross-section of a limb, fitted in the canonical pose and then bent at a
  joint with linear blend skinning (LBS).
  \textbf{Bottom:} anisotropic splats, isotropic splats and ours on the fox, driven by its
  ground-truth motion.
  Each Gaussian splat carries its own \emph{individual} shape, fitted in the canonical pose without
  knowing how the surface will later deform. LBS bends the surface non-rigidly: the outer side
  stretches, and the inner side compresses and thins. A splat, however, can only be moved,
  rotated, or rescaled, so its shape stays an ellipsoid and cannot bend with the surface. Long, thin anisotropic splats no
  longer lie along the bent surface and stick out as spikes. Making the splats isotropic removes the
  spikes, but the large round splats that fill the interior keep their size as the limb thins, and
  push through the surface as blobs. PAPR instead builds shape from a \emph{set} of shape-free
  points: each pixel is recomposed at render time from its deformed neighbours, so the surface
  follows the bend.}
  \label{fig:why-interpolation}
\end{figure}

Most existing approaches to rigging neural point-based representations adopt Gaussian-splat primitives, deformed either by direct linear blend skinning (LBS) on the splats~\citep{yao2025riggs, wan2024template} or by binding them to a deforming mesh or parametric template~\citep{manigs, gaussianavatars}. In our setting both routes produce visible artifacts near the joints, where the deformation pushes splats far from their canonical configuration. The cause is the primitive rather than the deformation: each splat carries its own \emph{individual} shape, an ellipsoid fixed by its covariance, which can be moved, rotated, or rescaled, but never bent. Neighbouring splats are fitted to tile into a seamless surface in a single static pose, with no observation of the motion to come, so when LBS later bends a joint the tiling comes apart. Anisotropic splats stand out from the bent surface as spikes; constraining them to be isotropic only rounds these into blobs (\cref{fig:why-interpolation}).

Directly addressing the cause calls for a different representation: an \emph{interpolation-based} representation, in which no primitive carries a fixed shape parameter. Instead, the shape emerges from the spatial configuration of a \emph{set} of primitives, not from any single one. Under deformation, the positions of the nearby primitives change, and so the rendered pixels change accordingly, which allows the rendering to adapt to the deformed spatial configuration automatically. \cref{fig:why-interpolation} (bottom right) shows the consequence: the same deformation that broke the Gaussian tiling can be rendered as a seamless surface with interpolation. Proximity attention point rendering (PAPR)~\citep{zhang2023papr} realizes this representation: each primitive carries only a position and a view-independent feature, and each pixel is rendered as an attention-weighted aggregation of its top-$K$ nearest primitives, with weights computed at render time from their current positions. Under direct LBS, the attention weights update automatically with the deformed positions.

Building on this insight, we present \emph{RigPAPR}, a method that animates a static PAPR cloud from a single fixed-viewpoint driving video, captured or I2V-generated, and produces a re-posable rigged 3D asset without category templates. Given a pre-trained PAPR cloud and the driving video, the method first auto-rigs the point cloud through a transient mesh proxy, which is discarded once binding is set. The point cloud is then deformed directly under LBS, with joint rotations and skinning weights refined through two-phase optimization driven by the video. We evaluate on synthetic and real-world subjects against strong representative splatting-based and mesh-based baselines. At the supervised view RigPAPR is competitive with the strongest baseline. At held-out novel views it outperforms all baselines, by 3~dB in PSNR on synthetic subjects and about 2~dB on real-world subjects, and renders joint boundaries more cleanly.

\section{Related Work} \label{sec:related_work}

\paragraph{3D representations for animatable reconstruction.}
The choice of 3D representation determines how the reconstructed object's geometry is stored before animation. Implicit representations~\citep{mildenhall2020nerf,wang2021neus,li2023neuralangelo} describe geometry as a continuous function of space, whereas explicit representations such as meshes and point clouds store geometry as discrete primitives with positions and, in some cases, local shape parameters. For animation, explicit representations are convenient because their primitives can be transformed directly to change the geometry. Among explicit neural representations, point-based neural renderers reconstruct real-world captures from posed images at high fidelity, without the surface-reconstruction step that mesh-based pipelines require. 3D Gaussian splatting~\citep{kerbl3Dgaussians} (3DGS) is the dominant such representation. An axis that matters under deformation is how shape is encoded: 3DGS encodes shape \emph{per primitive} via an anisotropic 3D covariance (a local ellipsoidal kernel), whereas PAPR~\citep{zhang2023papr} carries no per-primitive shape parameter: shape emerges at render time from proximity attention over a neighbourhood of primitives.

\paragraph{Rig-based animation and its precedent.} Captured subjects can be made animatable along a spectrum of control interfaces. Per-frame deformation fields~\citep{park2021nerfies,pumarola2021d,yang2024deformable,wu20244d} reconstruct the observed motion as a continuous 4D field, but they replay rather than re-pose: the asset exposes no re-pose interface beyond the captured latent trajectory. Cage and control-point methods~\citep{NeRFshop23,huang2024sc} provide a re-posable asset, but steer it through a dense set of surface handles that remains cumbersome to manipulate directly. Rig-based methods expose a skeleton instead: a sparse, human-interpretable interface that the existing motion-capture data, retargeting tools, and authoring workflows of production pipelines are already built around. We target this setting.

Rig-based animation has been studied extensively for meshes. Automatic rig prediction~\citep{baran2007automatic,xu2020rignet,song2025puppeteer,zhang2025one,liu2025riganything,xu2022morig} makes rigging tractable without manual authoring, producing a skeleton with skinning weights bound to a mesh. Recent generative work extends this to motion: ActionMesh~\citep{Sabathier2026ActionMeshA3} generates an animated mesh from a video, and RigMo~\citep{Zhang2026RigMoUR} recovers a rig and its motion from such a mesh sequence. Their combination produces a rigged mesh but recovers no view-consistent appearance of the captured subject from its posed images. Skeleton-driven deformation on non-mesh primitives has precedent: meshless and point-sampled animation driven by skeletons dates to the mid-2000s~\citep{guo2005realtime,song2006meshless,rivers2007fastlsm}, with modern differentiable descendants~\citep{zhang2025physrig}. These methods drive point sets or lattices through physics-based deformation (meshless elasticity, shape matching) rather than linear blend skinning, and none target photorealistic rendering of captured scenes.

\paragraph{Fixed-viewpoint rig-based animation.} Rig-based work for neural representations has explored multi-view supervision during the driving sequence, whether from multiple synchronised cameras or from a deliberately moving camera that encircles the subject across timesteps~\citep{uzolas2023template,svgs2025,wan2024template}, coverage that disambiguates depth from articulation. We focus on the setting where the driving sequence is a single fixed-viewpoint video, the form I2V models~\citep{Chen2025KlingOmniTR, blattmann2023stable, brooks2024video} produce most directly. In this regime, depth and articulation explain the same 2D residual at every frame, and the problem is under-constrained. Methods that operate here constrain the optimization with auxiliary information in three ways: (a)~a strong generative prior based on an image or video generation model~\citep{sc4d2024,gu2025das}; (b)~a pre-existing category template such as SMPL~\citep{hugs,gaussianavatars,animatablegaussians}; (c)~monocular pre-trained priors that provide approximate per-frame guidance, such as off-the-shelf depth estimators~\citep{depthanything3} and 2D point trackers~\citep{karaev24cotracker3}. Our work and AnimaMimic~\citep{animamimic2026} both follow route (c). We differ on the \emph{representation axis}: AnimaMimic's pipeline runs on a mesh that must be supplied upstream, whereas our point-based neural representation is reconstructed directly from posed real images.

\paragraph{Rig-based animation of point-based neural representations.} Recent work has begun to rig point-based neural representations. Most efforts operate on Gaussian-splat primitives; one line operates on density-field neural points under volumetric rendering.

\emph{Direct LBS on Gaussian splats.} LBS is applied directly to the splats with learned per-primitive skinning weights~\citep{wan2024template,camo2026}. A parallel line~\citep{yao2025riggs,svgs2025} retains the direct-LBS move but adds a pose-dependent detail-deformation head that displaces splat positions at each timestep to compensate for rendering artifacts under articulation. Because the head is trained on the observed pose distribution, this pose-dependent compensation may overfit to those poses and generalize weakly to novel ones.

\emph{Mesh-proxy binding.} Splats are bound~\citep{guedon2023sugar,games,manigs,a3gs2026,li2024dreammesh4d} to a triangle mesh that provides a deforming surface reference: as the mesh deforms, splats follow it in position, and their covariances reorient with the local surface frame. A specialized sub-variant uses parametric templates such as SMPL~\citep{SMPL:2015} or FLAME~\citep{Li2017LearningAM} as the mesh proxy~\citep{hugs,animatablegaussians,gaussianavatars,qian20243dgs}, avoiding topology estimation but restricting the method to categories with a pre-existing template. Several of these works also adopt dual-quaternion skinning~\citep{kavan2007skinning,kavan2008dqs} or local rigidity regularizers such as ARAP~\citep{SorkineHornung2007AsrigidaspossibleSM} to soften the non-rigid blending of bone transforms at joint boundaries. These fixes adjust the deformed position, but the splat covariance remains calibrated to the canonical-pose surface, so the artifacts are softened but persist.

\emph{Cage or sparse control graph.} A third family operates on the same Gaussian-splat primitive but replaces the skeleton with a dense set of surface handles, so re-posing becomes geometric editing of the cage rather than skeletal manipulation~\citep{sc4d2024}. These methods sit adjacent to rig-based work and share the primitive-level limitation analyzed below. Recent variants approach the skeleton from the other side: a kinematic-tree control graph~\citep{mbgs2025}, or piecewise-rigid Gaussian clusters~\citep{deformsplat2025}. Both still expose a dense-handle interface rather than a sparse rig.

All the Gaussian-splat coping patterns above share a structural problem rooted in the primitive rather than the method. Each splat carries an individual shape, an ellipsoid fixed by its covariance, calibrated in the canonical pose to tile with its neighbours. A bone transform moves, rotates and rescales that ellipsoid, and recomputing its covariance from the local deformation only maps it to another ellipsoid. No covariance update bends the splat, so it cannot follow a surface that bends between two bones, and the canonical tiling comes apart at joint boundaries. The mesh-proxy, cage, and correction MLP mechanisms each target a symptom of this mismatch while leaving shape inside the primitive, where the artifact originates. We therefore build on PAPR, an interpolation-based representation whose primitives carry no such parameter: each pixel's rendered output is recomposed at render time from the current positions of its top-$K$ nearest primitives, so the rendered surface follows the deformed configuration.

\emph{Density-field neural points.} \citenamed{Uzolas et al.}{uzolas2023template} rig a Point-NeRF~\citep{xu2022point}-style density-field point cloud directly with LBS, with no pose-dependent correction MLP. Their points carry no explicit per-primitive shape, yet rendering aggregates per-primitive contributions independently rather than interpolating across primitives at the pixel. They further differ on supervision and on point distribution. They require multi-view supervision (a varying viewpoint across the sequence) to resolve the depth-articulation ambiguity, whereas we operate from a single fixed-viewpoint video. Their points also volumetrically sample the object interior for density-field rendering, which precludes feeding them to surface-based auto-riggers~\citep{song2025puppeteer,xu2020rignet,zhang2025one} that PAPR's surface-concentrated points support directly.

\section{Method} \label{sec:method}

\subsection{Overview} \label{sec:method-overview}

From a static PAPR~\citep{zhang2023papr} cloud and a fixed-viewpoint driving video, our method produces a rigged point-based asset that reproduces the driving motion, renders faithfully from unseen viewpoints, and is re-posable to unseen configurations. We write the PAPR scene as $N$ points $\{(\mathbf{p}_i,\mathbf{u}_i,\tau_i)\}_{i=1}^N$ with positions $\mathbf{p}_i\in\mathbb{R}^3$, view-independent features $\mathbf{u}_i\in\mathbb{R}^d$, and influence scores $\tau_i\in\mathbb{R}$; the skeleton has $B$ bones and each point carries a per-bone skinning weight $\mathbf{w}_i\in\mathbb{R}^B$. A pose at frame $t$ comprises one axis-angle rotation per bone, $\{\mathbf{R}_b^t\}_{b=1}^B\subset SO(3)$, plus a global root translation $\mathbf{t}^t\in\mathbb{R}^3$; forward kinematics composes these with the fixed bind-pose bone offsets to yield per-bone $SE(3)$ transforms $\mathbf{T}_b(\mathbf{R}^t)$. The driving video supplies $T$ frames $\{\mathbf{I}^t\}_{t=1}^T$ from a single camera $\mathcal{C}$.

\begin{figure}[t]
  \centering
  \includegraphics[width=\linewidth]{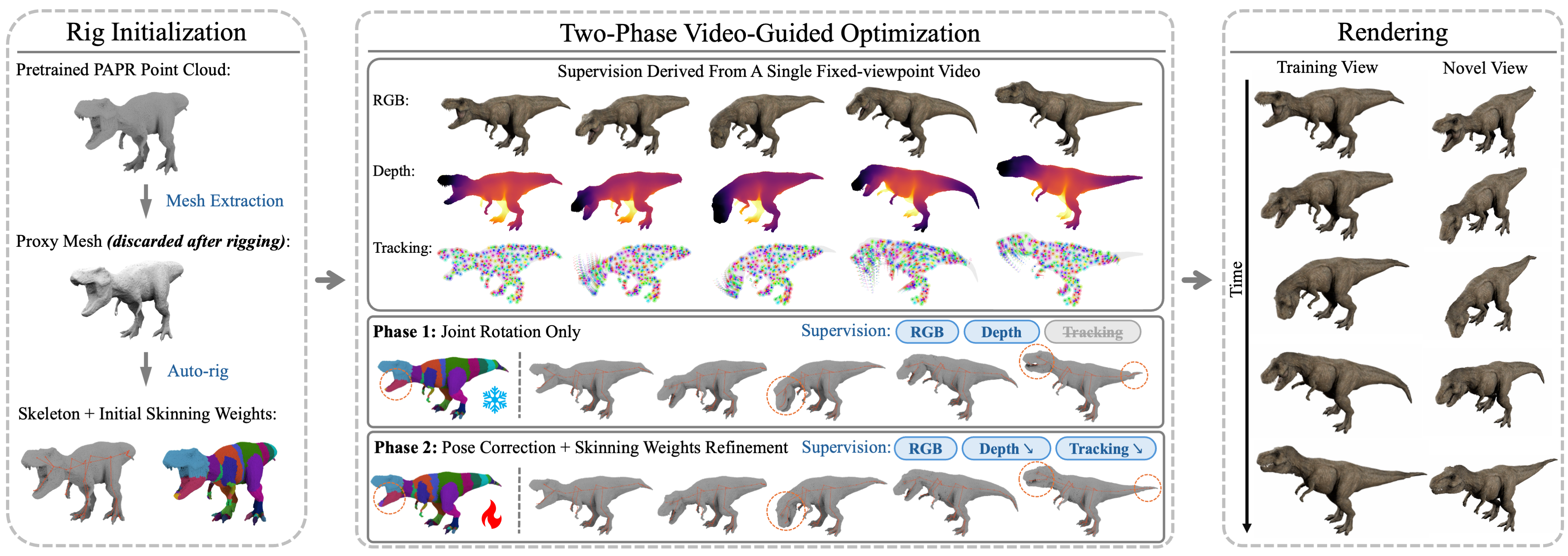}
  \caption{\textbf{RigPAPR overview.} A mesh proxy extracted from the PAPR cloud is auto-rigged and the resulting skeleton and initial skinning weights are transferred back onto the points; the proxy is then discarded (\S\ref{sec:lbs-on-papr}). A two-phase optimization drives the rigged cloud from the fixed-viewpoint video, recovering per-frame joint rotations and refining the skinning weights (\S\ref{sec:multi-phase}). At inference, LBS deforms the canonical points and the frozen PAPR renderer renders the asset from different viewpoints.}
  \label{fig:pipeline}
\end{figure}

Figure~\ref{fig:pipeline} shows the overview of our method. \S\ref{sec:papr} recalls the PAPR properties we exploit; \S\ref{sec:lbs-on-papr} attaches an articulated rig directly to the PAPR cloud, with no mesh proxy at deformation or render time; \S\ref{sec:multi-phase} drives the rig from the video through a two-phase optimization that resolves the depth-articulation ambiguity inherent to a single fixed view.

\subsection{Preliminaries: Proximity Attention Point Rendering} \label{sec:papr}

PAPR represents a scene as $N$ points whose positions and view-independent features are stored as learnable parameters and trained end-to-end against the input images. To render a pixel, PAPR casts a camera ray, selects the top-$K$ points nearest the ray, and aggregates them with a proximity-attention head: the head produces softmax weights over the top-$K$ points and aggregates their per-point contributions into the rendered pixel RGB~$\hat{\mathbf{I}}$. The attention weights and per-point contributions are computed from each point's relation to the ray (its position, its depth along the ray, and its perpendicular offset) together with the ray direction; we refer to~\citet{zhang2023papr} for the full attention design.

Two properties of PAPR matter for what follows. First, the same indexed point set persists across reconstruction and deformation, and the renderer relies on no surface connectivity (such as mesh edges) that LBS could break. Second, because the attention weights renormalize over the current top-$K$ neighbourhood at render time, every pixel still receives a non-vanishing gradient when points move far from the binding configuration; we exploit this property when fine-tuning the representation per frame. Backbone configuration and the variant we use are detailed in the appendix.

\subsection{Direct Linear Blend Skinning on Neural Points} \label{sec:lbs-on-papr}

We attach an articulated rig to the PAPR scene by treating each point as a free linear blend skinning (LBS) handle: the same point set used for static reconstruction is the point set we deform per frame, with no mesh proxy or resampling at deformation or render time and no retraining of the renderer.

\paragraph{Skeleton and weight initialization.} We mesh the trained PAPR cloud with OffsetOPT~\citep{offsetopt} and feed the proxy to Puppeteer~\citep{song2025puppeteer}, which returns a $B$-bone skeleton and per-vertex skinning weights; we transfer the weights to the PAPR points to obtain $\mathbf{w}_i\in\mathbb{R}^B$. The proxy is used only here, to seed the skeleton and skinning weights, and is discarded afterwards; it participates in neither deformation nor rendering. \S\ref{sec:multi-phase} then refines $\mathbf{w}$ and recovers per-frame rotations against the driving video.

\paragraph{LBS on PAPR points.}
Let $\{\mathbf{R}_b^t\}_{b=1}^B$ be the bone rotations at frame $t$
and $\mathbf{T}_b(\mathbf{R}^t)\in SE(3)$ the corresponding skinning
transform, defined as the bone-to-world transform at frame $t$
composed with the world-to-bone transform at the binding pose, so
that $\mathbf{T}_b(\mathbf{R}^0)=\mathbf{I}$ for every bone and the
canonical configuration is the fixed point of
Equation~\eqref{eq:lbs}. With positions written in homogeneous
coordinates, the deformed position of point $i$ at frame $t$ is
\begin{equation}
    \mathbf{p}_i^t \;=\; \mathbf{t}^t \;+\; \sum_{b=1}^B w_{i,b}\,
        \mathbf{T}_b(\mathbf{R}^t)\,\mathbf{p}_i^0,
    \label{eq:lbs}
\end{equation}
where $\mathbf{p}_i^0$ is the canonical position and $\mathbf{t}^t$
the global root translation. Only positions are deformed: $\mathbf{u}_i$ and $\tau_i$
travel rigidly with the point, and the proximity-attention renderer
re-aggregates the deformed neighbourhood at render time. Each PAPR point is therefore a
persistent, appearance-stable handle: articulation moves the handle
while leaving its appearance content untouched.

A useful side benefit follows directly: with no edges to preserve, the representation does not freeze connectivity at the binding frame. Deformations that require connectivity changes, such as an arm separating from a torso it touches at binding, remain expressible, whereas mesh-bound schemes propagate the binding-frame topology forward indefinitely through their splat-to-triangle pinning.

\subsection{Video-Guided Two-Phase Optimization} \label{sec:multi-phase}

The single-view setting is under-constrained: depth and articulation explain the same 2D residual at every frame, so a rendering loss alone cannot distinguish the correct configuration from one that renders identically at $\mathcal{C}$ but differs from every other viewpoint. Monocular depth and 2D tracks resolve the ambiguity but inject their own bias and noise; we therefore use them as decaying scaffolding: present early to break the ambiguity, then faded out so the converged optimum is not pinned by their failure modes. Loss weights and schedules are tabulated in the appendix.

\paragraph{Phase 1: Depth-supervised joint-rotation initialization.} Phase~1 freezes the skinning weights and optimizes the per-frame joint rotations $\{\mathbf{R}_b^t\}_{t=1}^T$ together with per-frame global root translations $\{\mathbf{t}^t\}_{t=1}^T\subset\mathbb{R}^3$, fit sequentially from the binding frame with each frame warm-started from the previous. The objective combines a rendering loss $\mathcal{L}_{\text{rgb}}$ on the rendered patch with a relative-depth term $\mathcal{L}_{\Delta\text{depth}}$ and a descendant-weighted rotation regularizer that keeps joint rotations small in proportion to how many child joints they propagate to. Monocular predictors such as Depth Anything~3~\citep{depthanything3} do not produce depths in PAPR units. We use the metric-depth variant, whose frame-to-frame \emph{relative} signal at the same canonical point tracks true depth changes reasonably well up to a per-frame scalar. We compute that scalar in closed form on each step's sample (held constant in the backward pass) and penalize the calibrated residual in MSE; full construction in the appendix. We do not use 2D tracks at this phase: relative depth alone breaks the front-versus-back ambiguity, and tracking would overconstrain the rotations before they have settled.

\paragraph{Phase 2: Pose corrections and skinning-weight refinement.} On top of the Phase~1 trajectory, Phase~2 unfreezes two additional parameter sets: per-joint axis-angle \emph{corrections} produced by a small temporal MLP, and the skinning-weight logits that yield $\mathbf{w}$. The Phase~1 rotations $\{\mathbf{R}_b^t\}$ and the Phase~1 translations $\{\mathbf{t}^t\}$ remain frozen; each correction is added to the Phase~1 rotation of its joint and propagates to points only through LBS with the refined weights. Unlike the per-primitive correction MLPs of Gaussian-splat-based methods (\S\ref{sec:related_work}), these corrections act at the joint level, on bone rotations rather than primitive parameters.

The Phase~2 objective extends Phase~1 with a CoTracker~3~\citep{karaev24cotracker3} tracking loss $\mathcal{L}_{\text{track}}$, an as-rigid-as-possible distance-preservation term over each point's $k$-nearest canonical neighbours that anchors local geometry through deformation, and standard smoothness and magnitude regularizers on the corrections and weights. CoTracker queries are seeded from PAPR's per-pixel attention dominance at the canonical view rather than raw projection, so each track corresponds to a point the renderer actually relies on (details in appendix). Both $\lambda_{\Delta\text{depth}}$ and $\lambda_{\text{track}}$ decay to zero over Phase~2, leaving a pure rendering-driven refinement on top of the joint trajectories that depth and tracking established.

\paragraph{Rendering at novel views.} At inference, novel views require a single forward pass: LBS-deform the canonical points with the optimized $(\mathbf{w}, \{\mathbf{R}^t\}, \{\mathbf{t}^t\})$, then render with the frozen PAPR network. The same canonical asset re-poses to arbitrary novel rotations and renders from arbitrary novel views without per-frame or per-view fitting. The one-time fit takes about an hour per phase on a single RTX~3090 for a 16-frame sequence.

\section{Experiments} \label{sec:experiment}

\subsection{Experimental Setup} \label{sec:setup}

\paragraph{Baselines.} We compare against baselines that represent the deformation models our analysis contrasts with. Mani-GS~\citep{manigs} deforms Gaussian splats through a mesh: each splat is bound to a triangle of a mesh proxy that is deformed by the rig. RigGS~\citep{yao2025riggs} instead applies linear blend skinning directly to the splats, with a pose-conditioned network that corrects the skinned result. Puppeteer~\citep{song2025puppeteer} serves as a mesh-based animation baseline. We adapt both splatting-based baselines to our setting, giving them the same canonical reconstruction, auto-rig, and two-stage optimization as our method; implementation details are in appendix~\ref{sec:supp-baselines}.

\paragraph{Datasets.} For synthetic evaluation we pick five objects from Objaverse~\citep{objaverse} spanning three morphologies: bipeds (T-Rex, Simpsons), quadrupeds (fox, wolf), and an octopod (spider). For each scene we render a fixed-viewpoint $512{\times}512$ video as the training view. For the splatting-based baselines and our method we additionally pre-train on the first frame using multi-view reconstruction, rendered from a ring of cameras around the subject; Puppeteer takes the ground-truth mesh and texture directly. For real-world evaluation we use a short clip of each of two subjects (313 and 390) from ZJU-MoCap~\citep{Peng2020NeuralBI}, a synchronized multi-camera capture of human motion: we pre-train on the first frame of the multi-view capture, use one camera as the fixed-viewpoint driving video, and hold out the remaining cameras as ground truth for novel views. Across both settings, sequences are 16--40 frames; per-scene lengths are tabulated in the appendix. Finally, as a demonstration of the image-to-video use case that motivates the fixed-viewpoint setting, we capture two static scenes of our own (robot, statue) and animate them from a Kling~3.0~\citep{Chen2025KlingOmniTR} video generated at one canonical view; we show qualitative results only (Figure~\ref{fig:captured-main}).

\begin{table}[t]
\centering
\caption{Quantitative comparison on synthetic sequences (5-scene mean) and on
ZJU-MoCap (2-scene mean). We report PSNR, SSIM and LPIPS at
the training view and at four held-out novel views. Puppeteer~\citep{song2025puppeteer}
is evaluated on synthetic data only. Baselines are Mani-GS~\citep{manigs} and
RigGS~\citep{yao2025riggs}. \textbf{Best} and \underline{second best} per column;
values that print identically are underlined on both sides.}
\label{tab:quantitative}
\scriptsize
\setlength{\tabcolsep}{2pt}
\begin{tabular}{l|ccc|ccc|ccc|ccc}
\toprule
 & \multicolumn{6}{c|}{Synthetic} & \multicolumn{6}{c}{ZJU-MoCap} \\
\cmidrule(lr){2-7}\cmidrule(lr){8-13}
 & \multicolumn{3}{c|}{Train view} & \multicolumn{3}{c|}{Novel view}
 & \multicolumn{3}{c|}{Train view} & \multicolumn{3}{c}{Novel view} \\
Method & PSNR$\uparrow$ & SSIM$\uparrow$ & LPIPS$\downarrow$
       & PSNR$\uparrow$ & SSIM$\uparrow$ & LPIPS$\downarrow$
       & PSNR$\uparrow$ & SSIM$\uparrow$ & LPIPS$\downarrow$
       & PSNR$\uparrow$ & SSIM$\uparrow$ & LPIPS$\downarrow$ \\
\midrule
Puppeteer & 22.48 & 0.941 & 0.044 & 19.27 & 0.923 & 0.075
          & --- & --- & --- & --- & --- & --- \\
Mani-GS & 28.53 & \underline{0.974} & \textbf{0.026}
        & \underline{20.51} & \underline{0.927} & 0.072
        & 26.38 & \textbf{0.974} & \textbf{0.022}
        & \underline{21.33} & \underline{0.937} & 0.065 \\
RigGS & \textbf{29.52} & \textbf{0.975} & 0.029
      & 20.37 & 0.926 & \underline{0.069}
      & \underline{27.06} & 0.969 & 0.037
      & 21.21 & \underline{0.937} & \underline{0.057} \\
Ours & \underline{29.04} & 0.971 & \underline{0.027}
     & \textbf{23.51} & \textbf{0.944} & \textbf{0.045}
     & \textbf{27.67} & \underline{0.972} & \underline{0.034}
     & \textbf{23.20} & \textbf{0.952} & \textbf{0.044} \\
\bottomrule
\end{tabular}
\end{table}

\paragraph{Metrics.} We report PSNR, SSIM~\citep{wang2004ssim}, and LPIPS~\citep{zhang2018lpips} (VGG backbone) at the training view and at four held-out cameras. On synthetic data the held-out cameras are placed uniformly on a ring around the subject; on ZJU-MoCap they are the capture camera nearest the driving camera and those at roughly $90^\circ$, $180^\circ$ and $270^\circ$ from it, with the subject's foreground composited on a white background. All metrics are averaged over the sequence with the canonical frame excluded.

\subsection{Quantitative Results} \label{sec:quantitative}

Table~\ref{tab:quantitative} reports the quantitative comparison. At the training view, our method is competitive with both splatting-based baselines on synthetic scenes and on ZJU-MoCap, and outperforms Puppeteer on every metric. At held-out novel views, our method outperforms all baselines on every metric on both datasets, by 3~dB PSNR on synthetic scenes and about 2~dB on ZJU-MoCap. Although every method loses quality away from the supervised view, ours best preserves novel-view quality.

\subsection{Qualitative Results} \label{sec:qualitative}

Video renderings for all results below are provided on the \href{https://niopeng.github.io/RigPAPR/}{project website}; we encourage the reader to consult them, as articulation artifacts are most evident in motion.

\paragraph{Synthetic qualitative.} Figure~\ref{fig:qualitative-synthetic} shows the qualitative comparison on synthetic T-Rex; the remaining four subjects (fox, Simpsons, wolf, spider) are deferred to the appendix (Figures~\ref{fig:qualitative-synthetic-fox-supp}--\ref{fig:qualitative-synthetic-spider-supp}). Mani-GS and RigGS show visible artifacts under articulation (highlighted in red), most pronounced around the head and belly in the novel view. Puppeteer fails to fit the correct geometry, exemplified by the head not pointing in the right orientation. Our method avoids both failure modes.

\def\synimgw{0.1525\linewidth}
\def\synimgh{0.104\linewidth}
\def\synlabelw{0.07\linewidth}
\def\synblockgap{4pt}
\providecommand{\synlabel}[1]{\parbox[b][\synimgh][c]{\synlabelw}{\centering\scriptsize #1}}

\begin{figure}[t]
\vspace{-1em}
\centering
\footnotesize
\setlength{\tabcolsep}{0pt}
\renewcommand{\arraystretch}{0.4}
\begin{tabular}{@{}c@{\hspace{1pt}}ccc@{\hspace{\synblockgap}}ccc@{}}
& \multicolumn{3}{c}{\textbf{Training View}} & \multicolumn{3}{c}{\textbf{Novel View}} \\
\cmidrule(lr){2-4}\cmidrule(lr){5-7}
& {\scriptsize $t_0$ (canonical)} & {\scriptsize $t_1$} & {\scriptsize $t_2$}
& {\scriptsize $t_0$ (canonical)} & {\scriptsize $t_1$} & {\scriptsize $t_2$} \\
\synlabel{Puppeteer} &
\includegraphics[width=\synimgw]{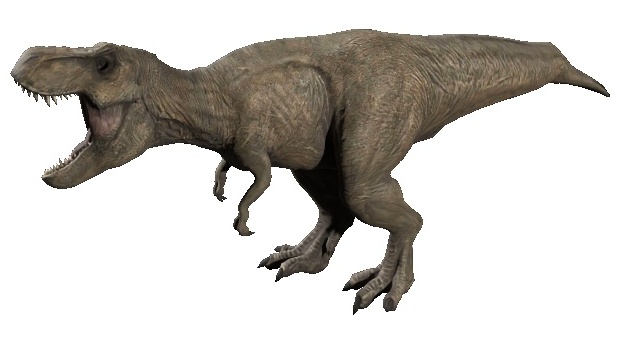} &
\includegraphics[width=\synimgw]{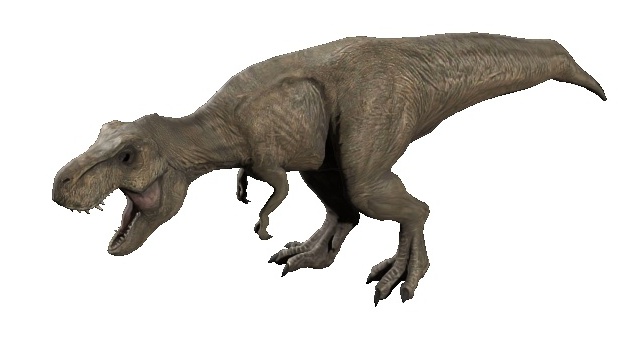} &
\includegraphics[width=\synimgw]{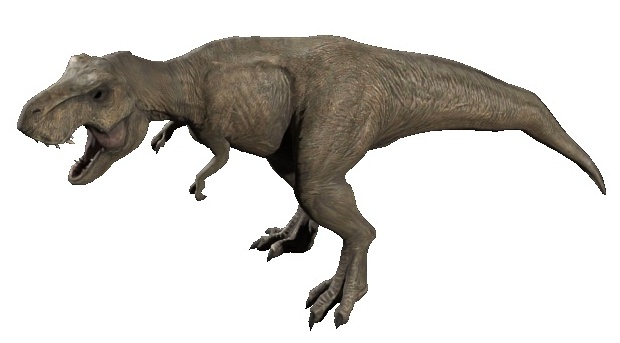} &
\includegraphics[width=\synimgw]{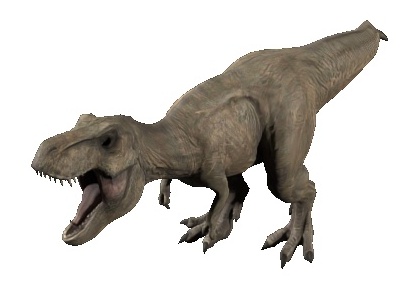} &
\includegraphics[width=\synimgw]{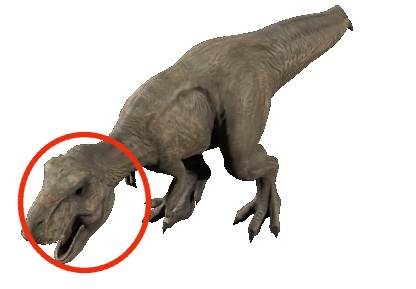} &
\includegraphics[width=\synimgw]{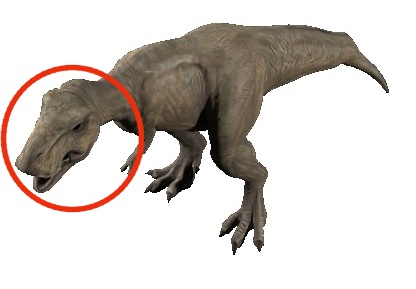} \\
\synlabel{Mani-GS} &
\includegraphics[width=\synimgw]{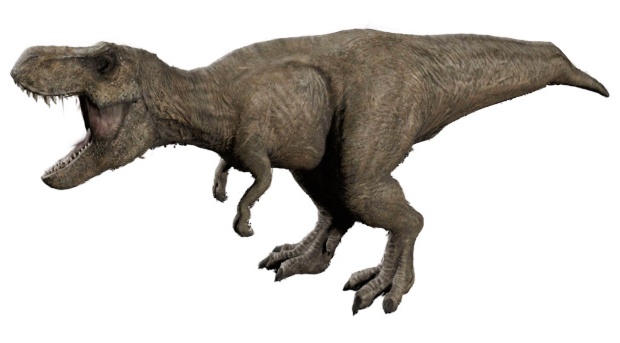} &
\includegraphics[width=\synimgw]{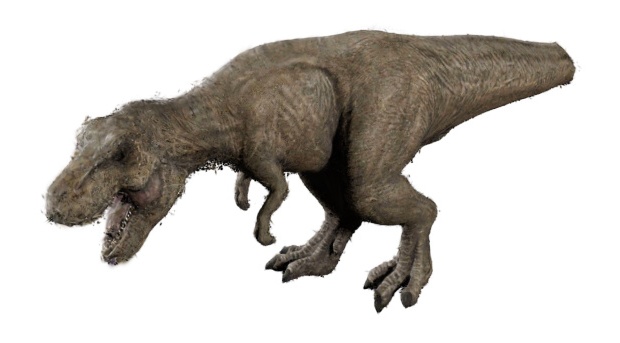} &
\includegraphics[width=\synimgw]{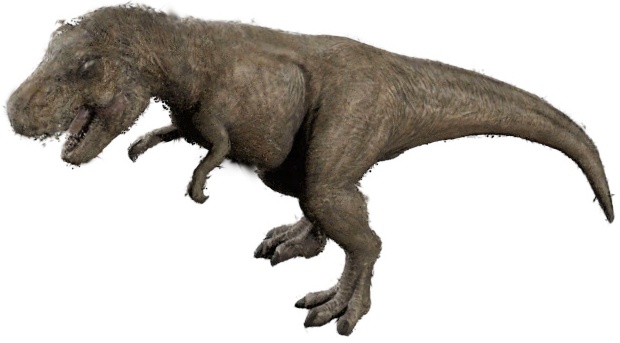} &
\includegraphics[width=\synimgw]{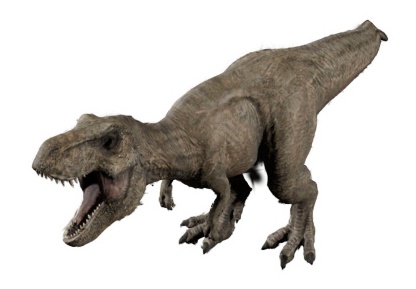} &
\includegraphics[width=\synimgw]{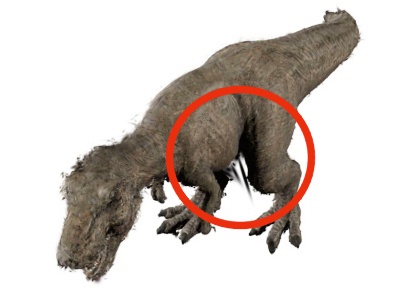} &
\includegraphics[width=\synimgw]{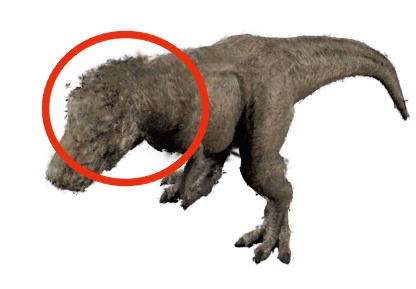} \\
\synlabel{RigGS} &
\includegraphics[width=\synimgw]{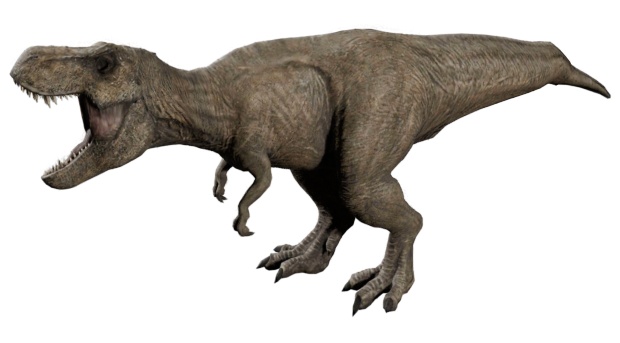} &
\includegraphics[width=\synimgw]{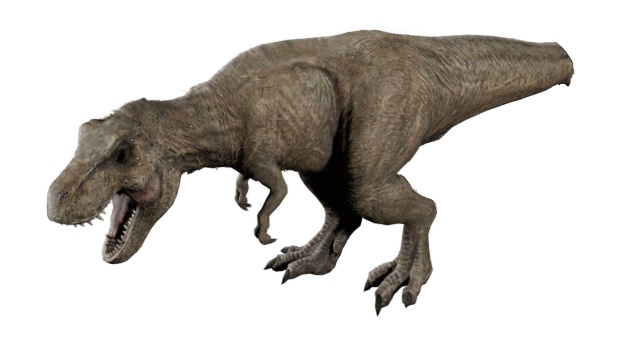} &
\includegraphics[width=\synimgw]{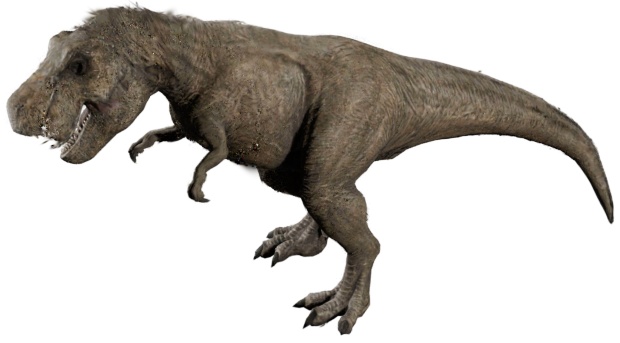} &
\includegraphics[width=\synimgw]{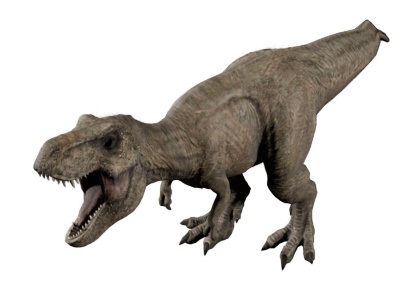} &
\includegraphics[width=\synimgw]{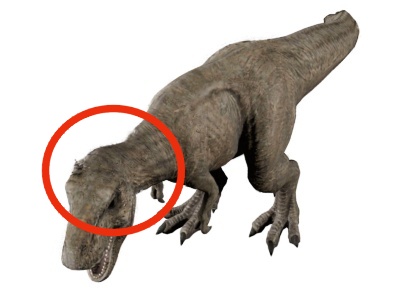} &
\includegraphics[width=\synimgw]{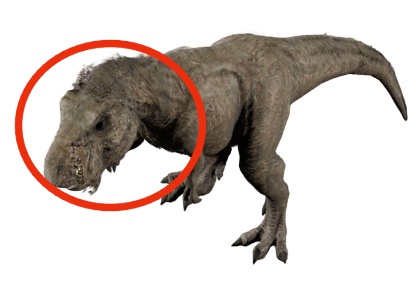} \\
\synlabel{Ours} &
\includegraphics[width=\synimgw]{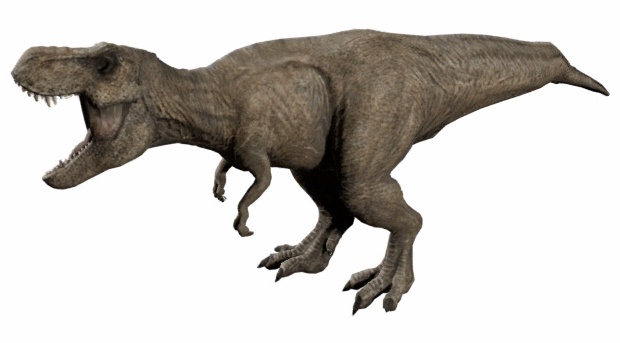} &
\includegraphics[width=\synimgw]{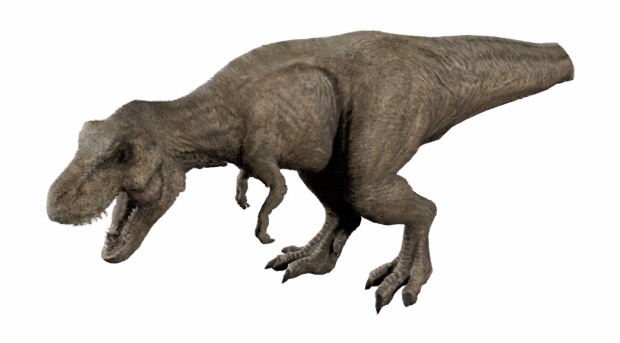} &
\includegraphics[width=\synimgw]{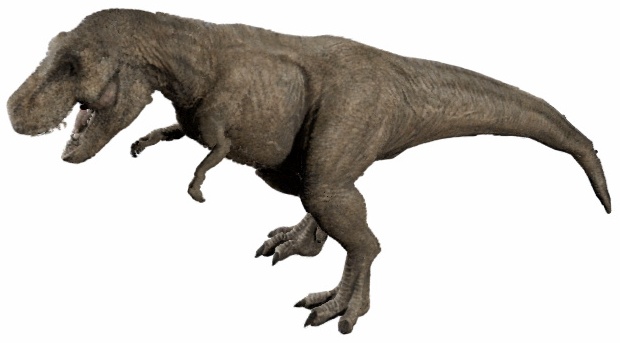} &
\includegraphics[width=\synimgw]{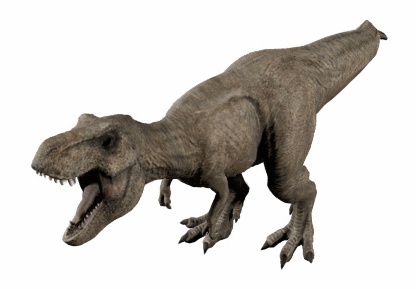} &
\includegraphics[width=\synimgw]{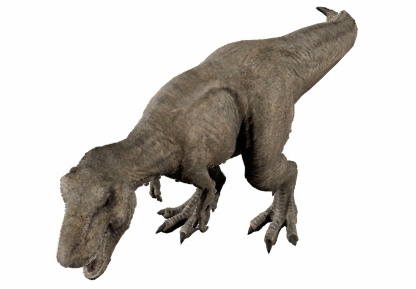} &
\includegraphics[width=\synimgw]{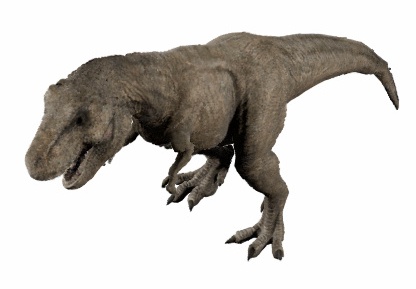} \\
\synlabel{GT} &
\includegraphics[width=\synimgw]{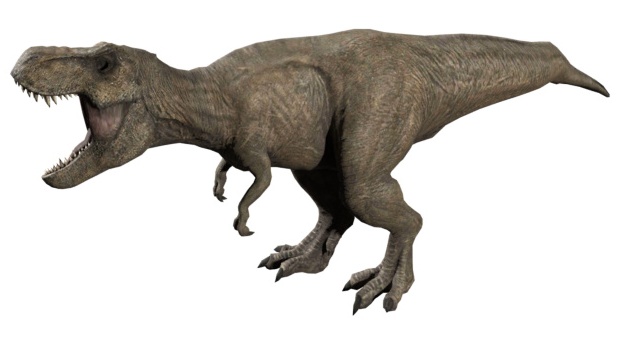} &
\includegraphics[width=\synimgw]{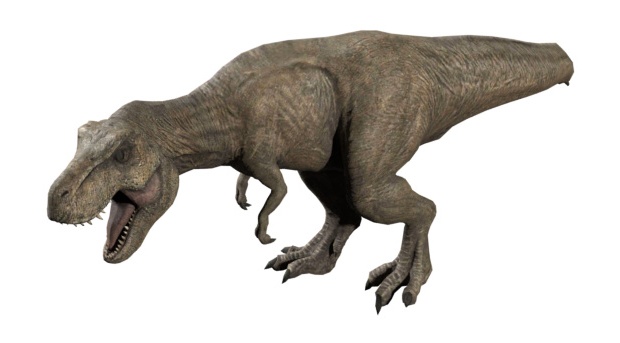} &
\includegraphics[width=\synimgw]{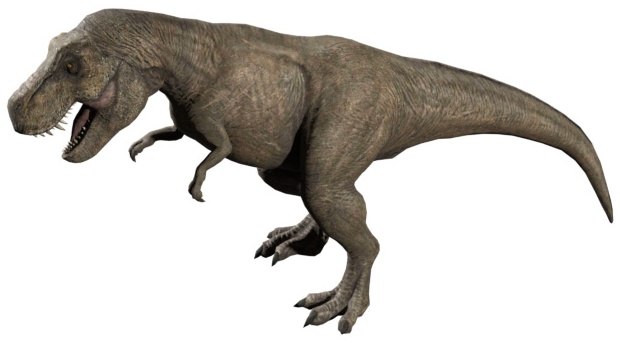} &
\includegraphics[width=\synimgw]{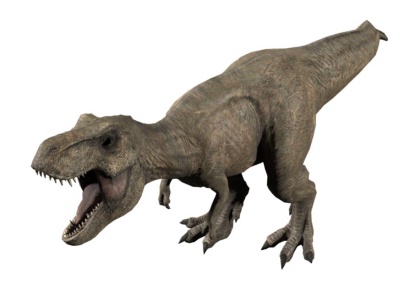} &
\includegraphics[width=\synimgw]{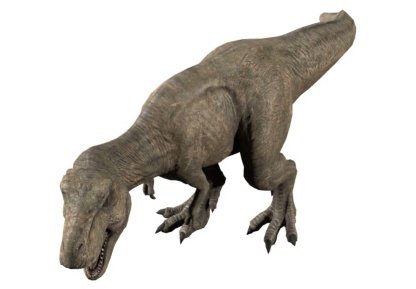} &
\includegraphics[width=\synimgw]{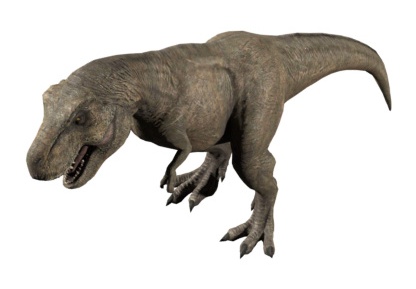} \\
\end{tabular}
\caption{\textbf{Qualitative comparison on synthetic T-Rex.} Rows are methods, columns are frames; left block is the supervised \emph{Training View}, right block is a held-out \emph{Novel View}. Column $t_0$ is the canonical (binding) frame; $t_1, t_2$ are progressively articulated frames. Red highlights mark Puppeteer's joint-rotation errors and the articulation artifacts of Mani-GS and RigGS; best viewed zoomed in.}
\label{fig:qualitative-synthetic}
\end{figure}

\begin{figure}[t]
\centering
\def\zjuimgh{0.172\linewidth}
\def\zjulabelw{0.05\linewidth}
\def\zjublockgap{4pt}
\footnotesize
\setlength{\tabcolsep}{0pt}
\renewcommand{\arraystretch}{0.4}
\newcommand{\zjulabel}[1]{\parbox[b][\zjuimgh][c]{\zjulabelw}{\centering\scriptsize #1}}
\begin{tabular}{@{}c@{\hspace{1pt}}cccc@{\hspace{\zjublockgap}}cccc@{}}
& \multicolumn{4}{c}{\textbf{Training View}} & \multicolumn{4}{c}{\textbf{Novel View}} \\
\cmidrule(lr){2-5}\cmidrule(lr){6-9}
& {\scriptsize Mani-GS} & {\scriptsize RigGS} & {\scriptsize Ours} & {\scriptsize GT}
& {\scriptsize Mani-GS} & {\scriptsize RigGS} & {\scriptsize Ours} & {\scriptsize GT} \\
\zjulabel{$t_0$} &
\includegraphics[height=\zjuimgh]{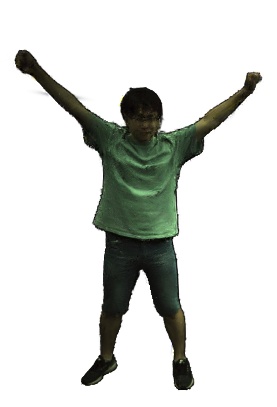} & \includegraphics[height=\zjuimgh]{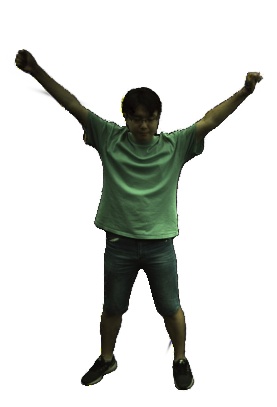} & \includegraphics[height=\zjuimgh]{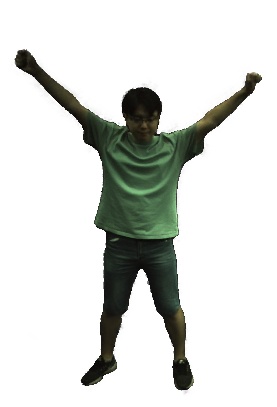} & \includegraphics[height=\zjuimgh]{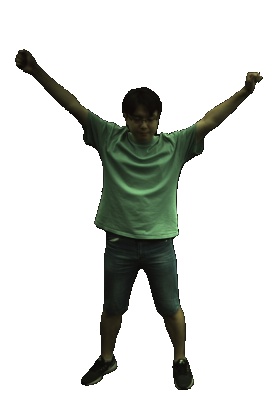} &
\includegraphics[height=\zjuimgh]{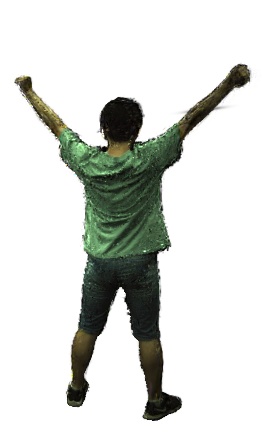} & \includegraphics[height=\zjuimgh]{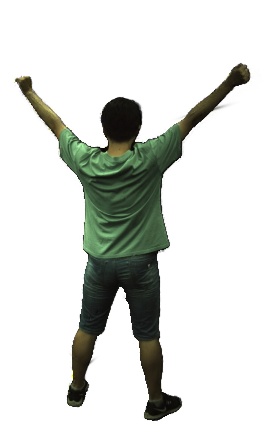} & \includegraphics[height=\zjuimgh]{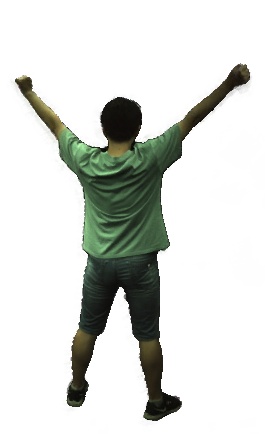} & \includegraphics[height=\zjuimgh]{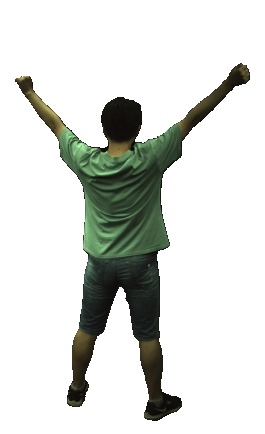} \\
\zjulabel{$t_1$} &
\includegraphics[height=\zjuimgh]{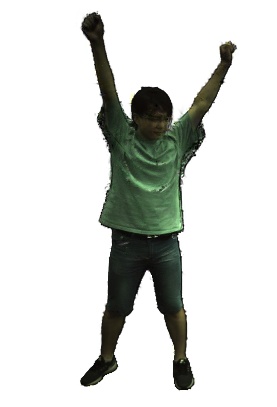} & \includegraphics[height=\zjuimgh]{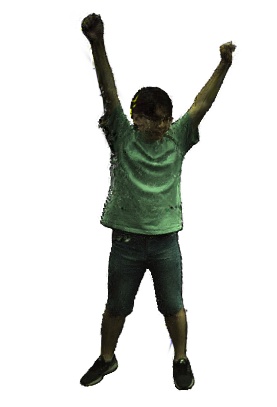} & \includegraphics[height=\zjuimgh]{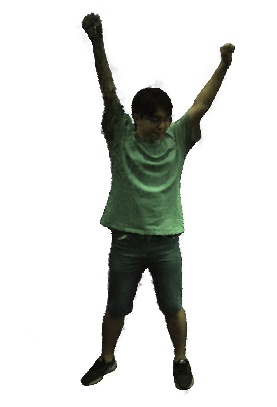} & \includegraphics[height=\zjuimgh]{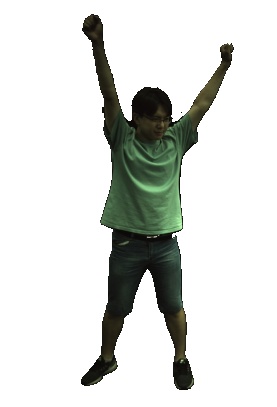} &
\includegraphics[height=\zjuimgh]{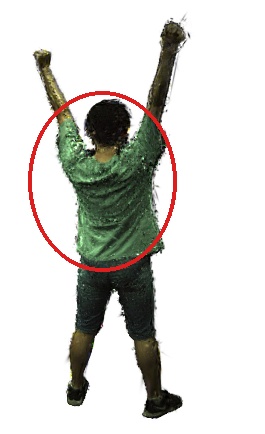} & \includegraphics[height=\zjuimgh]{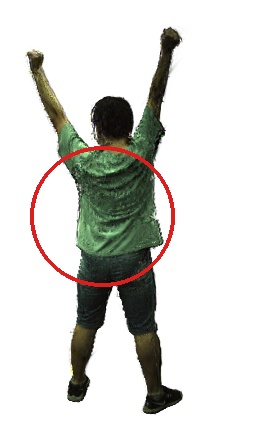} & \includegraphics[height=\zjuimgh]{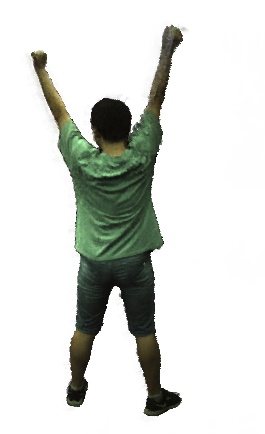} & \includegraphics[height=\zjuimgh]{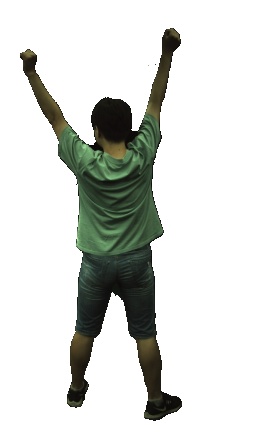} \\
\end{tabular}

\caption{\textbf{Qualitative comparison on ZJU-MoCap.} Rows are frames; columns are split into the supervised \emph{Training View} (left) and a held-out \emph{Novel View} (right), each showing Mani-GS~\citep{manigs}, RigGS~\citep{yao2025riggs}, ours and the ground truth. $t_0$ is the canonical (binding) frame; $t_1$ is an articulated frame. In the novel view, our renderings are cleaner, without the spikes or gaps of the baselines (red); best viewed zoomed in.}
\label{fig:qualitative-realworld}
\end{figure}

\paragraph{Real-world qualitative.} Figure~\ref{fig:qualitative-realworld} shows one subject from ZJU-MoCap at the training view and at a held-out view; a second subject is in the appendix (Figure~\ref{fig:qualitative-zju390-supp}). At the training view all methods reproduce the captured frames. At the held-out view each baseline degrades in its own way: Mani-GS shows needle-like spikes over the back and head, and RigGS opens gaps at the waist, while ours remains clean.

\begin{figure}[t]
\centering
\def\npimgw{0.119\linewidth}
\footnotesize
\setlength{\tabcolsep}{0pt}
\renewcommand{\arraystretch}{0.4}
\begin{tabular}{@{}cccc@{\hspace{4pt}}cccc@{}}
\multicolumn{4}{c}{\textbf{Unseen pose 1}} & \multicolumn{4}{c}{\textbf{Unseen pose 2}} \\
\cmidrule(lr){1-4}\cmidrule(lr){5-8}
{\scriptsize Mani-GS} & {\scriptsize RigGS} & {\scriptsize Ours} & {\scriptsize GT} &
{\scriptsize Mani-GS} & {\scriptsize RigGS} & {\scriptsize Ours} & {\scriptsize GT} \\
\includegraphics[width=\npimgw]{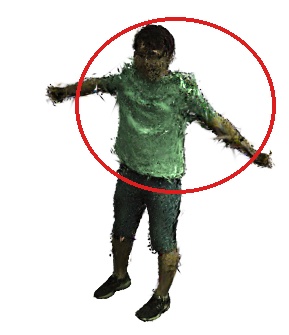} & \includegraphics[width=\npimgw]{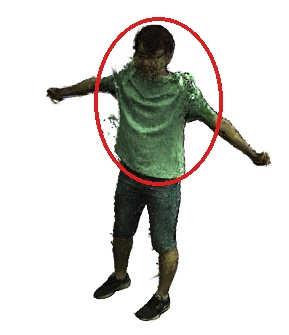} & \includegraphics[width=\npimgw]{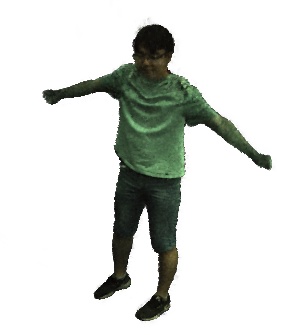} & \includegraphics[width=\npimgw]{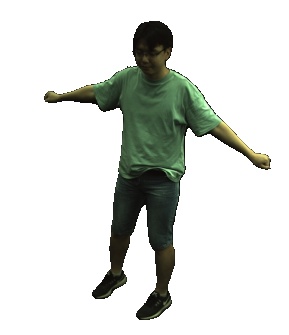} &
\includegraphics[width=\npimgw]{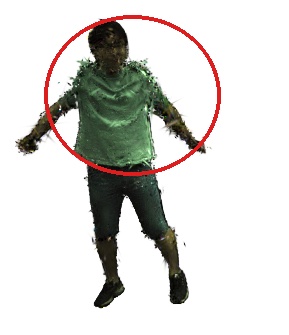} & \includegraphics[width=\npimgw]{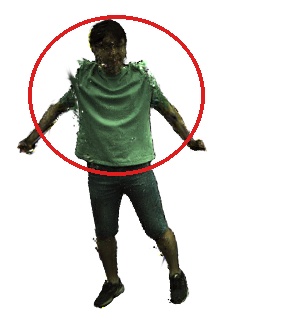} & \includegraphics[width=\npimgw]{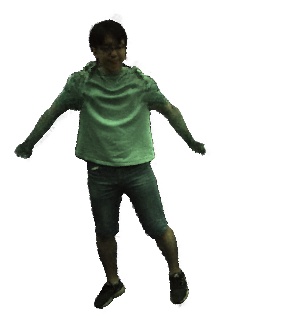} & \includegraphics[width=\npimgw]{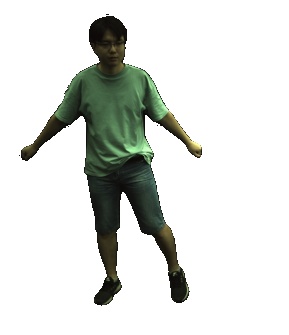} \\
\end{tabular}
\caption{\textbf{Novel pose driving on ZJU-MoCap.} Each method is driven at test time with poses from a later part of the capture that was not part of the fitting trajectory; the ground truth is the captured frame. Columns show Mani-GS, RigGS, ours and the ground truth. Our renderings are cleaner, without the spikes or gaps of the baselines (red). Best viewed zoomed in.}
\label{fig:motion-transfer}
\end{figure}

\paragraph{Novel pose driving.} Figure~\ref{fig:motion-transfer} drives each fitted model with poses from a later part of the ZJU-MoCap capture that were not part of the fitting trajectory, retargeted to each method's own rig. Only pose-independent outputs carry over to unseen poses: the refined skinning weights for every method, and for Mani-GS additionally the refined per-splat binding rotation and offset; components tied to the training frames (our Phase~2 correction MLP and RigGS's detail head) are disabled at test time. Both splatting-based baselines show artifacts at the joints, while ours does not.

\begin{figure}[t]
\centering
\def\capmainh{0.255\linewidth}
\footnotesize
\setlength{\tabcolsep}{0pt}
\renewcommand{\arraystretch}{0.4}
\begin{tabular}{@{}ccc@{\hspace{8pt}}ccc@{}}
\multicolumn{3}{c}{\textbf{Robot}} & \multicolumn{3}{c}{\textbf{Statue}} \\
\cmidrule(lr){1-3}\cmidrule(lr){4-6}
{\scriptsize\makecell{Generated\\driving video}} & {\scriptsize\makecell{Ours\\(train view)}} & {\scriptsize\makecell{Ours\\(novel view)}} &
{\scriptsize\makecell{Generated\\driving video}} & {\scriptsize\makecell{Ours\\(train view)}} & {\scriptsize\makecell{Ours\\(novel view)}} \\
\includegraphics[height=\capmainh]{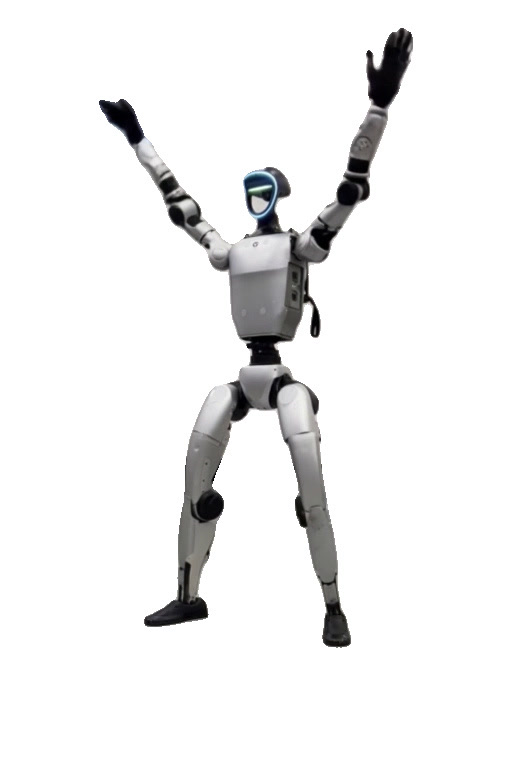} & \includegraphics[height=\capmainh]{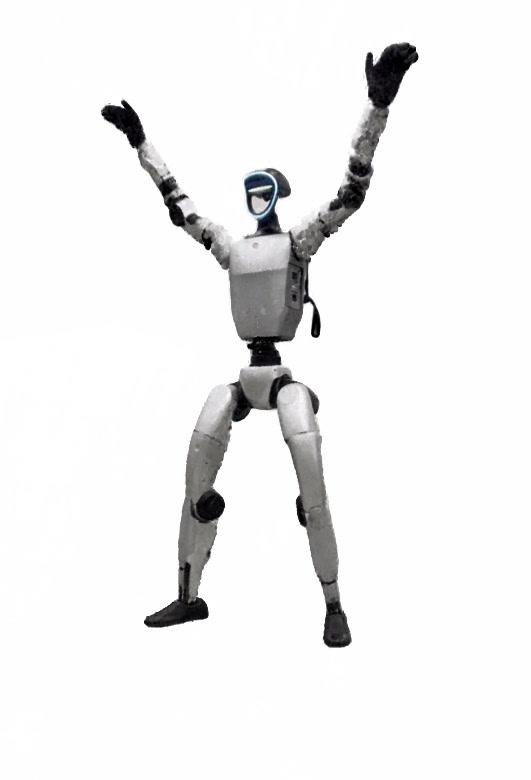} & \includegraphics[height=\capmainh]{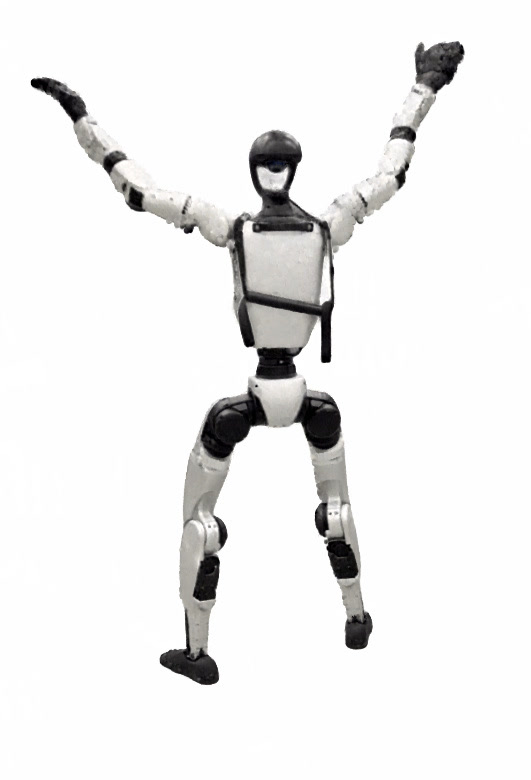} &
\includegraphics[height=\capmainh]{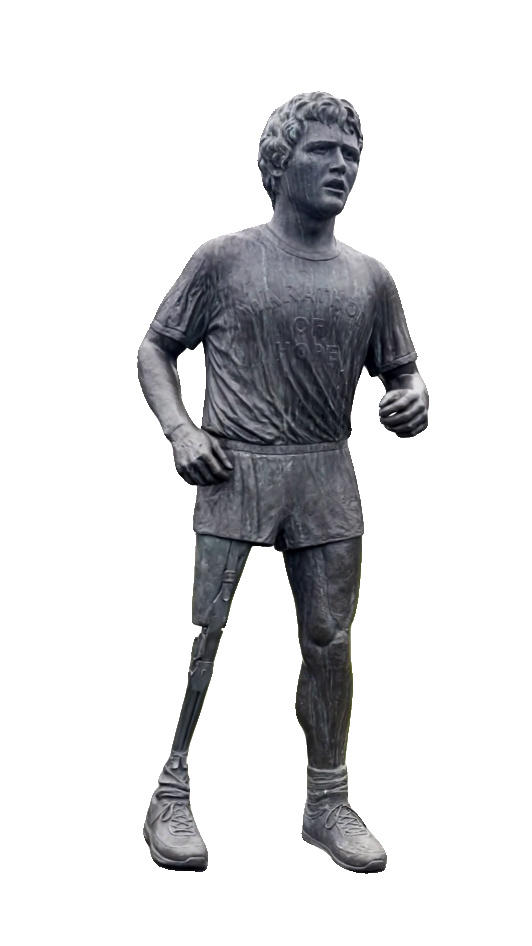} & \includegraphics[height=\capmainh]{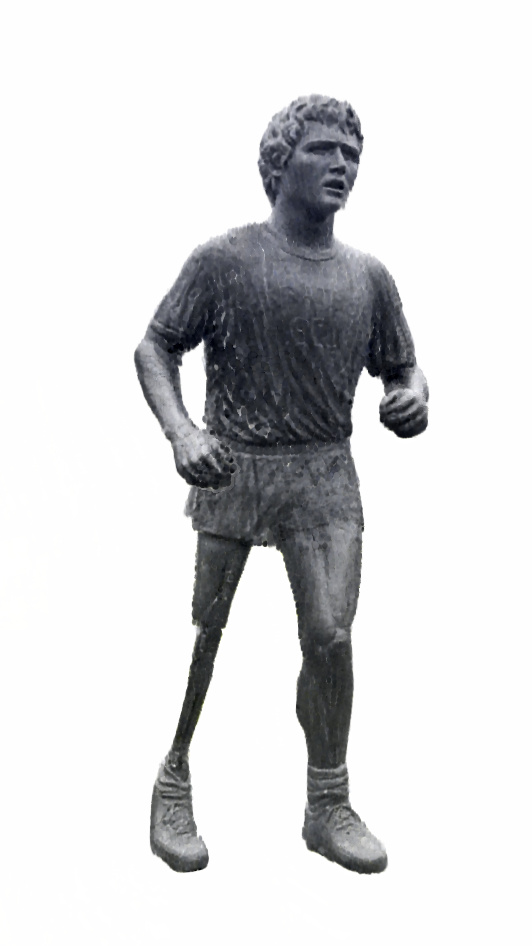} & \includegraphics[height=\capmainh]{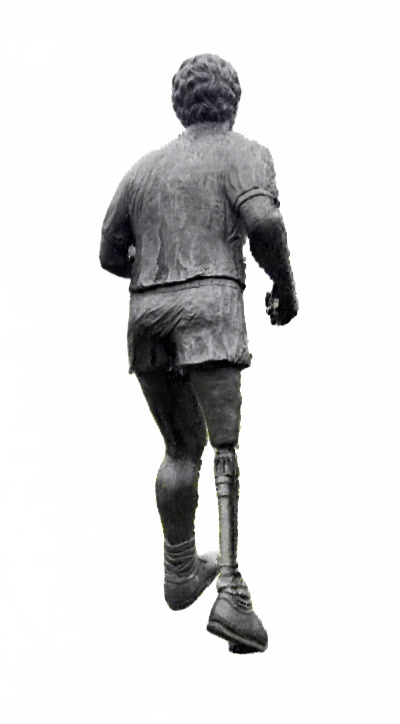} \\
\end{tabular}
\caption{\textbf{Captured scenes.} Each static capture is animated from a generated driving video. For each scene, columns show one frame of the driving video and our result at the training view and at a novel view.}
\label{fig:captured-main}
\end{figure}

\paragraph{Captured scenes.} Figure~\ref{fig:captured-main} shows our results on the two captured scenes, each animated from its generated driving video. At the training view our model follows the generated motion, and from a novel view it renders a clean surface as the subject moves. More frames and novel pose driving of the robot are in the appendix (Figures~\ref{fig:qualitative-realworld-robot-supp}--\ref{fig:supp-robot-novel-pose}).

\subsection{Ablations} \label{sec:ablations}

\begin{figure}[t]
\centering
\def\gtmainw{0.18\linewidth}
\footnotesize
\setlength{\tabcolsep}{0pt}
\renewcommand{\arraystretch}{0.4}
\begin{tabular}{@{}c@{\hspace{4pt}}c@{\hspace{4pt}}c@{\hspace{4pt}}c@{}}
{\scriptsize Mani-GS} & {\scriptsize Mani-GS (isotropic)} & {\scriptsize RigGS} & {\scriptsize Ours} \\
\includegraphics[width=\gtmainw]{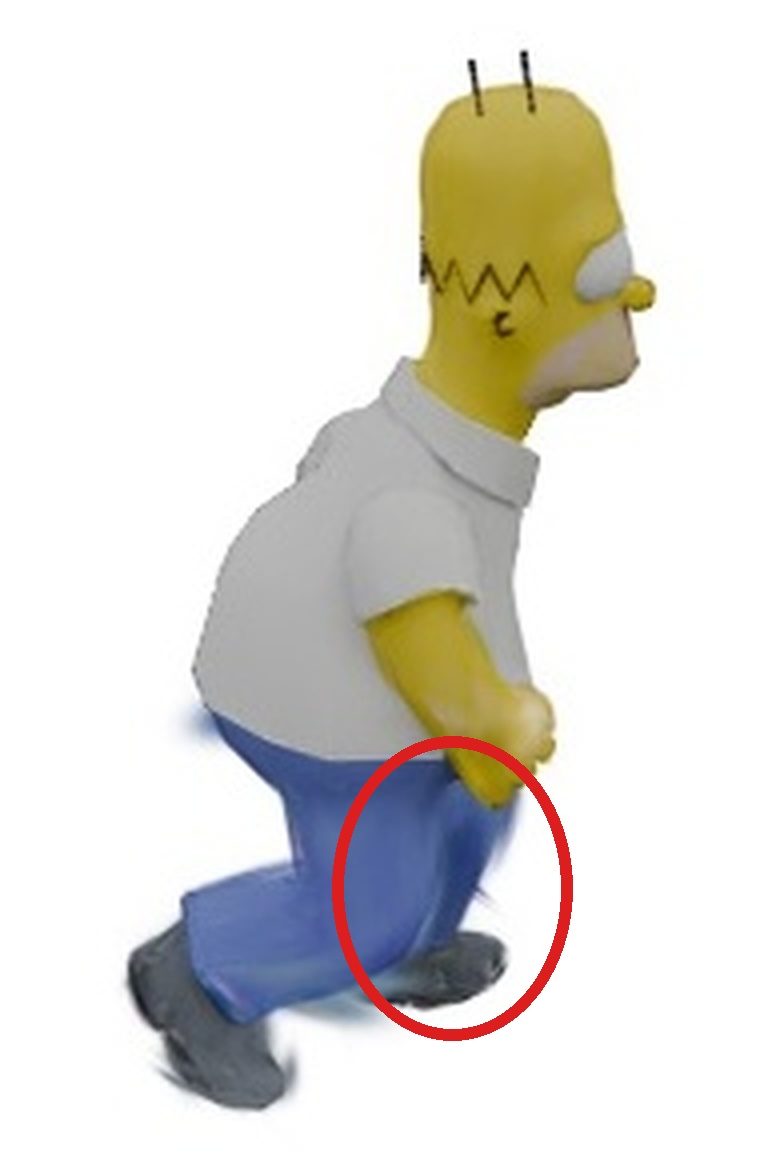} &
\includegraphics[width=\gtmainw]{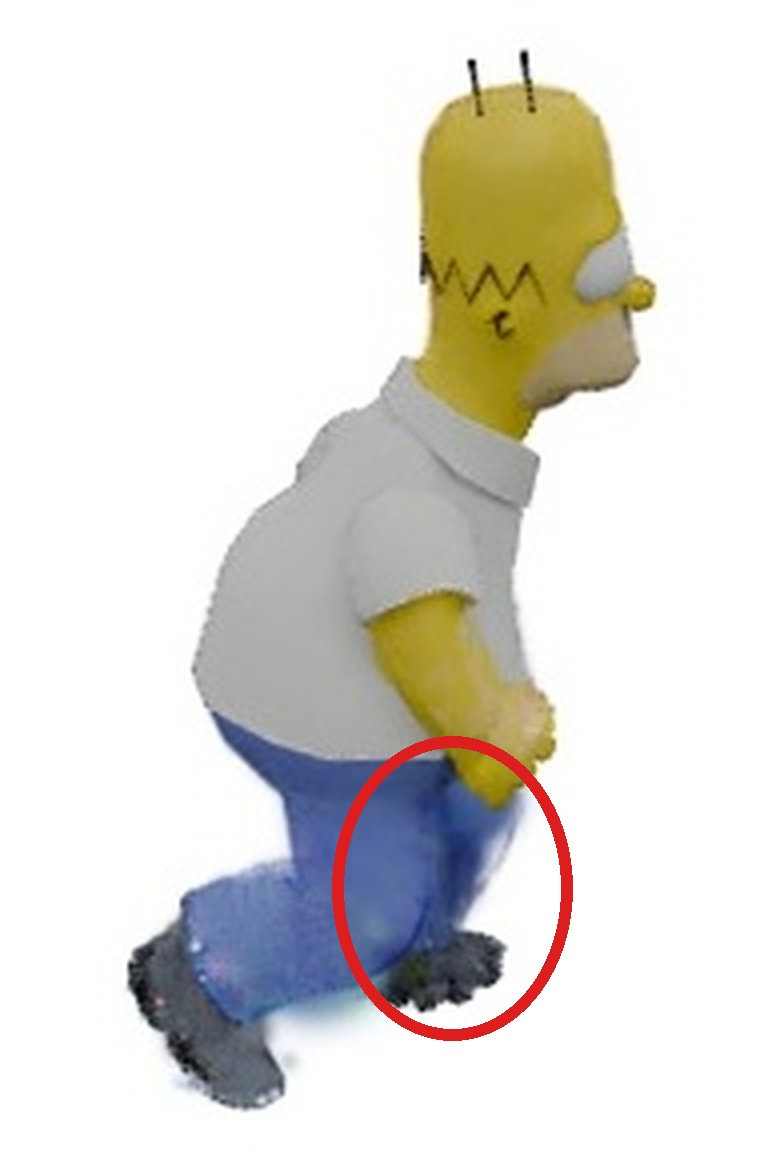} &
\includegraphics[width=\gtmainw]{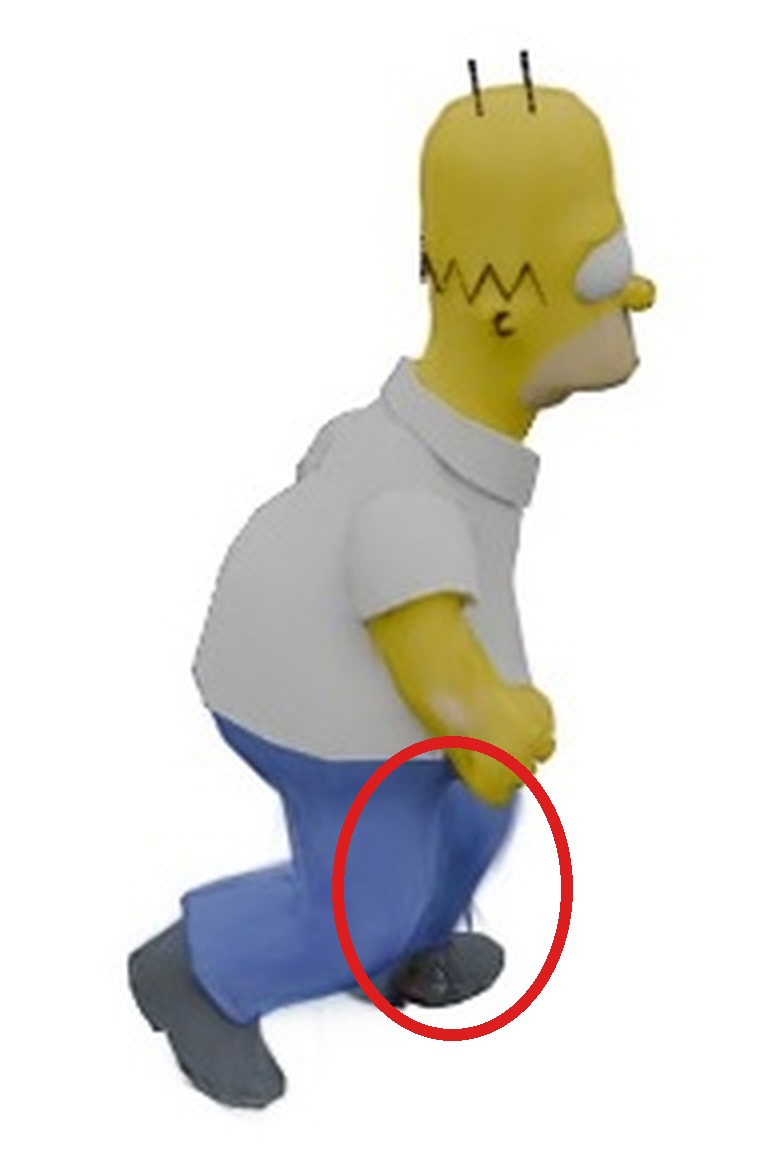} &
\includegraphics[width=\gtmainw]{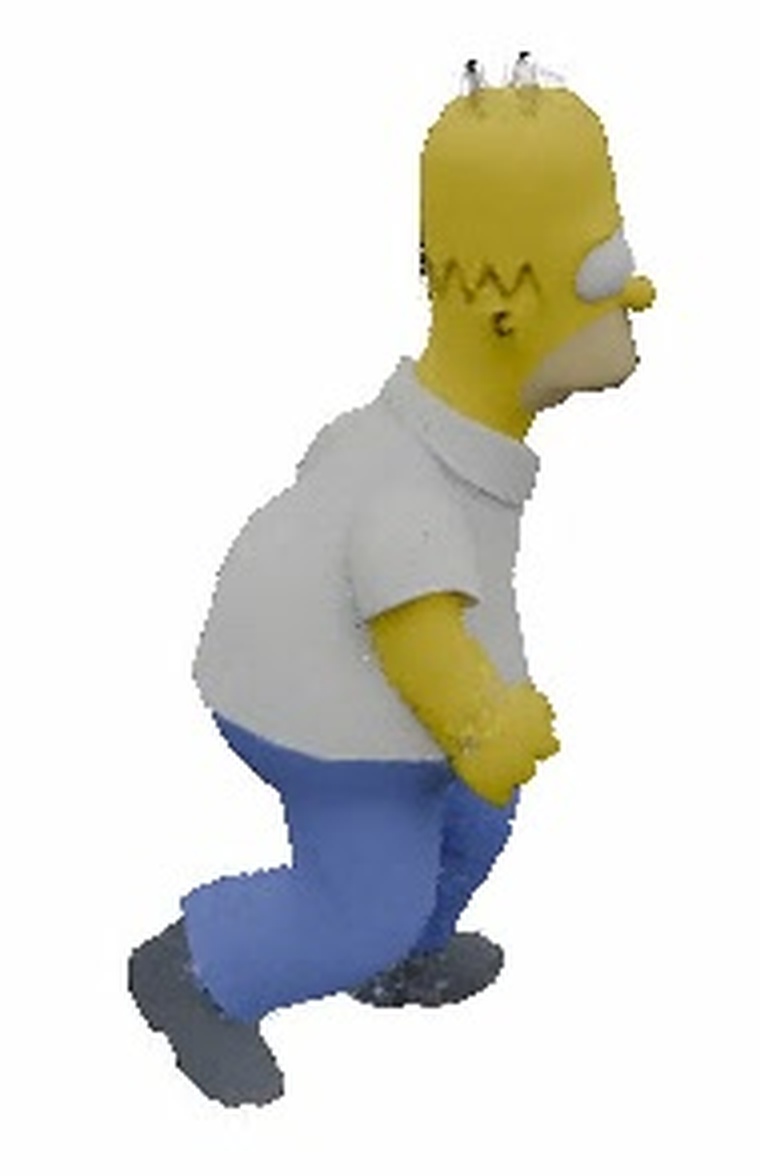} \\
\end{tabular}
\caption{\textbf{Ground-truth deformation on Simpsons.} Every method is deformed with the ground truth, so no rig error enters the comparison. The splat-based methods produce artifacts at the legs when the knees bend (red); ours does not. Best viewed zoomed in.}
\label{fig:gt-deform-main}
\end{figure}

\paragraph{Ground-truth deformation.} To separate the rendering representation from rig error, we deform every method with the ground-truth motion of the synthetic subject and fine-tune none of them per frame. We compare three splat variants with ours: Mani-GS~\citep{manigs}, which binds its splats to the ground-truth mesh; its isotropic variant; and RigGS~\citep{yao2025riggs}, which skins its splats directly with LBS, with its pose-dependent correction head omitted since the deformation is given. Every splat variant uses more primitives than ours, from about two to fourteen times as many. Even so, on Simpsons (Figure~\ref{fig:gt-deform-main}), all three produce artifacts at the legs when the knees bend, while ours does not. At the binding frame all methods render comparably, and fox and T-Rex show the same pattern (appendix~\ref{sec:gt-mesh-ablation}). With the deformation exact, the artifact remains whether the splats follow a mesh or are skinned directly, and whether they are anisotropic or isotropic; it disappears only when the primitive changes.

\paragraph{Two-phase optimization.} Table~\ref{tab:phase-rexy} ablates the two phases of our method on T-Rex, including the relative-depth term of Phase~1. Without $\mathcal{L}_{\Delta\text{depth}}$, Phase~1 fits the training view more closely, by 1.5~dB PSNR, yet the chamfer distance nearly triples and novel-view PSNR drops by 2.7~dB. From a single fixed view, a joint rotated wrongly in depth can still match the pixels well, so a closer training-view fit does not mean a more accurate 3D pose. Phase~2 then raises training-view quality and lowers the chamfer distance further by correcting Phase~1's joint-rotation errors, such as the T-Rex head and tail in Figure~\ref{fig:pipeline}, while novel-view quality stays about the same. Across methods, Phase~2 likewise raises training-view PSNR while leaving novel-view quality essentially unchanged (appendix~\ref{sec:supp-phase-ablation}).

\begin{table}[t]
\centering
\caption{\textbf{Two-phase ablation of our method on T-Rex.} Training- and novel-view metrics use the same cameras as Table~\ref{tab:quantitative}; chamfer distance is measured against the ground-truth mesh and averaged over the driving-video frames. Removing $\mathcal{L}_{\Delta\text{depth}}$ improves the training view at the cost of novel-view quality and 3D accuracy; Phase~2 improves both the training view and the chamfer distance.}
\label{tab:phase-rexy}
\footnotesize
\setlength{\tabcolsep}{3pt}
\begin{tabular}{l|ccc|ccc|c}
\toprule
 & \multicolumn{3}{c|}{Training view} & \multicolumn{3}{c|}{Novel view} & \\
Configuration & PSNR$\uparrow$ & SSIM$\uparrow$ & LPIPS$\downarrow$
              & PSNR$\uparrow$ & SSIM$\uparrow$ & LPIPS$\downarrow$
              & Chamfer ($\times 10^{-3}$)$\downarrow$ \\
\midrule
Phase~1 w/o $\mathcal{L}_{\Delta\text{depth}}$                & 29.21 & 0.953 & 0.034 & 21.38 & 0.934 & 0.045 & 15.64 \\
Phase~1                                                       & 27.72 & 0.945 & 0.035 & 24.05 & 0.942 & 0.039 & 5.68 \\
Full                                                          & 31.08 & 0.972 & 0.025 & 23.95 & 0.945 & 0.034 & 4.60 \\
\bottomrule
\end{tabular}
\end{table}

\paragraph{Loss-weight sensitivity.} We scale each loss-weight group by $\times\frac{1}{3}$ and $\times 3$ on T-Rex and fox, with all other weights at their defaults (appendix~\ref{sec:supp-sensitivity}). Novel-view PSNR varies by at most 0.54~dB for every group except the relative-depth weight (1.55~dB), which matters most, consistent with Table~\ref{tab:phase-rexy}.

\section{Final Discussion and Conclusion}

\paragraph{Limitations.}
RigPAPR models skeletal articulation under LBS; muscle bulging, soft tissue, cloth, hair, and facial motion are not modelled. Within this scope, RigPAPR inherits the failure modes of its 2D scaffolding. Phase~1 relies on monocular depth and Phase~2 further on 2D tracking; where these priors fail, the errors propagate into the recovered articulation and cannot be corrected by the rendering loss, since the single fixed view does not constrain the failure direction. The auto-rigging stage is similarly load-bearing: a poor skeleton or miscomputed skinning seeds an optimum that downstream refinement only partially escapes, particularly on morphologies the auto-rigger was not trained for. Finally, regions occluded in the binding frame receive no appearance supervision during canonical pretraining, so articulation that later exposes them tends to render as holes or stray colors rather than a coherent surface.

\paragraph{Conclusion.}
We traced joint-boundary artifacts in Gaussian-splat-based rigging methods to a representation-level incompatibility with LBS, and showed that covariance-free, interpolation-based point primitives address the underlying cause rather than patching the symptom. RigPAPR auto-rigs a static PAPR cloud and animates it from a fixed-viewpoint driving video, captured or I2V-generated. It renders cleanly under articulation at both the supervised and novel views. The rigged cloud is a reusable 3D asset, re-posable to new motions beyond the driving video.

\begin{ack}
We thank Dun Yang for allowing us to capture the robot and for helping us set it up.
This research was enabled in part by support provided by NSERC, the BC DRI Group, the Digital Research Alliance of Canada and the Canada CIFAR AI Chairs Program.
We declare no competing interests.
\end{ack}

\clearpage
\newpage
\bibliography{neurips_2026}
\bibliographystyle{plain}

\clearpage
\newpage
\appendix
This appendix is organized into four parts. Section~\ref{sec:supp-qualitative} extends the qualitative comparison of the main paper to the four remaining synthetic subjects (fox, Simpsons, wolf, spider) and to the second ZJU-MoCap subject, and shows our results on two scenes we captured ourselves (robot, statue), animated from a generated video and driven to novel poses (Section~\ref{sec:supp-realworld}). Section~\ref{sec:supp-ablations} groups two additional ablations: the two-phase optimization ablation across methods, with a closer look at our method and its depth supervision on T-Rex (Section~\ref{sec:supp-phase-ablation}), and a ground-truth deformation ablation that isolates the rendering representation from geometry and skinning quality (Section~\ref{sec:gt-mesh-ablation}). Section~\ref{sec:supp-sensitivity} reports the sensitivity of the two-phase optimization to its loss weights. Section~\ref{sec:supp-impl} collects additional implementation details: dataset sequence lengths (Section~\ref{sec:supp-data-lengths}), the PAPR backbone and auto-rig pipeline (Section~\ref{sec:supp-backbone}), the baseline implementations (Section~\ref{sec:supp-baselines}) and a Mani-GS configuration comparison that motivates disabling its per-triangle adaptive scaling (Section~\ref{sec:supp-mani-scaling}), the two-phase optimization losses and schedules (Section~\ref{sec:supp-optim}), and the attention-guided track-seeding procedure (Section~\ref{sec:supp-tracks}).

\section{Additional Qualitative Results} \label{sec:supp-qualitative}

\begin{figure}[!htbp]
\centering
\def\synimgw{0.20\linewidth}  %
\def\synimgh{0.133\linewidth}
\footnotesize
\setlength{\tabcolsep}{0pt}
\renewcommand{\arraystretch}{0.4}

\begin{tabular}{@{}c@{\hspace{1pt}}ccc@{}}
\multicolumn{4}{@{}l}{\textbf{Training View}} \\
\cmidrule(lr){2-4}
& {\scriptsize $t_0$ (canonical)} & {\scriptsize $t_1$} & {\scriptsize $t_2$} \\
\synlabel{Puppeteer} &
\includegraphics[width=\synimgw]{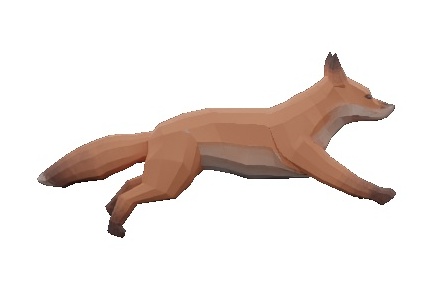} &
\includegraphics[width=\synimgw]{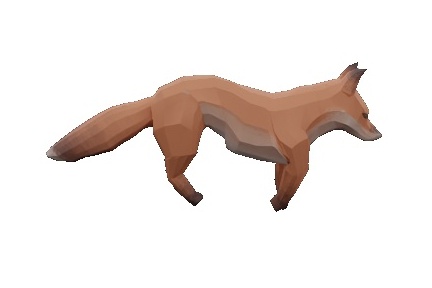} &
\includegraphics[width=\synimgw]{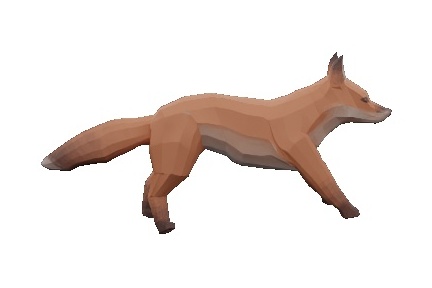} \\
\synlabel{Mani-GS} &
\includegraphics[width=\synimgw]{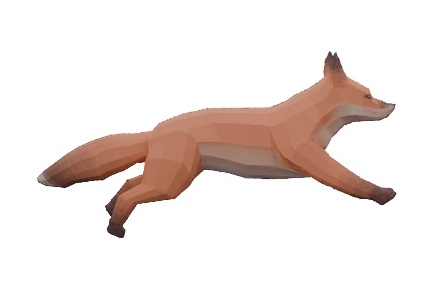} &
\includegraphics[width=\synimgw]{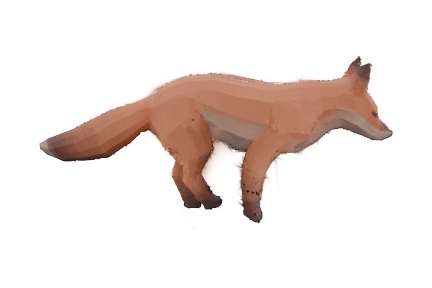} &
\includegraphics[width=\synimgw]{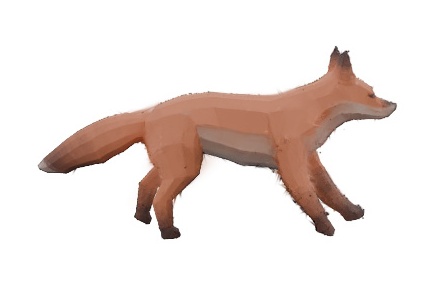} \\
\synlabel{RigGS} &
\includegraphics[width=\synimgw]{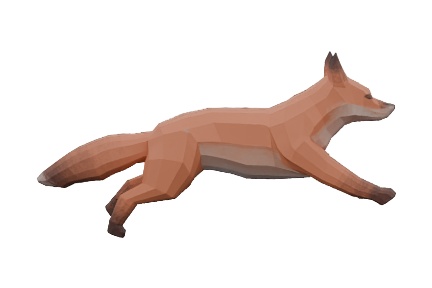} &
\includegraphics[width=\synimgw]{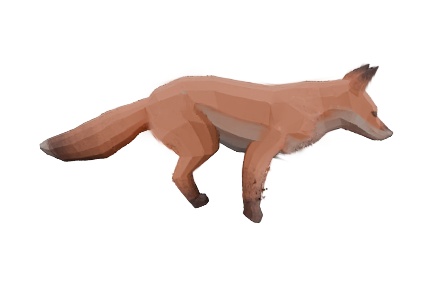} &
\includegraphics[width=\synimgw]{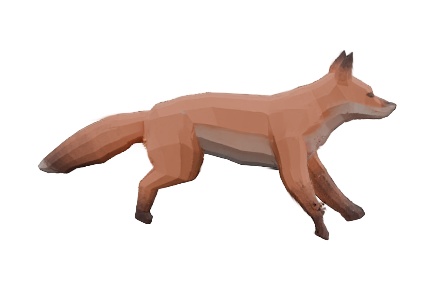} \\
\synlabel{Ours} &
\includegraphics[width=\synimgw]{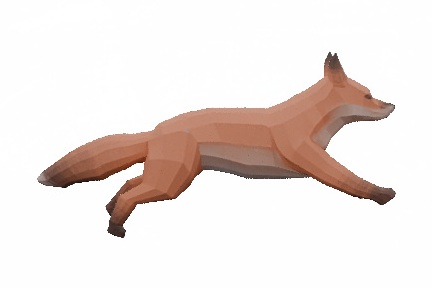} &
\includegraphics[width=\synimgw]{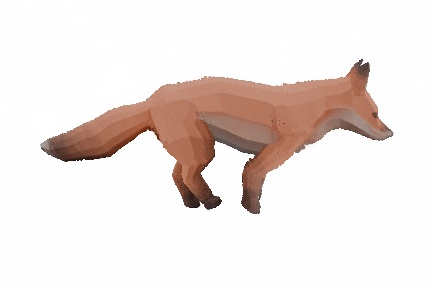} &
\includegraphics[width=\synimgw]{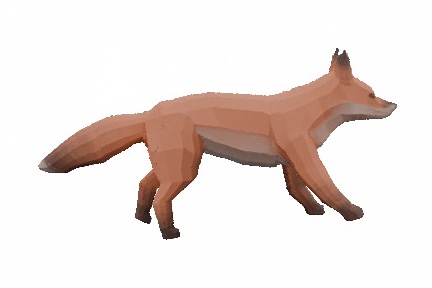} \\
\synlabel{GT} &
\includegraphics[width=\synimgw]{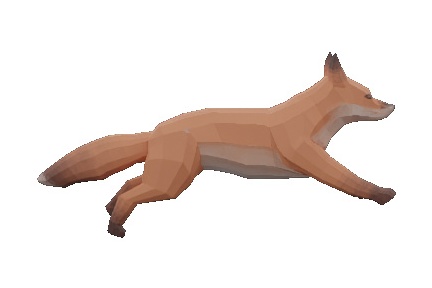} &
\includegraphics[width=\synimgw]{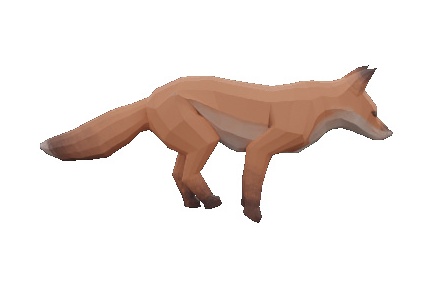} &
\includegraphics[width=\synimgw]{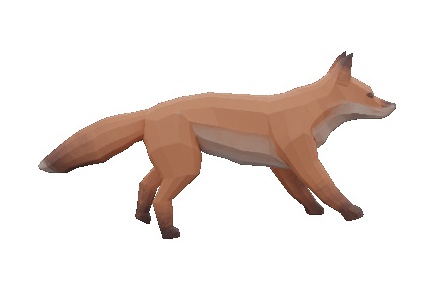} \\[6pt]
\multicolumn{4}{@{}l}{\textbf{Novel View}} \\
\cmidrule(lr){2-4}
& {\scriptsize $t_0$ (canonical)} & {\scriptsize $t_1$} & {\scriptsize $t_2$} \\
\synlabel{Puppeteer} &
\includegraphics[width=\synimgw]{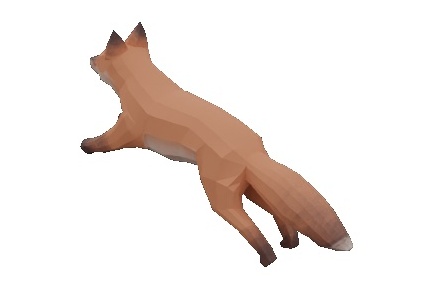} &
\includegraphics[width=\synimgw]{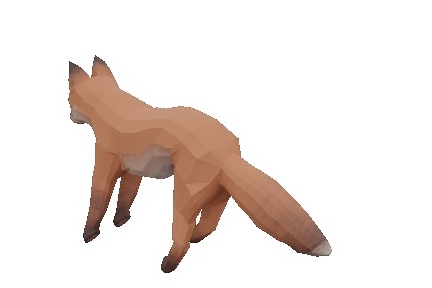} &
\includegraphics[width=\synimgw]{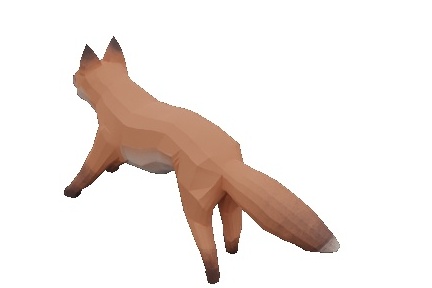} \\
\synlabel{Mani-GS} &
\includegraphics[width=\synimgw]{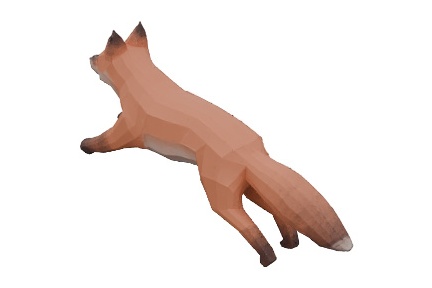} &
\includegraphics[width=\synimgw]{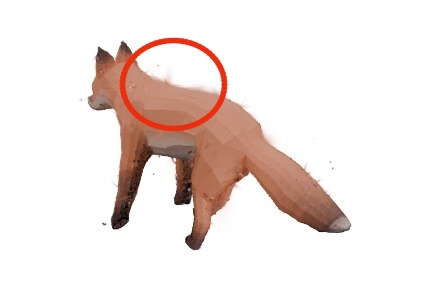} &
\includegraphics[width=\synimgw]{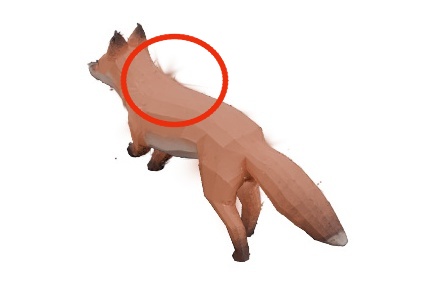} \\
\synlabel{RigGS} &
\includegraphics[width=\synimgw]{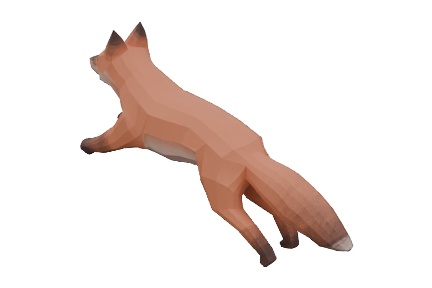} &
\includegraphics[width=\synimgw]{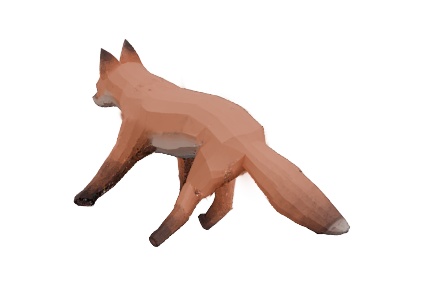} &
\includegraphics[width=\synimgw]{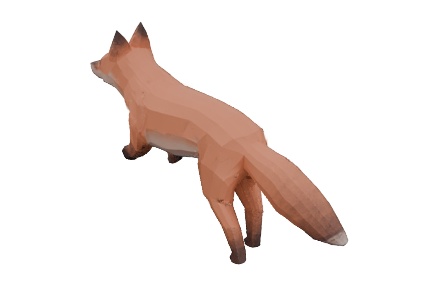} \\
\synlabel{Ours} &
\includegraphics[width=\synimgw]{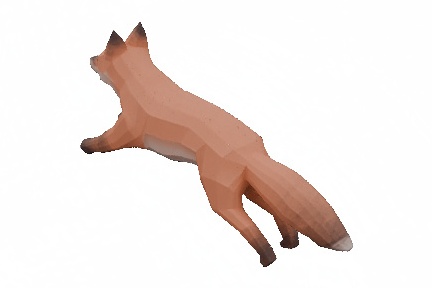} &
\includegraphics[width=\synimgw]{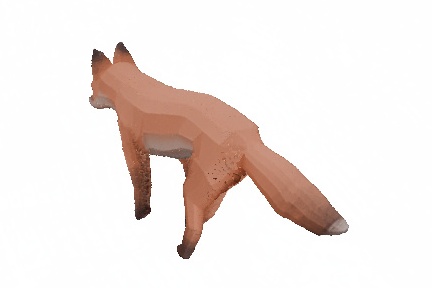} &
\includegraphics[width=\synimgw]{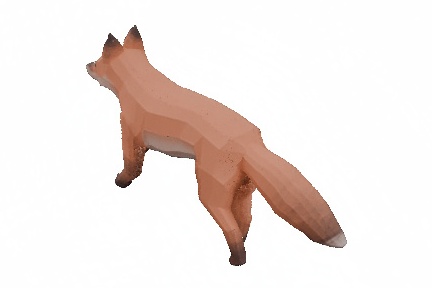} \\
\synlabel{GT} &
\includegraphics[width=\synimgw]{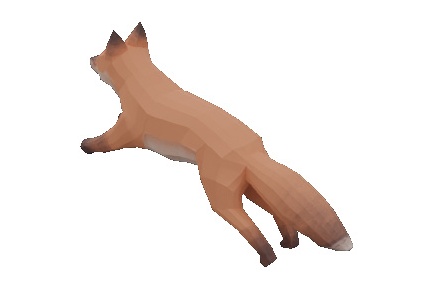} &
\includegraphics[width=\synimgw]{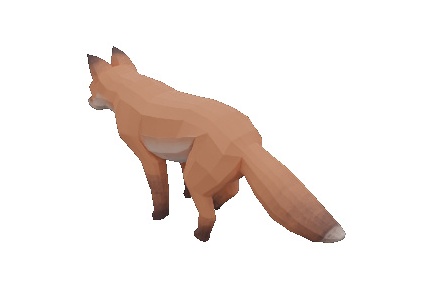} &
\includegraphics[width=\synimgw]{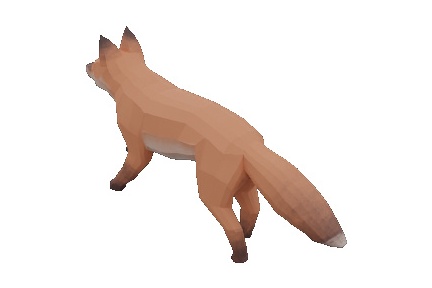} \\
\end{tabular}

\caption{\textbf{Qualitative comparison on synthetic fox.} Training View is on top and Novel View is on the bottom. Rows are methods; columns are matched frames, with $t_0$ the canonical (binding) frame and $t_1,\,t_2$ progressively articulated.}
\label{fig:qualitative-synthetic-fox-supp}
\end{figure}

\begin{figure}[!htbp]
\centering
\def\synimgw{0.13\linewidth}
\def\synimgh{0.26\linewidth}
\footnotesize
\setlength{\tabcolsep}{0pt}
\renewcommand{\arraystretch}{0.4}

\begin{tabular}{@{}c@{\hspace{1pt}}ccc@{\hspace{\synblockgap}}ccc@{}}
& \multicolumn{3}{c}{\textbf{Training View}} & \multicolumn{3}{c}{\textbf{Novel View}} \\
\cmidrule(lr){2-4}\cmidrule(lr){5-7}
& {\scriptsize $t_0$ (canonical)} & {\scriptsize $t_1$} & {\scriptsize $t_2$}
& {\scriptsize $t_0$ (canonical)} & {\scriptsize $t_1$} & {\scriptsize $t_2$} \\
\synlabel{Puppeteer} &
\includegraphics[width=\synimgw]{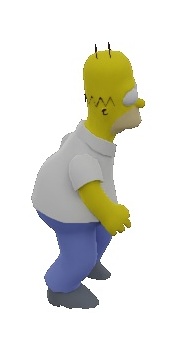} &
\includegraphics[width=\synimgw]{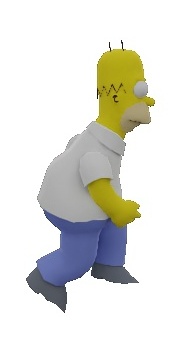} &
\includegraphics[width=\synimgw]{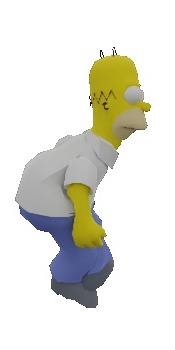} &
\includegraphics[width=\synimgw]{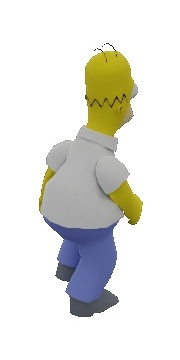} &
\includegraphics[width=\synimgw]{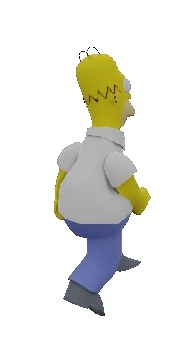} &
\includegraphics[width=\synimgw]{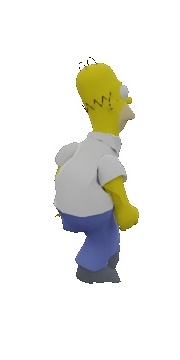} \\
\synlabel{Mani-GS} &
\includegraphics[width=\synimgw]{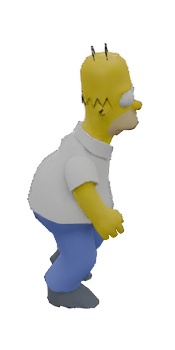} &
\includegraphics[width=\synimgw]{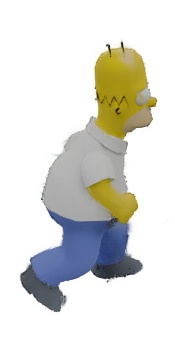} &
\includegraphics[width=\synimgw]{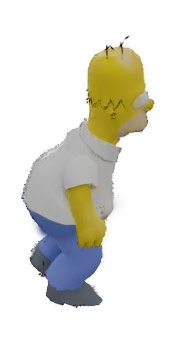} &
\includegraphics[width=\synimgw]{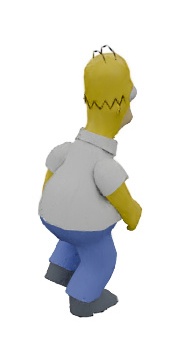} &
\includegraphics[width=\synimgw]{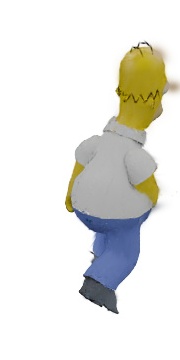} &
\includegraphics[width=\synimgw]{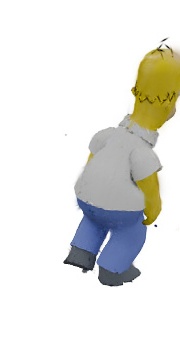} \\
\synlabel{RigGS} &
\includegraphics[width=\synimgw]{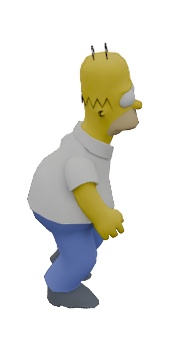} &
\includegraphics[width=\synimgw]{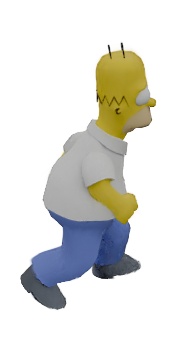} &
\includegraphics[width=\synimgw]{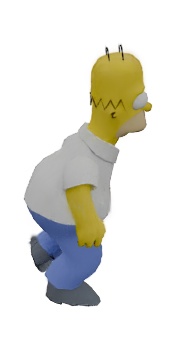} &
\includegraphics[width=\synimgw]{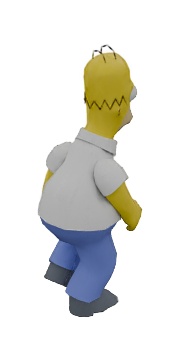} &
\includegraphics[width=\synimgw]{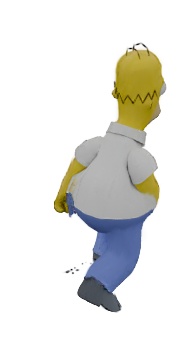} &
\includegraphics[width=\synimgw]{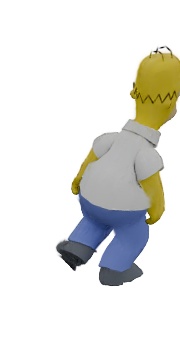} \\
\synlabel{Ours} &
\includegraphics[width=\synimgw]{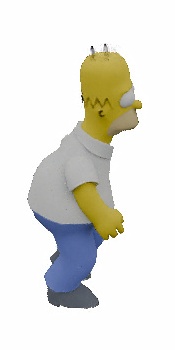} &
\includegraphics[width=\synimgw]{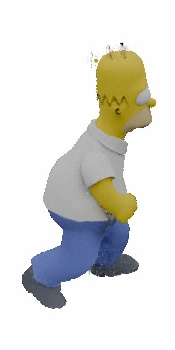} &
\includegraphics[width=\synimgw]{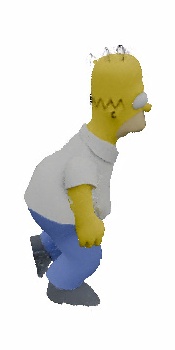} &
\includegraphics[width=\synimgw]{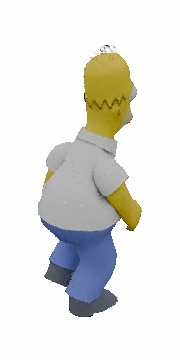} &
\includegraphics[width=\synimgw]{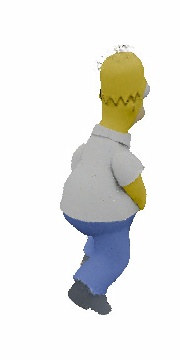} &
\includegraphics[width=\synimgw]{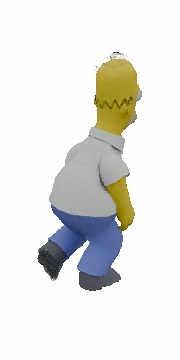} \\
\synlabel{GT} &
\includegraphics[width=\synimgw]{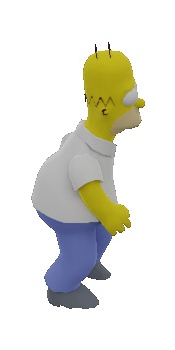} &
\includegraphics[width=\synimgw]{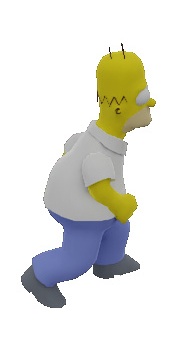} &
\includegraphics[width=\synimgw]{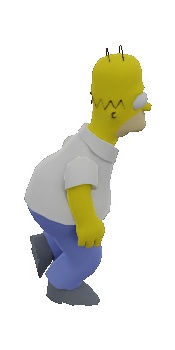} &
\includegraphics[width=\synimgw]{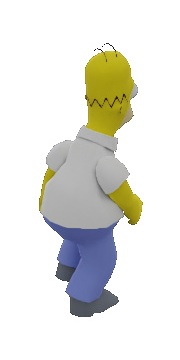} &
\includegraphics[width=\synimgw]{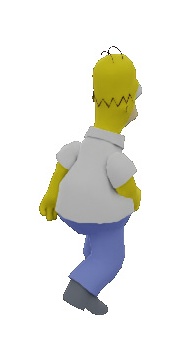} &
\includegraphics[width=\synimgw]{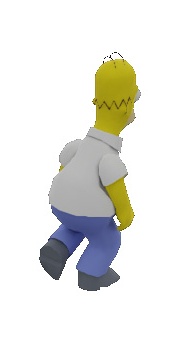} \\
\end{tabular}

\caption{\textbf{Qualitative comparison on synthetic Simpsons.} Rows are methods; columns are matched frames split into a supervised \emph{Training View} (left) and a held-out \emph{Novel View} (right). $t_0$ is the canonical (binding) frame; $t_1, t_2$ are progressively articulated frames.}
\label{fig:qualitative-synthetic-simpsons-supp}
\end{figure}

\begin{figure}[!htbp]
\centering
\def\synimgw{0.19\linewidth}  %
\def\synimgh{0.134\linewidth}
\footnotesize
\setlength{\tabcolsep}{0pt}
\renewcommand{\arraystretch}{0.4}

\begin{tabular}{@{}c@{\hspace{1pt}}ccc@{}}
\multicolumn{4}{@{}l}{\textbf{Training View}} \\
\cmidrule(lr){2-4}
& {\scriptsize $t_0$ (canonical)} & {\scriptsize $t_1$} & {\scriptsize $t_2$} \\
\synlabel{Puppeteer} &
\includegraphics[width=\synimgw]{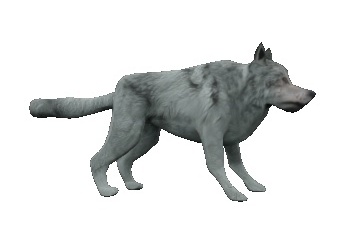} &
\includegraphics[width=\synimgw]{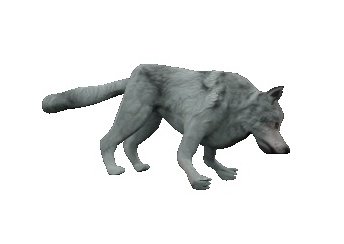} &
\includegraphics[width=\synimgw]{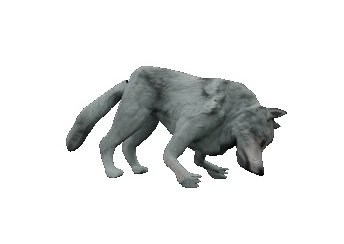} \\
\synlabel{Mani-GS} &
\includegraphics[width=\synimgw]{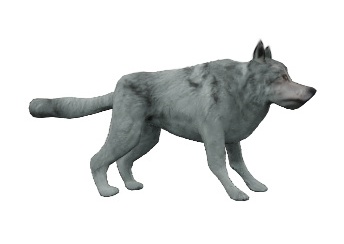} &
\includegraphics[width=\synimgw]{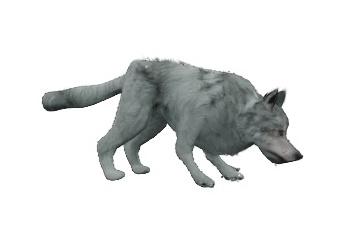} &
\includegraphics[width=\synimgw]{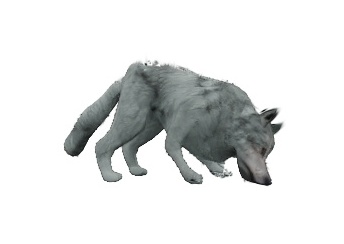} \\
\synlabel{RigGS} &
\includegraphics[width=\synimgw]{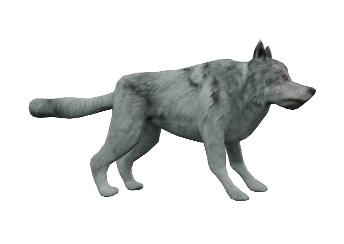} &
\includegraphics[width=\synimgw]{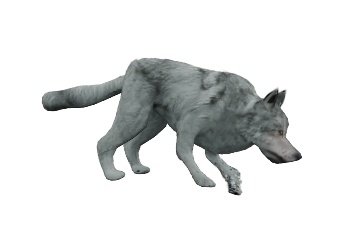} &
\includegraphics[width=\synimgw]{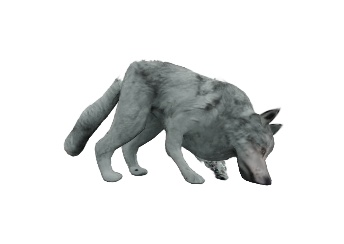} \\
\synlabel{Ours} &
\includegraphics[width=\synimgw]{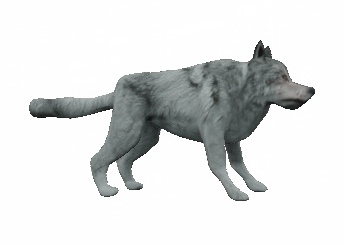} &
\includegraphics[width=\synimgw]{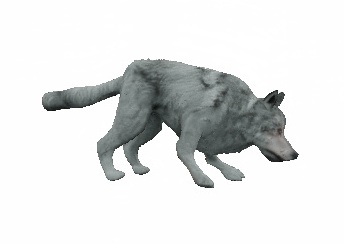} &
\includegraphics[width=\synimgw]{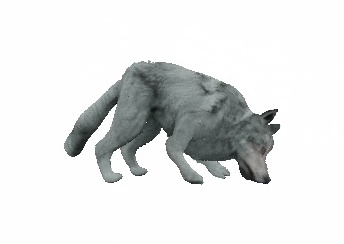} \\
\synlabel{GT} &
\includegraphics[width=\synimgw]{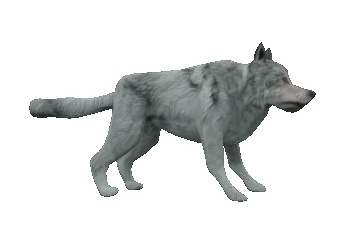} &
\includegraphics[width=\synimgw]{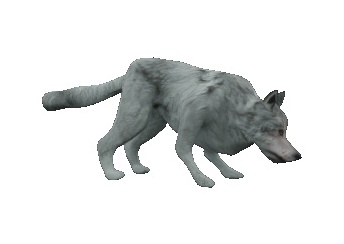} &
\includegraphics[width=\synimgw]{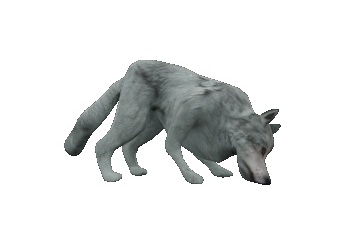} \\[6pt]
\multicolumn{4}{@{}l}{\textbf{Novel View}} \\
\cmidrule(lr){2-4}
& {\scriptsize $t_0$ (canonical)} & {\scriptsize $t_1$} & {\scriptsize $t_2$} \\
\synlabel{Puppeteer} &
\includegraphics[width=\synimgw]{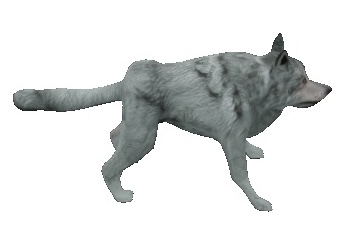} &
\includegraphics[width=\synimgw]{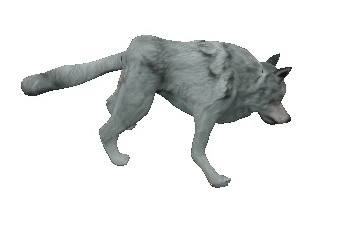} &
\includegraphics[width=\synimgw]{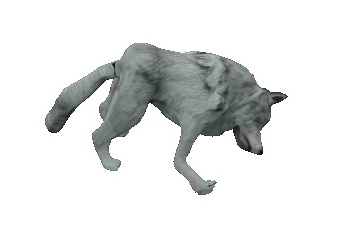} \\
\synlabel{Mani-GS} &
\includegraphics[width=\synimgw]{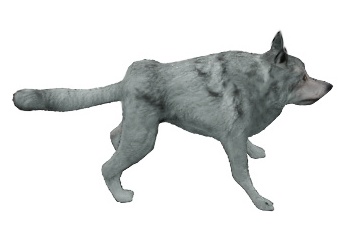} &
\includegraphics[width=\synimgw]{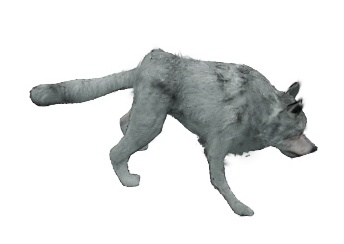} &
\includegraphics[width=\synimgw]{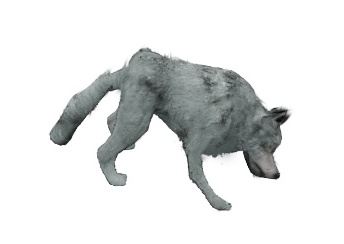} \\
\synlabel{RigGS} &
\includegraphics[width=\synimgw]{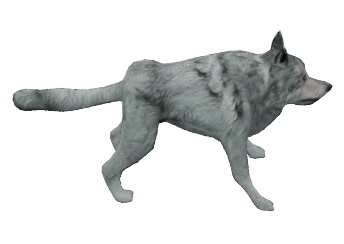} &
\includegraphics[width=\synimgw]{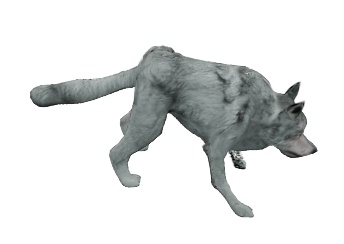} &
\includegraphics[width=\synimgw]{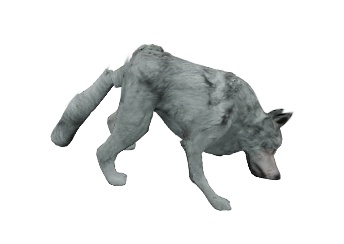} \\
\synlabel{Ours} &
\includegraphics[width=\synimgw]{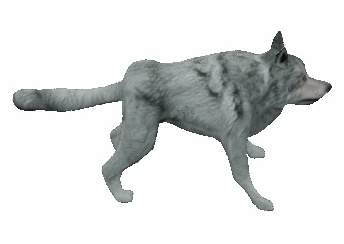} &
\includegraphics[width=\synimgw]{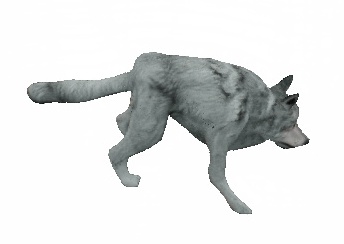} &
\includegraphics[width=\synimgw]{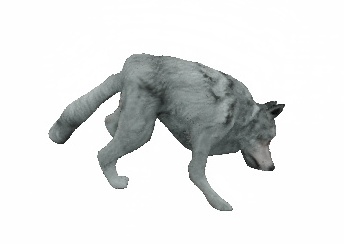} \\
\synlabel{GT} &
\includegraphics[width=\synimgw]{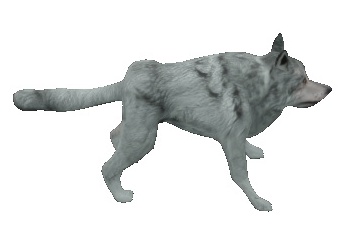} &
\includegraphics[width=\synimgw]{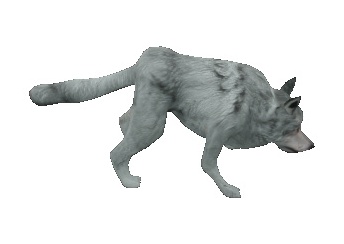} &
\includegraphics[width=\synimgw]{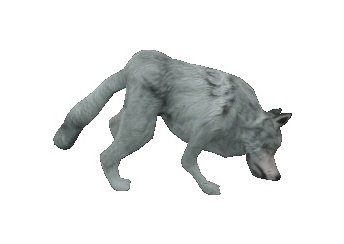} \\
\end{tabular}

\caption{\textbf{Qualitative comparison on synthetic wolf.} Training View is on top and Novel View is on the bottom. Rows are methods; columns are matched frames, with $t_0$ the canonical (binding) frame and $t_1,\,t_2$ progressively articulated.}
\label{fig:qualitative-synthetic-wolf-supp}
\end{figure}

\begin{figure}[!htbp]
\centering
\def\synimgw{0.22\linewidth}
\def\synimgh{0.22\linewidth}
\footnotesize
\setlength{\tabcolsep}{0pt}
\renewcommand{\arraystretch}{0.4}

\begin{tabular}{@{}c@{\hspace{1pt}}cc@{\hspace{\synblockgap}}cc@{}}
& \multicolumn{2}{c}{\textbf{Training View}} & \multicolumn{2}{c}{\textbf{Novel View}} \\
\cmidrule(lr){2-3}\cmidrule(lr){4-5}
& {\scriptsize $t_0$ (canonical)} & {\scriptsize $t_1$}
& {\scriptsize $t_0$ (canonical)} & {\scriptsize $t_1$} \\
\synlabel{Puppeteer} &
\includegraphics[width=\synimgw]{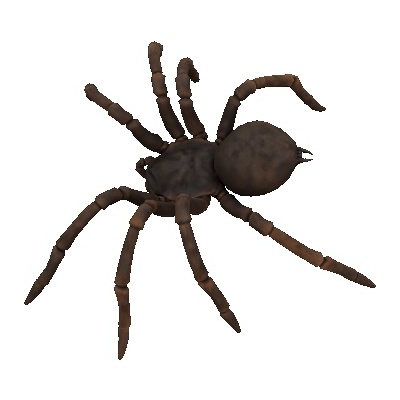} &
\includegraphics[width=\synimgw]{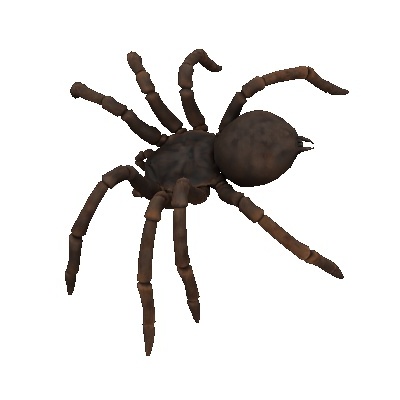} &
\includegraphics[width=\synimgw]{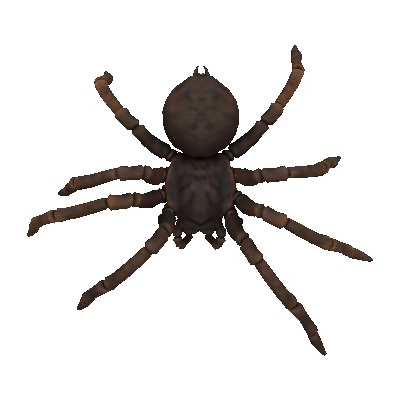} &
\includegraphics[width=\synimgw]{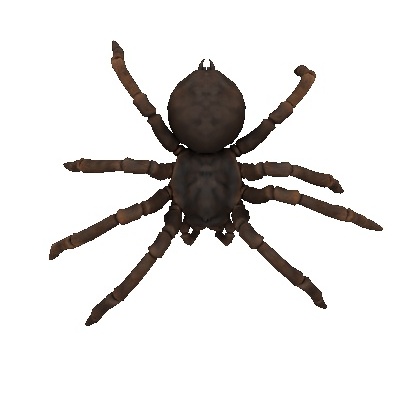} \\
\synlabel{Mani-GS} &
\includegraphics[width=\synimgw]{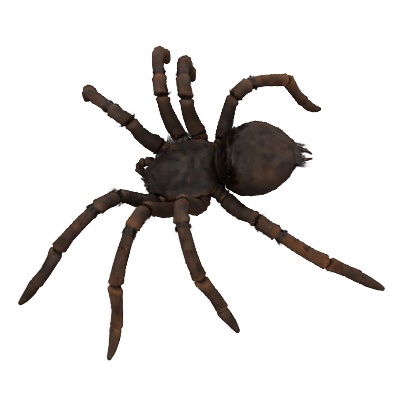} &
\includegraphics[width=\synimgw]{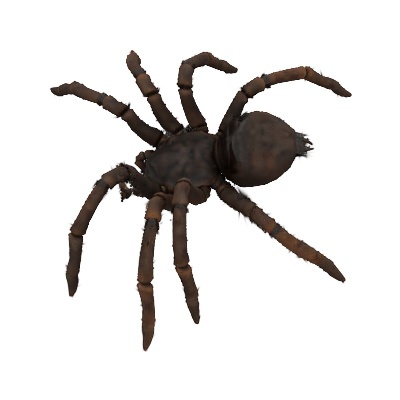} &
\includegraphics[width=\synimgw]{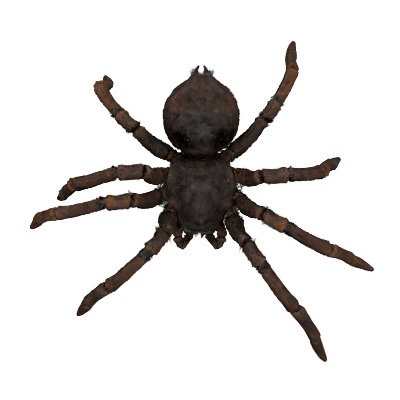} &
\includegraphics[width=\synimgw]{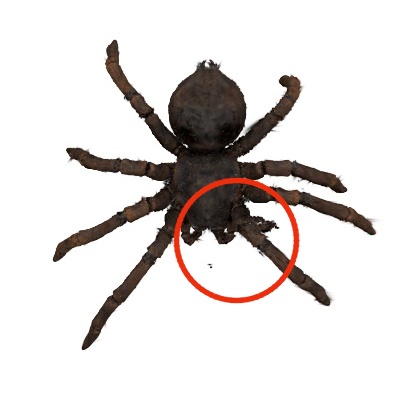} \\
\synlabel{RigGS} &
\includegraphics[width=\synimgw]{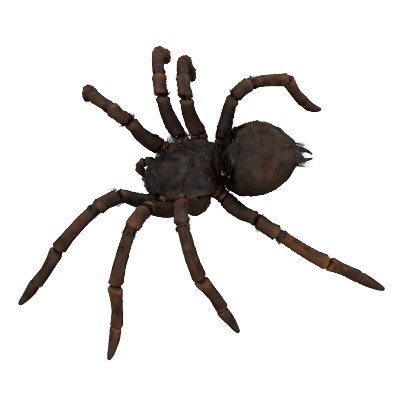} &
\includegraphics[width=\synimgw]{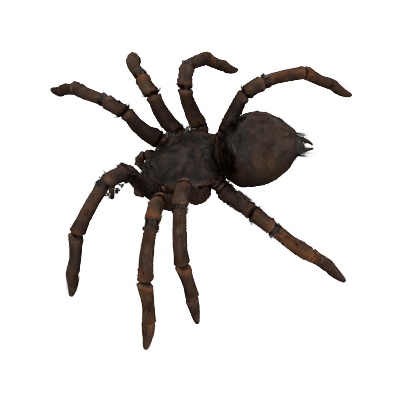} &
\includegraphics[width=\synimgw]{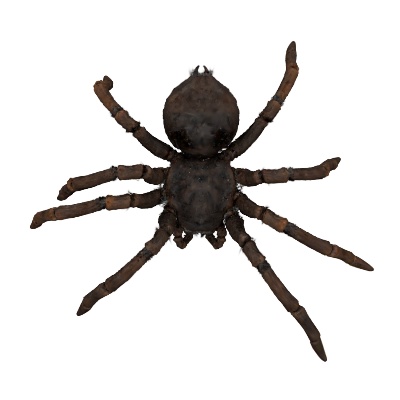} &
\includegraphics[width=\synimgw]{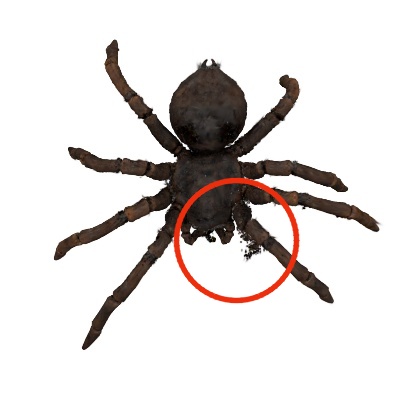} \\
\synlabel{Ours} &
\includegraphics[width=\synimgw]{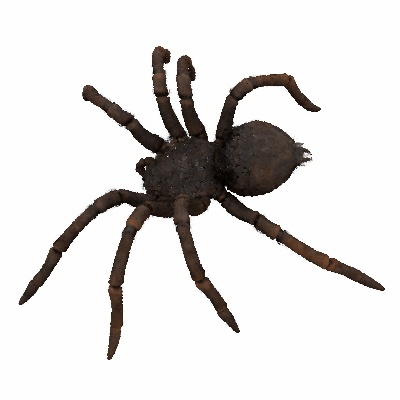} &
\includegraphics[width=\synimgw]{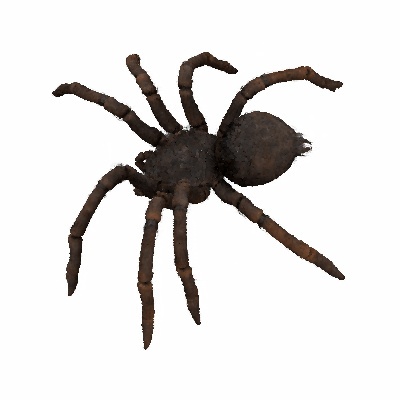} &
\includegraphics[width=\synimgw]{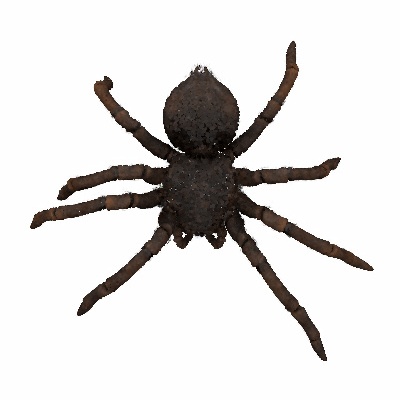} &
\includegraphics[width=\synimgw]{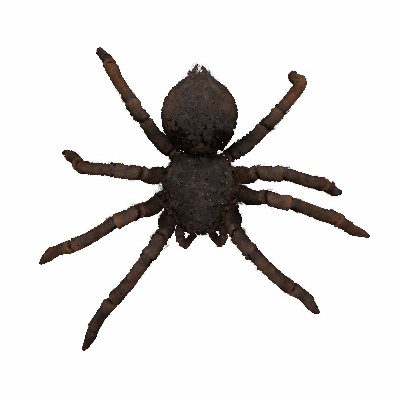} \\
\synlabel{GT} &
\includegraphics[width=\synimgw]{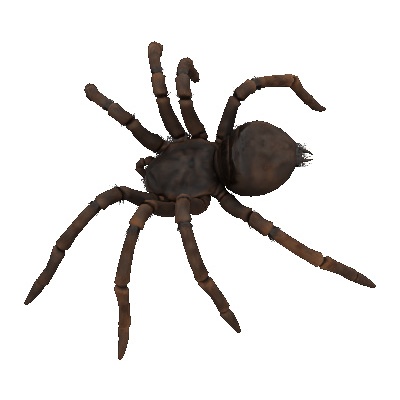} &
\includegraphics[width=\synimgw]{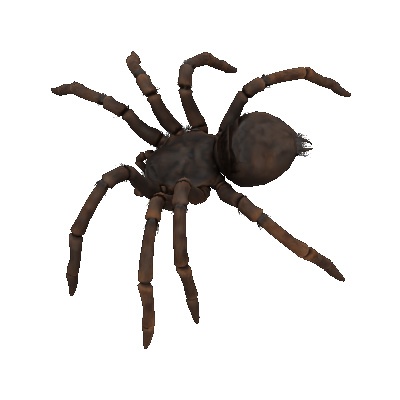} &
\includegraphics[width=\synimgw]{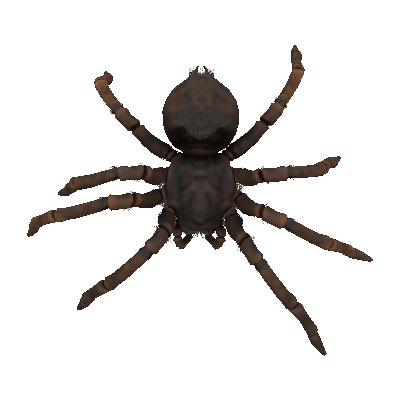} &
\includegraphics[width=\synimgw]{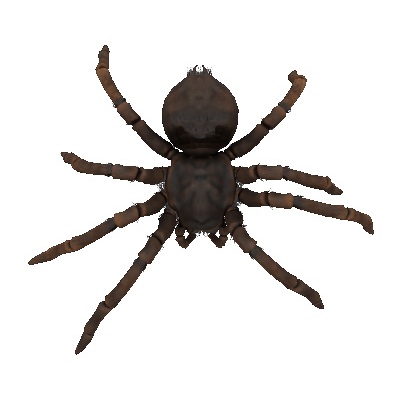} \\
\end{tabular}

\caption{\textbf{Qualitative comparison on synthetic spider.} Rows are methods; columns are matched frames split into a supervised \emph{Training View} (left) and a held-out \emph{Novel View} (right). $t_0$ is the canonical (binding) frame and $t_1$ is the articulated frame shown; the additional $t_2$ is omitted to allow larger images.}
\label{fig:qualitative-synthetic-spider-supp}
\end{figure}

Figures~\ref{fig:qualitative-synthetic-fox-supp}--\ref{fig:qualitative-synthetic-spider-supp} extend the synthetic qualitative comparison of Section~\ref{sec:qualitative} (Figure~\ref{fig:qualitative-synthetic}) to the four remaining synthetic subjects. Rows are methods (Puppeteer~\citep{song2025puppeteer}, Mani-GS~\citep{manigs}, RigGS~\citep{yao2025riggs}, ours, ground truth) and columns are matched frames; each figure shows the supervised \emph{Training View} and a held-out \emph{Novel View} with $t_0$ the canonical (binding) frame. Spider is shown with two frames per view rather than three so each image stays large enough to read the limb-junction detail. The same failure modes observed in the main figure recur across morphologies. Mani-GS produces spiky artifacts, especially around joint boundaries: the neck and back for fox, the arm and leg for Simpsons, the chest and tail for wolf, and the legs for spider. RigGS shows similar artifacts at the joints: the feet for Simpsons, the tail and front leg for wolf, and the front leg next to the jaws for spider. Puppeteer fails to deform the geometry to match the observed motion sequence accurately. While its result is sometimes passable, it is not consistently accurate, with limbs and body parts often not reaching their target positions and the geometry sometimes structurally incorrect. Our method recovers the joint rotations more accurately throughout the motion sequence and renders cleanly without artifacts. For parts of the subject that remain occluded under the fixed training view, however, the geometry does not always track as accurately: the 2D priors we use to disambiguate single-view supervision provide no signal in occluded regions either (e.g., the left arm of Simpsons).

\begin{figure}[!htbp]
\centering
\def\zjusimgh{0.168\linewidth}
\def\zjuslabelw{0.05\linewidth}
\footnotesize
\setlength{\tabcolsep}{0pt}
\renewcommand{\arraystretch}{0.4}
\newcommand{\zjuslabel}[1]{\parbox[b][\zjusimgh][c]{\zjuslabelw}{\centering\scriptsize #1}}
\begin{tabular}{@{}c@{\hspace{1pt}}cccc@{\hspace{4pt}}cccc@{}}
& \multicolumn{4}{c}{\textbf{Training View}} & \multicolumn{4}{c}{\textbf{Novel View}} \\
\cmidrule(lr){2-5}\cmidrule(lr){6-9}
& {\scriptsize Mani-GS} & {\scriptsize RigGS} & {\scriptsize Ours} & {\scriptsize GT}
& {\scriptsize Mani-GS} & {\scriptsize RigGS} & {\scriptsize Ours} & {\scriptsize GT} \\
\zjuslabel{$t_0$} &
\includegraphics[height=\zjusimgh]{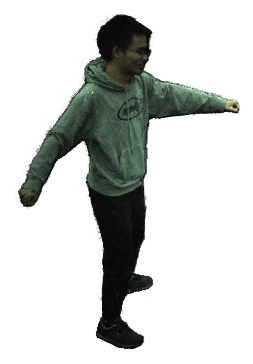} & \includegraphics[height=\zjusimgh]{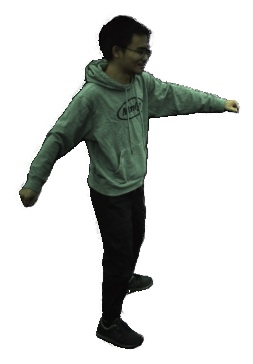} & \includegraphics[height=\zjusimgh]{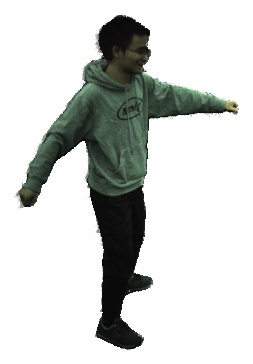} & \includegraphics[height=\zjusimgh]{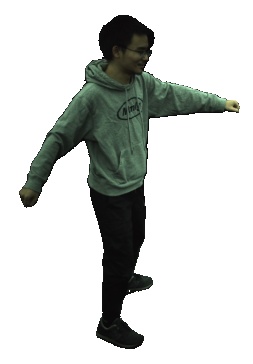} &
\includegraphics[height=\zjusimgh]{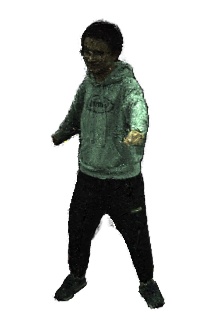} & \includegraphics[height=\zjusimgh]{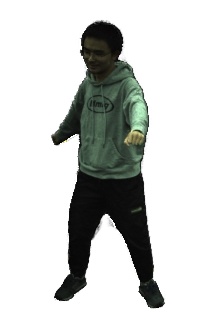} & \includegraphics[height=\zjusimgh]{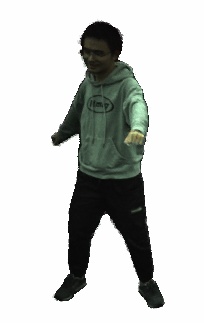} & \includegraphics[height=\zjusimgh]{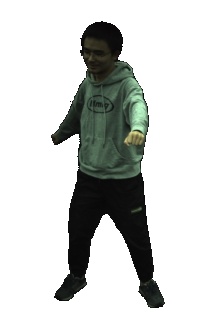} \\
\zjuslabel{$t_1$} &
\includegraphics[height=\zjusimgh]{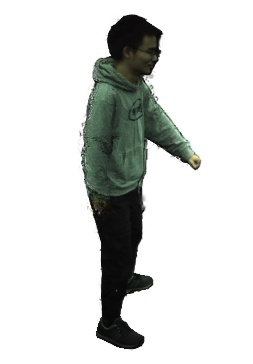} & \includegraphics[height=\zjusimgh]{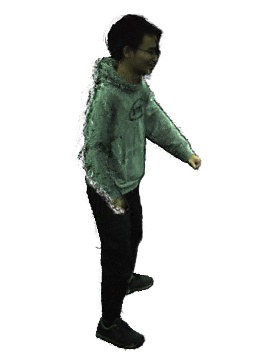} & \includegraphics[height=\zjusimgh]{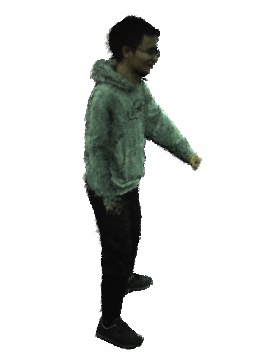} & \includegraphics[height=\zjusimgh]{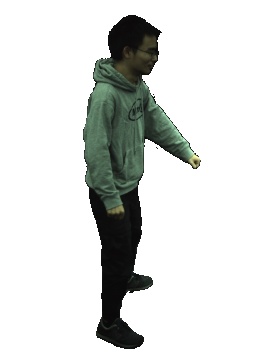} &
\includegraphics[height=\zjusimgh]{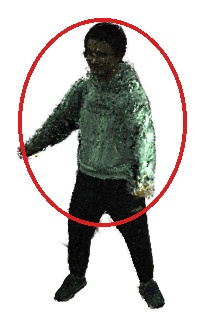} & \includegraphics[height=\zjusimgh]{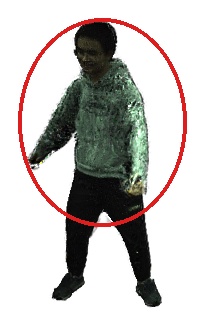} & \includegraphics[height=\zjusimgh]{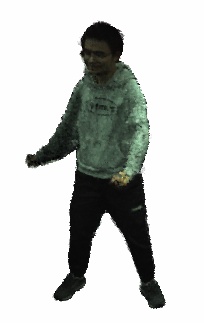} & \includegraphics[height=\zjusimgh]{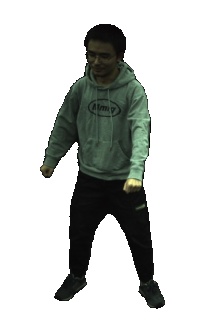} \\
\end{tabular}

\caption{\textbf{Qualitative comparison on the second ZJU-MoCap subject.} Layout as in Figure~\ref{fig:qualitative-realworld}: rows are frames; columns are split into the supervised \emph{Training View} (left) and a held-out \emph{Novel View} (right), each showing Mani-GS~\citep{manigs}, RigGS~\citep{yao2025riggs}, ours and the ground truth. $t_0$ is the canonical (binding) frame; $t_1$ is an articulated frame. In the novel view, our renderings are cleaner, without the spikes or gaps of the baselines (red); best viewed zoomed in.}
\label{fig:qualitative-zju390-supp}
\end{figure}

Figure~\ref{fig:qualitative-zju390-supp} shows the comparison on the second ZJU-MoCap subject. At the held-out view, both Mani-GS and RigGS show artifacts on the arms, while ours remains clean.

\subsection{Captured Scenes} \label{sec:supp-realworld}
\begin{figure}[!htbp]
\centering
\def\capimgw{0.18\linewidth}
\def\capimgh{0.264\linewidth} %
\def\caplabelw{0.08\linewidth}
\def\capblockgap{4pt}
\footnotesize
\setlength{\tabcolsep}{0pt}
\renewcommand{\arraystretch}{0.4}
\providecommand{\capturedlab}{}\renewcommand{\capturedlab}[1]{\parbox[b][\capimgh][c]{\caplabelw}{\centering\scriptsize #1}}

\begin{tabular}{@{}c@{\hspace{1pt}}cc@{\hspace{\capblockgap}}c@{}}
& \multicolumn{2}{c}{\textbf{Training View}} & \textbf{Novel View} \\
\cmidrule(lr){2-3}\cmidrule(lr){4-4}
& {\scriptsize Ours} & {\scriptsize Driving video} & {\scriptsize Ours} \\
\capturedlab{$t_0$ (canonical)} &
\includegraphics[width=\capimgw]{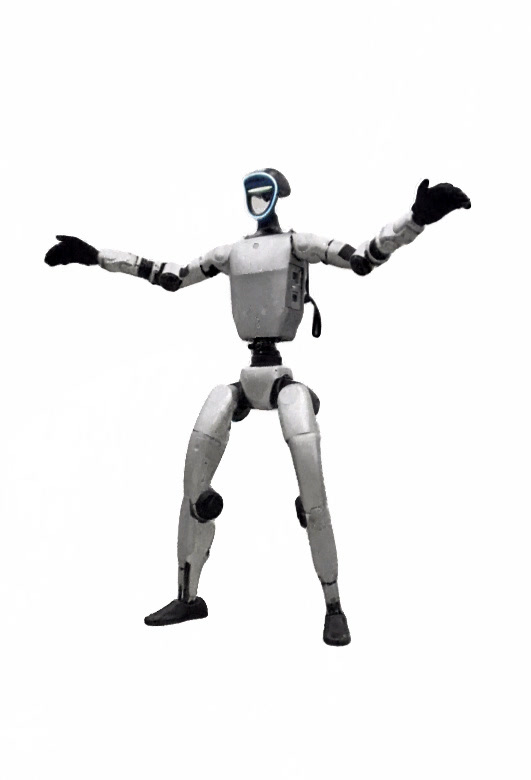} &
\includegraphics[width=\capimgw]{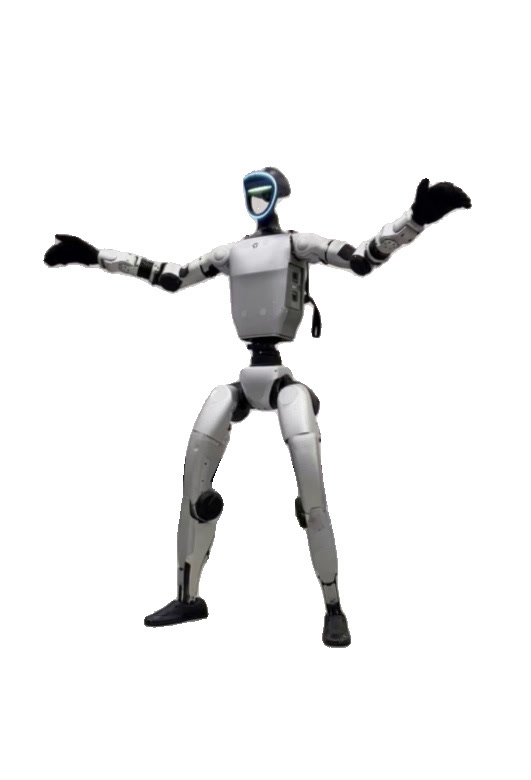} &
\includegraphics[width=\capimgw]{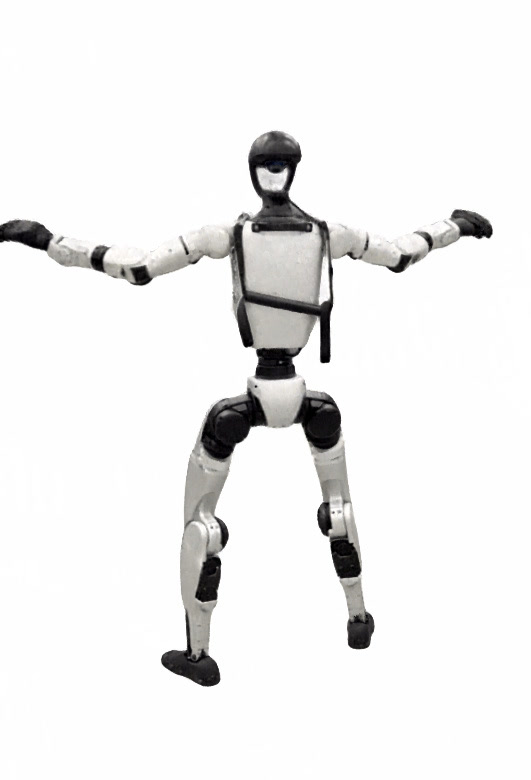} \\
\capturedlab{$t_1$} &
\includegraphics[width=\capimgw]{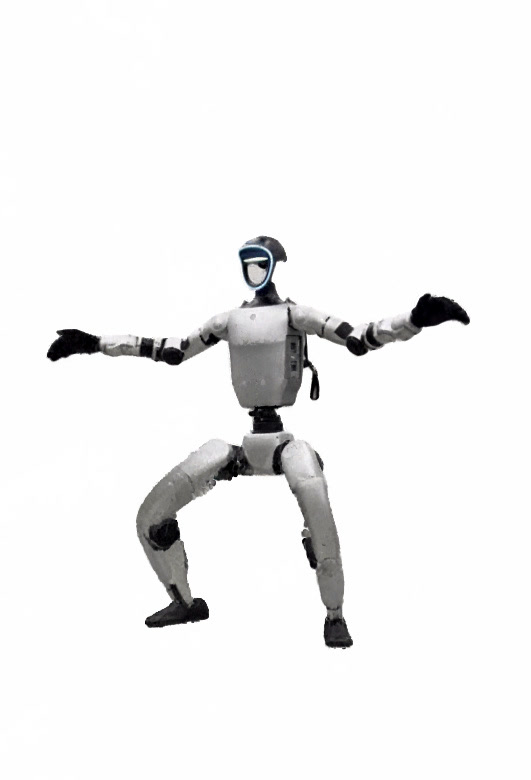} &
\includegraphics[width=\capimgw]{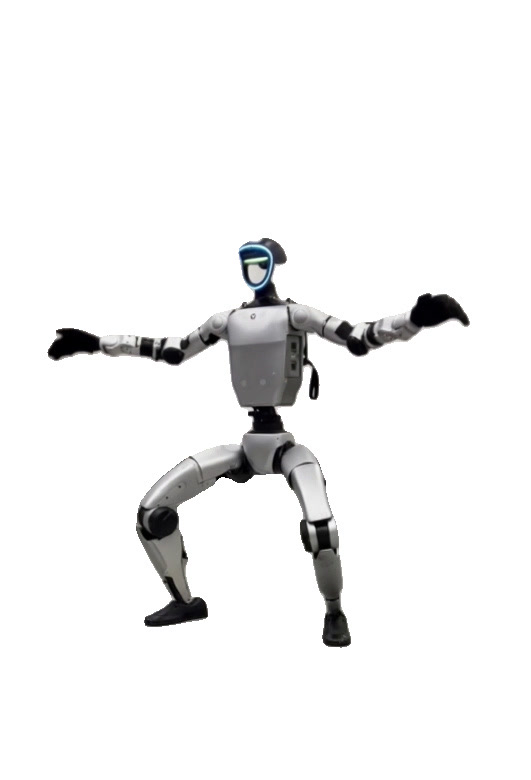} &
\includegraphics[width=\capimgw]{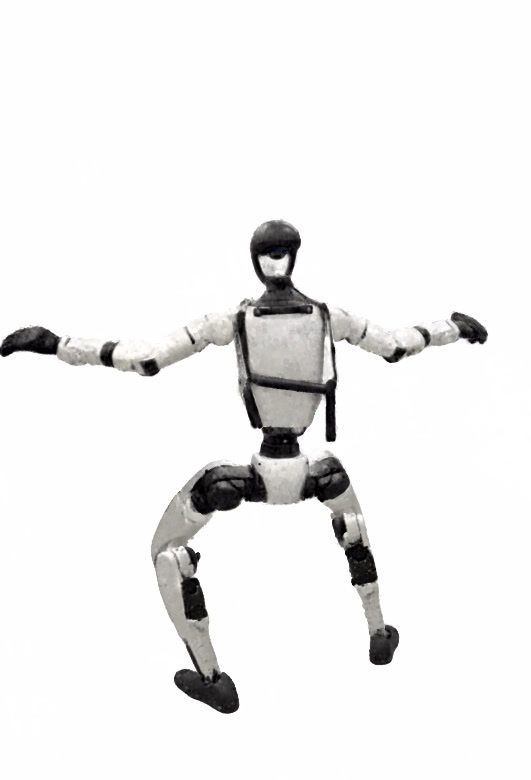} \\
\capturedlab{$t_2$} &
\includegraphics[width=\capimgw]{robot/train_ours_phase2_30.jpg} &
\includegraphics[width=\capimgw]{robot/train_gt_30.jpg} &
\includegraphics[width=\capimgw]{robot/ours_phase2_view10_30.jpg} \\
\end{tabular}

\caption{\textbf{Our results on the captured robot.} Rows are matched frames ($t_0$ is the canonical/binding frame; $t_1, t_2$ are progressively articulated); columns are split into the supervised \emph{Training View} (ours and the generated driving video) and a held-out \emph{Novel View} (ours).}
\label{fig:qualitative-realworld-robot-supp}
\end{figure}

\begin{figure}[!htbp]
\centering
\def\capimgw{0.18\linewidth}
\def\capimgh{0.320\linewidth} %
\def\caplabelw{0.08\linewidth}
\def\capblockgap{4pt}
\footnotesize
\setlength{\tabcolsep}{0pt}
\renewcommand{\arraystretch}{0.4}
\providecommand{\capturedlab}{}\renewcommand{\capturedlab}[1]{\parbox[b][\capimgh][c]{\caplabelw}{\centering\scriptsize #1}}

\begin{tabular}{@{}c@{\hspace{1pt}}cc@{\hspace{\capblockgap}}c@{}}
& \multicolumn{2}{c}{\textbf{Training View}} & \textbf{Novel View} \\
\cmidrule(lr){2-3}\cmidrule(lr){4-4}
& {\scriptsize Ours} & {\scriptsize Driving video} & {\scriptsize Ours} \\
\capturedlab{$t_0$ (canonical)} &
\includegraphics[width=\capimgw]{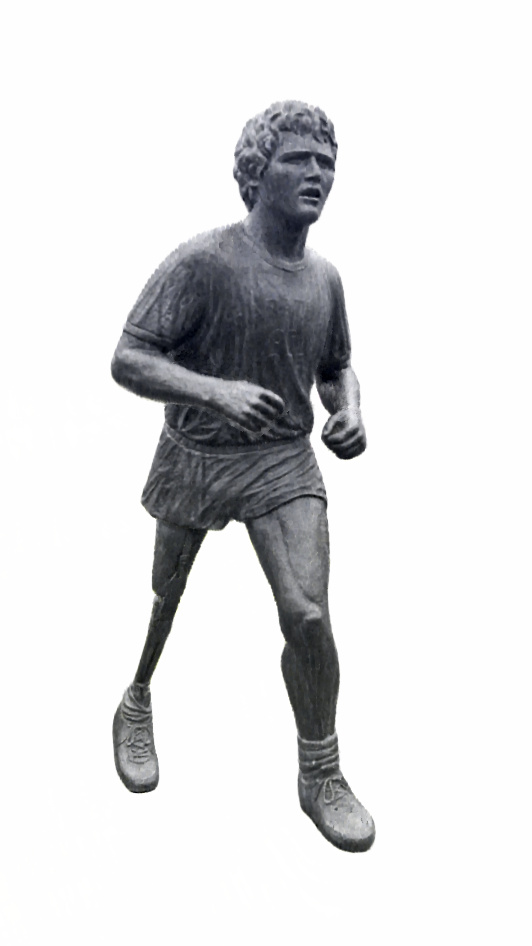} &
\includegraphics[width=\capimgw]{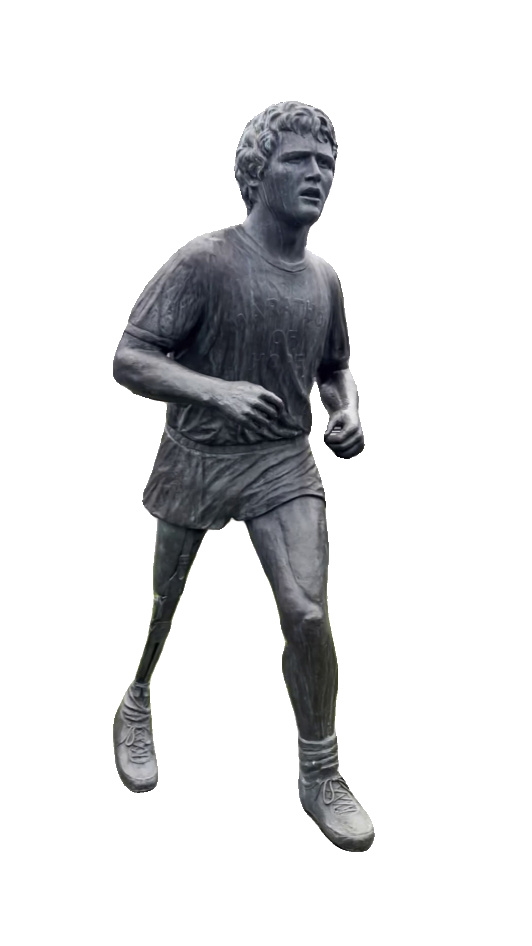} &
\includegraphics[width=\capimgw]{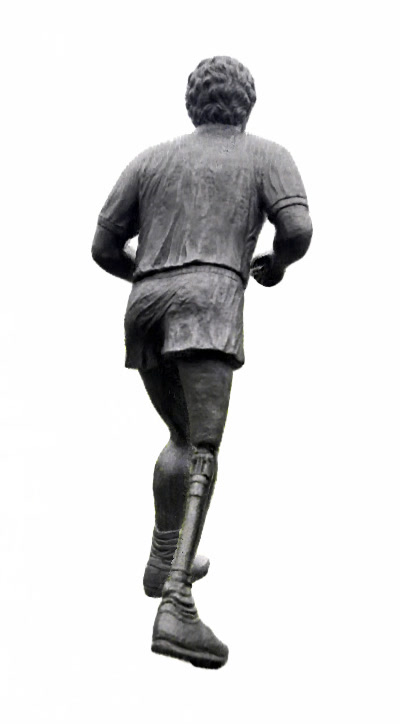} \\
\capturedlab{$t_1$} &
\includegraphics[width=\capimgw]{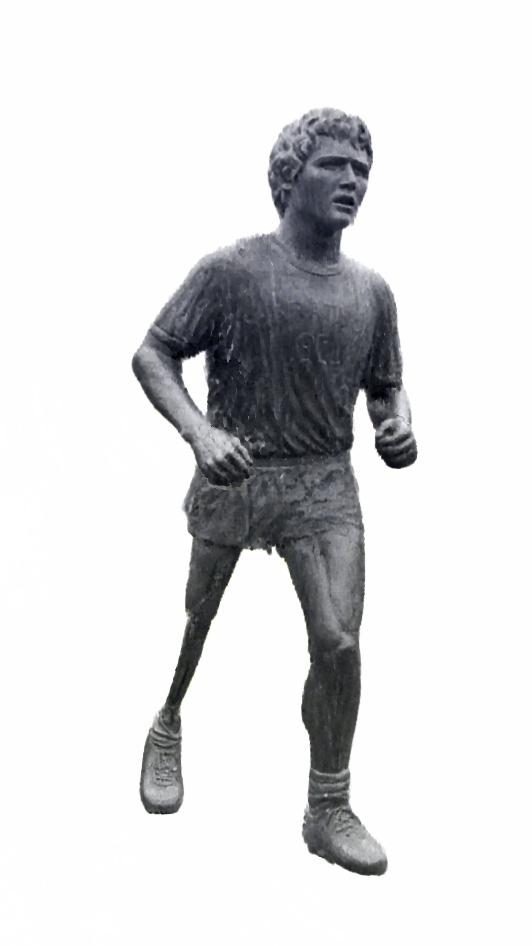} &
\includegraphics[width=\capimgw]{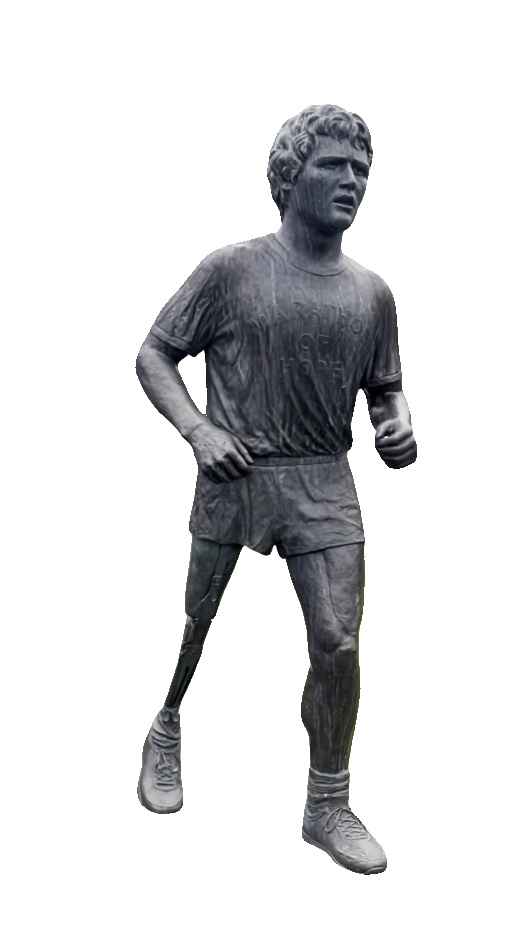} &
\includegraphics[width=\capimgw]{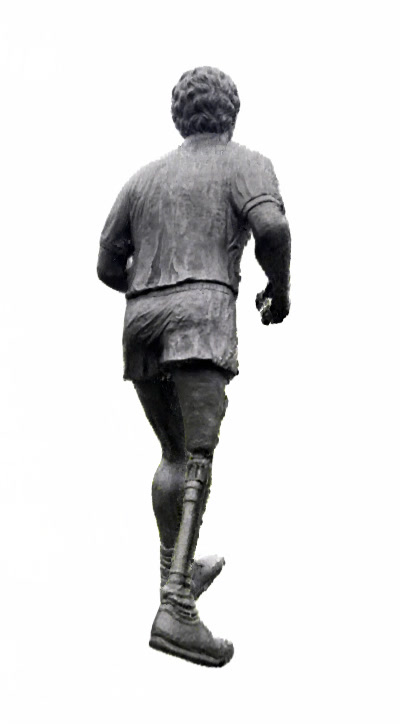} \\
\capturedlab{$t_2$} &
\includegraphics[width=\capimgw]{statue/train_ours_phase2_36.jpg} &
\includegraphics[width=\capimgw]{statue/train_gt_36.jpg} &
\includegraphics[width=\capimgw]{statue/ours_phase2_view3_36.jpg} \\
\end{tabular}

\caption{\textbf{Our results on the captured statue.} Layout as in Figure~\ref{fig:qualitative-realworld-robot-supp}.}
\label{fig:qualitative-realworld-supp}
\end{figure}

This section adds frames to the captured scenes of Figure~\ref{fig:captured-main}. We photograph a static subject (a robot and a statue) from multiple views, pre-train a PAPR cloud on the photos, generate a driving video at one canonical view with Kling~3.0~\citep{Chen2025KlingOmniTR}, and animate the cloud from that video. Because the driving video is generated, no ground truth exists at novel views, so we show qualitative results only. Figures~\ref{fig:qualitative-realworld-robot-supp} and~\ref{fig:qualitative-realworld-supp} show our results at the training view beside the driving video, and from a held-out viewpoint.

\begin{figure}[!htbp]
\centering
\def\mtsimgw{0.18\linewidth}
\footnotesize
\setlength{\tabcolsep}{0pt}
\renewcommand{\arraystretch}{0.4}
\begin{tabular}{@{}c@{\hspace{4pt}}cc@{\hspace{4pt}}cc@{}}
& \multicolumn{2}{c}{\textbf{Sequence 1}} & \multicolumn{2}{c}{\textbf{Sequence 2}} \\
\cmidrule(lr){2-3}\cmidrule(lr){4-5}
{\scriptsize $t_0$ (canonical)} & {\scriptsize $t_1$} & {\scriptsize $t_2$} & {\scriptsize $t_1$} & {\scriptsize $t_2$} \\
\includegraphics[width=\mtsimgw]{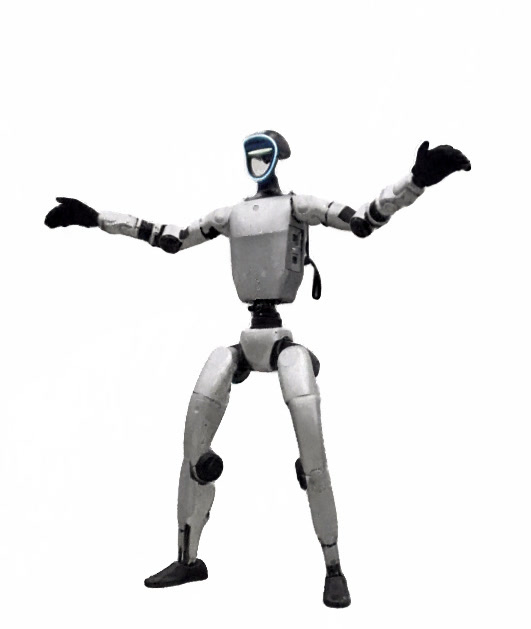} &
\includegraphics[width=\mtsimgw]{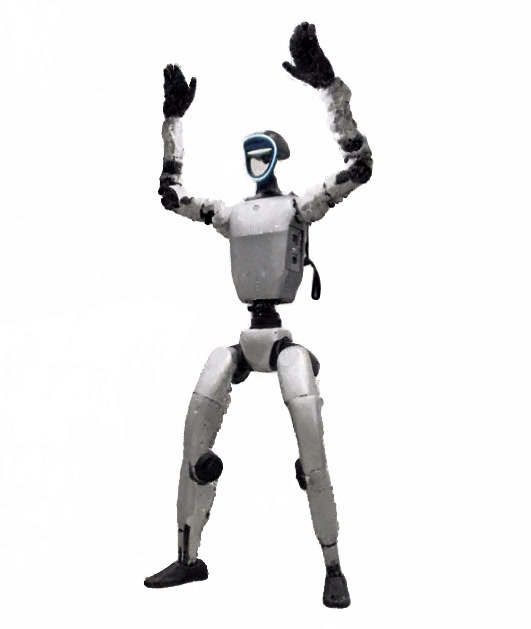} &
\includegraphics[width=\mtsimgw]{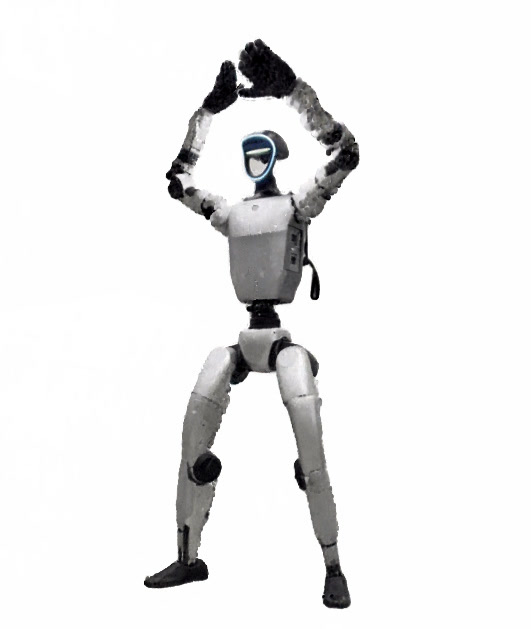} &
\includegraphics[width=\mtsimgw]{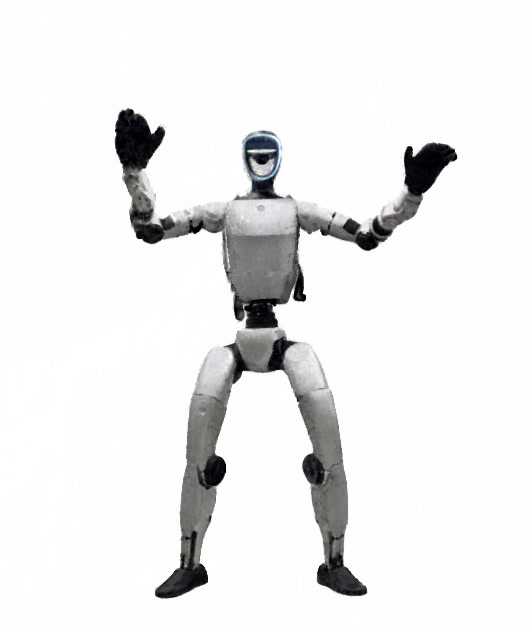} &
\includegraphics[width=\mtsimgw]{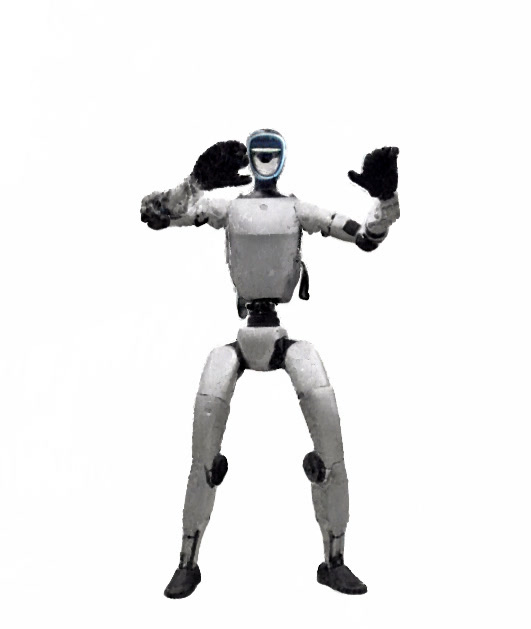} \\
\end{tabular}
\caption{\textbf{Novel pose driving on the captured robot.} Our model is driven at test time by two pose sequences not seen during fitting. Columns show the canonical (binding) frame $t_0$ followed by two frames from each driving sequence. No ground truth is available for the driven poses.}
\label{fig:supp-robot-novel-pose}
\end{figure}

We further drive the captured robot with two pose sequences unseen in the driving video (Figure~\ref{fig:supp-robot-novel-pose}). The model follows both sequences, including arm poses far from any in the driving video.

\clearpage
\section{Additional Ablations} \label{sec:supp-ablations}

This section reports two additional ablations. Section~\ref{sec:supp-phase-ablation} compares Phase~1 alone with the full two-phase optimization for every method, and examines what the two phases and the depth supervision each contribute to our method on T-Rex. Section~\ref{sec:gt-mesh-ablation} isolates the rendering representation from any geometry or skinning quality by deforming every method with the ground truth.

\subsection{Two-Phase Optimization Ablation} \label{sec:supp-phase-ablation}

Table~\ref{tab:multiphase} compares Phase~1 alone with the full two-phase optimization on all synthetic scenes. Phase~2 raises training-view PSNR for every method ($+1.8$~dB on Mani-GS, $+2.7$~dB on RigGS, $+1.0$~dB on ours), while novel-view PSNR stays within $0.4$~dB of Phase~1. For our method this gain comes with visible geometric corrections: Phase~2 fixes per-frame joint rotations that Phase~1 gets wrong, which we examine on T-Rex below.

\begin{table}[h]
\centering
\caption{Two-phase ablation on the synthetic scenes. \emph{Phase~1} stops after each method's first training phase; \emph{Full} adds Phase~2 (Section~\ref{sec:multi-phase}) on top. \textbf{Best} in bold, \underline{second-best} underlined.}
\label{tab:multiphase}
\footnotesize
\begin{tabular}{l|ccc|ccc}
\toprule
 & \multicolumn{3}{c|}{Training view} & \multicolumn{3}{c}{Novel view} \\
Configuration & PSNR$\uparrow$ & SSIM$\uparrow$ & LPIPS$\downarrow$
              & PSNR$\uparrow$ & SSIM$\uparrow$ & LPIPS$\downarrow$ \\
\midrule
Mani-GS (Phase~1)                   & 26.71 & 0.962 & 0.037 & 20.14 & 0.925 & 0.073 \\
Mani-GS (Full)                      & 28.53 & \underline{0.974} & \textbf{0.026} & 20.51 & 0.927 & 0.072 \\
RigGS (Phase~1)                     & 26.80 & 0.963 & 0.036 & 20.19 & 0.924 & 0.071 \\
RigGS (Full)                        & \textbf{29.52} & \textbf{0.975} & 0.029 & 20.37 & 0.926 & 0.069 \\
Ours (Phase~1)                      & 28.02 & 0.964 & 0.029 & \textbf{23.70} & \textbf{0.946} & \textbf{0.044} \\
Ours (Full)                         & \underline{29.04} & 0.971 & \underline{0.027} & \underline{23.51} & \underline{0.944} & \underline{0.045} \\
\bottomrule
\end{tabular}
\end{table}

\begin{figure}[!htbp]
\centering
\def\synimgw{0.24\linewidth}
\def\synimgh{0.15\linewidth}
\footnotesize
\setlength{\tabcolsep}{0pt}
\renewcommand{\arraystretch}{0.4}

\begin{tabular}{@{}c@{\hspace{1pt}}ccc@{}}
\multicolumn{4}{@{}l}{\textbf{Training View}} \\
\cmidrule(lr){2-4}
& {\scriptsize $t_0$ (canonical)} & {\scriptsize $t_1$} & {\scriptsize $t_2$} \\
\synlabel{Phase~1 w/o $\mathcal{L}_{\Delta\text{depth}}$} &
\includegraphics[width=\synimgw]{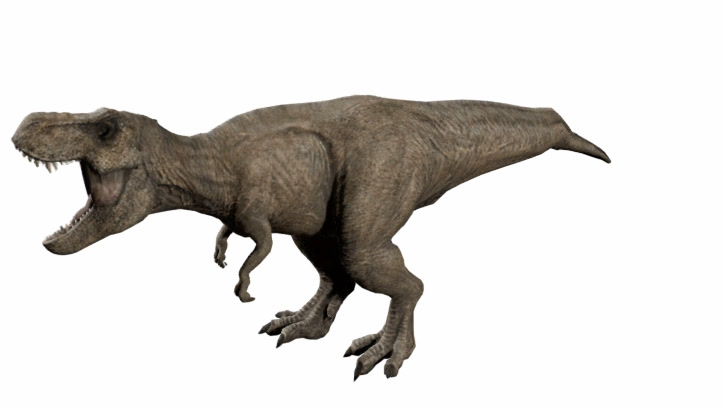} &
\includegraphics[width=\synimgw]{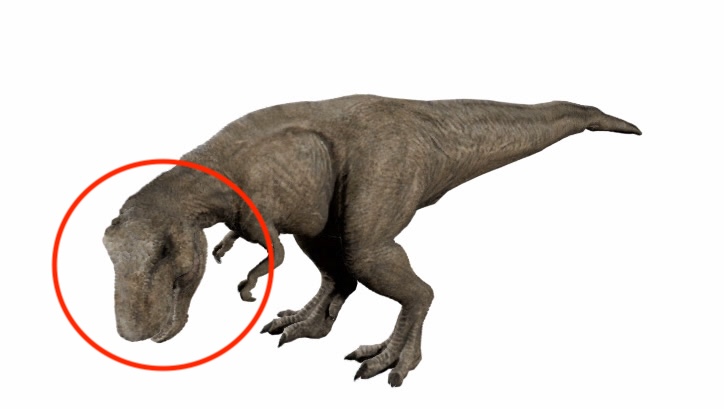} &
\includegraphics[width=\synimgw]{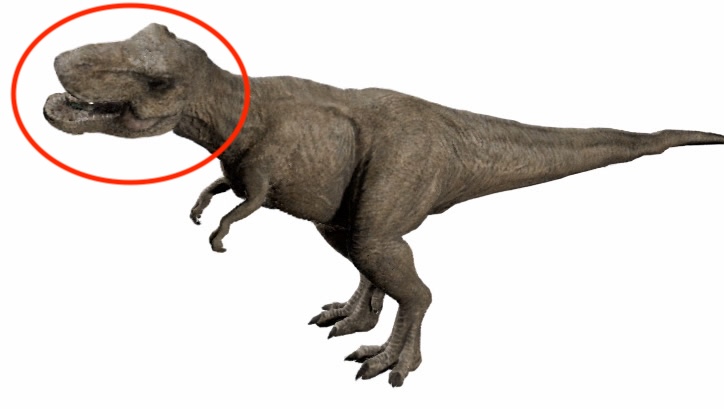} \\
\synlabel{Phase~1} &
\includegraphics[width=\synimgw]{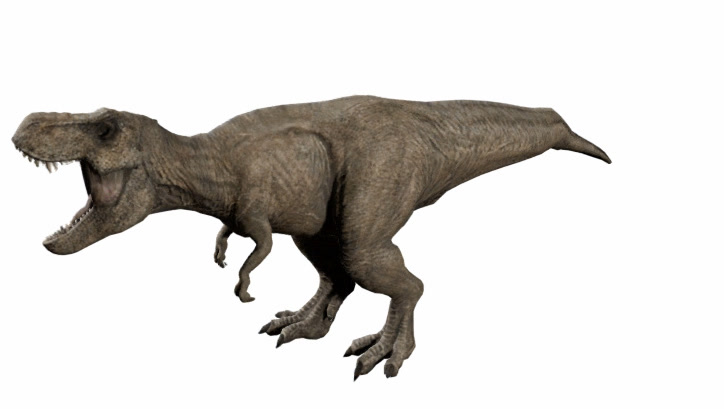} &
\includegraphics[width=\synimgw]{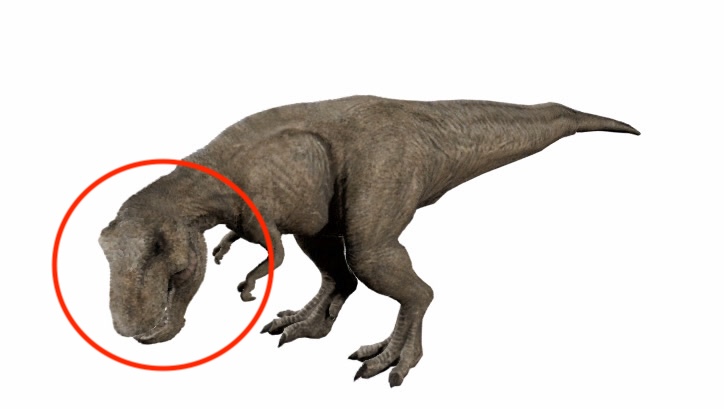} &
\includegraphics[width=\synimgw]{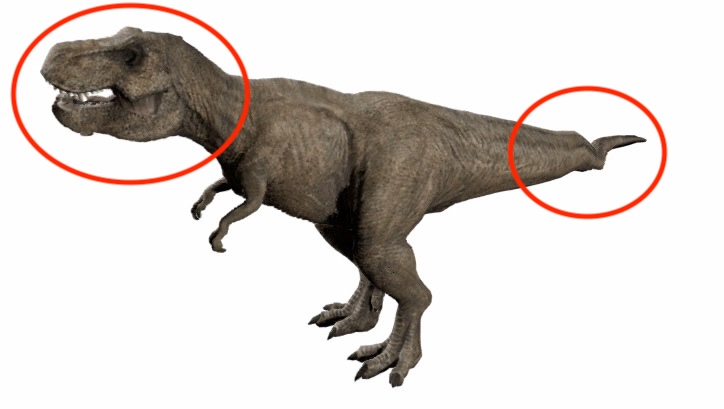} \\
\synlabel{Full} &
\includegraphics[width=\synimgw]{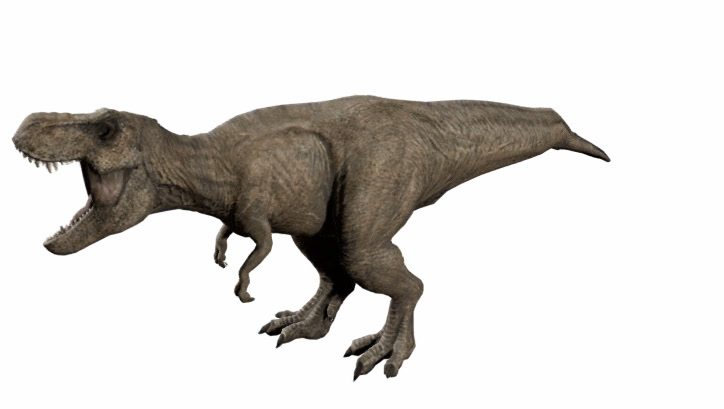} &
\includegraphics[width=\synimgw]{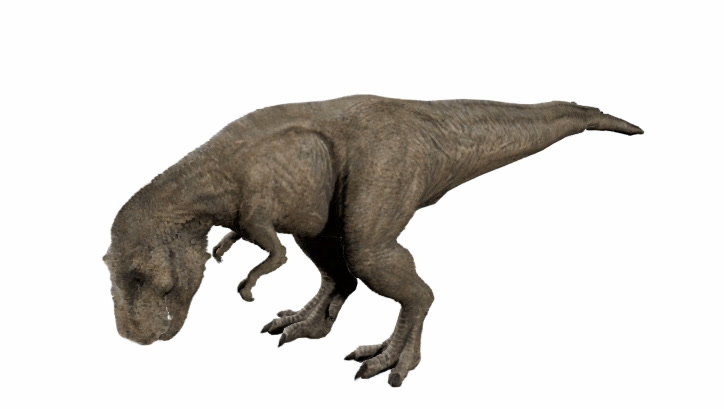} &
\includegraphics[width=\synimgw]{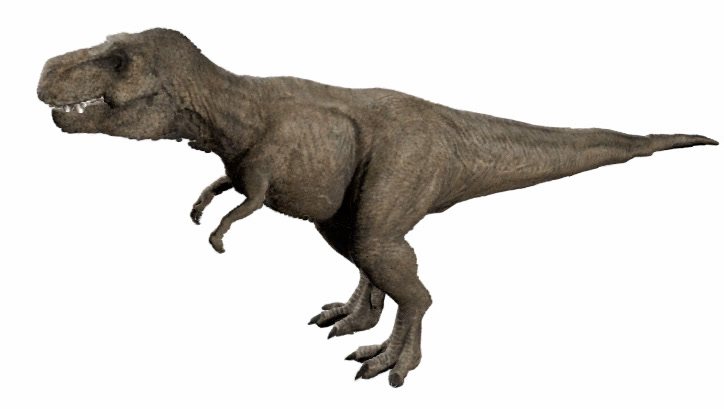} \\
\synlabel{GT} &
\includegraphics[width=\synimgw]{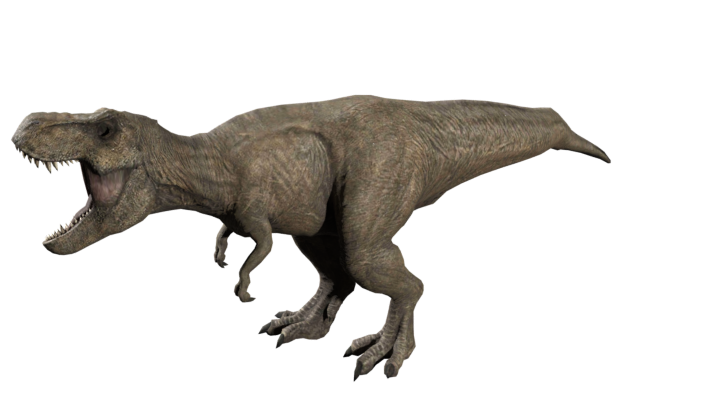} &
\includegraphics[width=\synimgw]{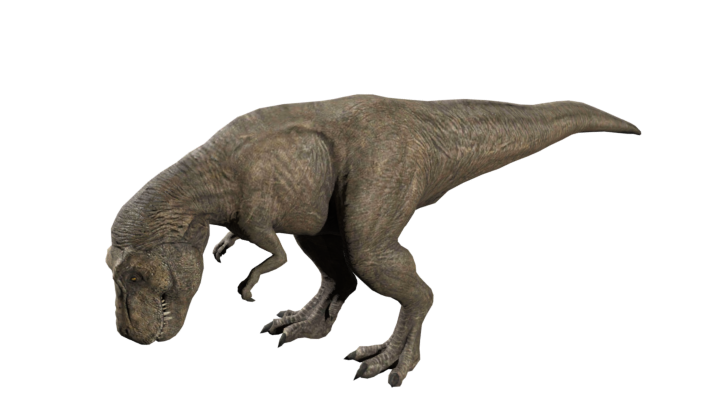} &
\includegraphics[width=\synimgw]{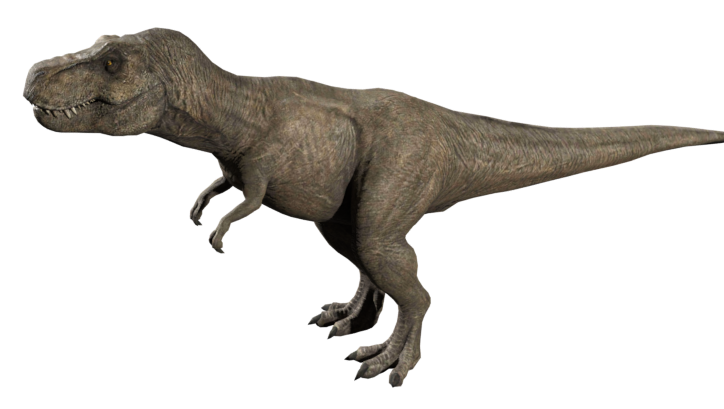} \\[6pt]
\multicolumn{4}{@{}l}{\textbf{Novel View}} \\
\cmidrule(lr){2-4}
& {\scriptsize $t_0$ (canonical)} & {\scriptsize $t_1$} & {\scriptsize $t_2$} \\
\synlabel{Phase~1 w/o $\mathcal{L}_{\Delta\text{depth}}$} &
\includegraphics[width=\synimgw]{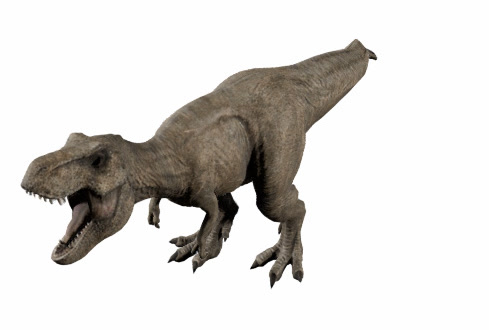} &
\includegraphics[width=\synimgw]{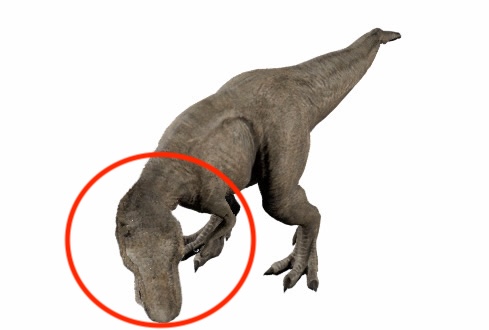} &
\includegraphics[width=\synimgw]{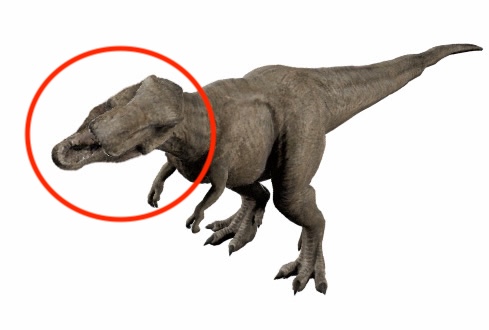} \\
\synlabel{Phase~1} &
\includegraphics[width=\synimgw]{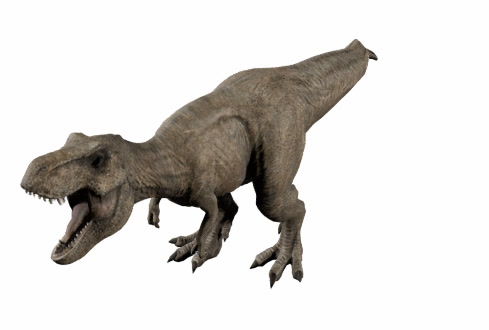} &
\includegraphics[width=\synimgw]{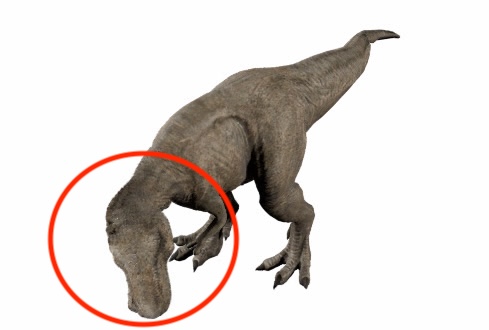} &
\includegraphics[width=\synimgw]{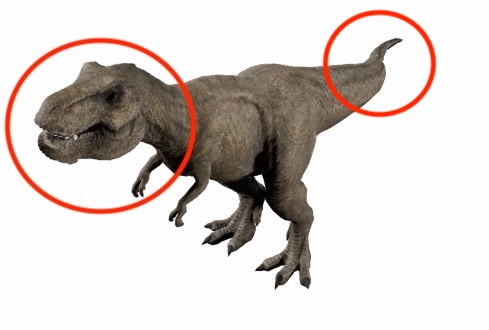} \\
\synlabel{Full} &
\includegraphics[width=\synimgw]{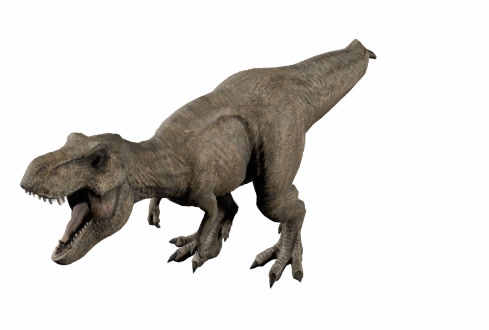} &
\includegraphics[width=\synimgw]{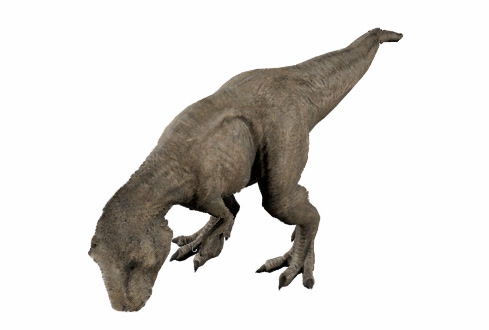} &
\includegraphics[width=\synimgw]{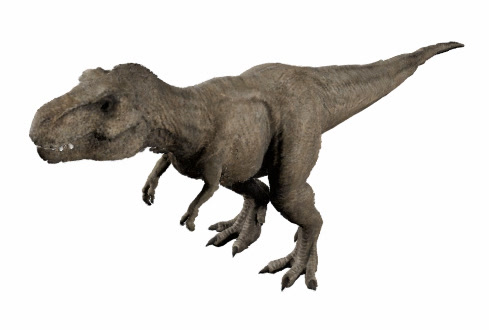} \\
\synlabel{GT} &
\includegraphics[width=\synimgw]{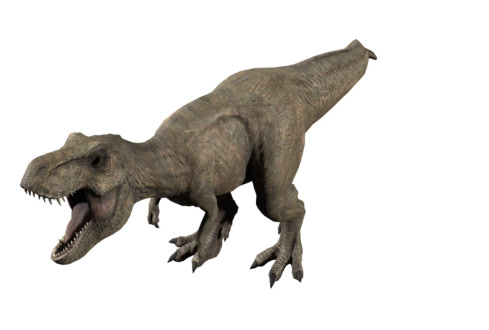} &
\includegraphics[width=\synimgw]{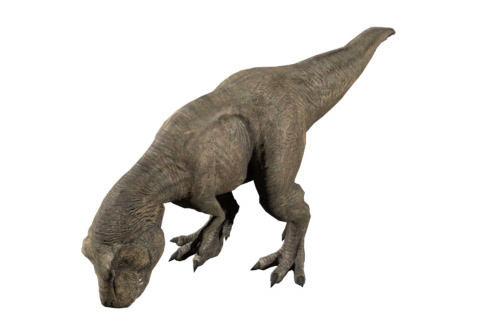} &
\includegraphics[width=\synimgw]{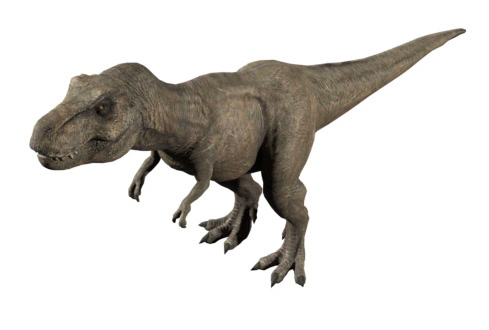} \\
\end{tabular}

\caption{\textbf{Two-phase optimization ablation on T-Rex.} Training View is on top and Novel View is on the bottom. Rows are configurations (Phase~1 w/o $\mathcal{L}_{\Delta\text{depth}}$, Phase~1, Full, GT); columns are matched frames with $t_0$ the canonical (binding) frame and $t_1, t_2$ progressively articulated.}
\label{fig:supp-phase-rexy}
\end{figure}

\paragraph{Our method on T-Rex.} We examine our method's two phases more closely on T-Rex, together with the relative-depth term of Phase~1. Figure~\ref{fig:supp-phase-rexy} shows three configurations side by side: Phase~1 with the relative-depth term $\mathcal{L}_{\Delta\text{depth}}$ disabled, the depth-supervised Phase~1, and Full (Phase~1 followed by Phase~2). Without $\mathcal{L}_{\Delta\text{depth}}$, the per-frame joint rotations are supervised only by the rendering loss at a single fixed view, giving the optimizer the freedom to twist rotations off their physically correct paths to improve the 2D pixel match. The no-depth variant therefore fits the training-view silhouette more aggressively, but misaligns the underlying 3D articulation across the sequence; the artifact is most visible at the head, whose rotation locks into a clearly wrong configuration that subsequent training cannot correct. The depth-supervised Phase~1, by contrast, keeps each rotation on a 3D-plausible trajectory, and with the auto-rigged skinning weights still unmodified at this stage (Phase~2 is what refines them), this fidelity to 3D motion comes at the cost of a slightly looser silhouette fit.

Building on the depth-supervised Phase~1, Phase~2 refines both the skinning weights and the per-joint rotation corrections, addressing both ends of the trade at once: with corrected skinning and refined rotations, the deformation matches the ground-truth silhouette without forcing rotations off their 3D-plausible paths. The visible corrections on T-Rex are highlighted in the figure: the head rotation at $t_1$ and the mouth articulation and tail position at $t_2$  are all brought closer to the ground truth. The metrics tell the same story: training-view PSNR climbs to its best across all configurations ($31.08$~dB), and the chamfer distance against the ground-truth mesh improves further from $5.68{\times}10^{-3}$ at Phase~1 to $4.60{\times}10^{-3}$ at Full.

\subsection{Ground-Truth Deformation} \label{sec:gt-mesh-ablation}

Section~\ref{sec:intro} attributes the joint-boundary artifacts of Gaussian-splat rigging to the primitive rather than to the deformation. This ablation tests that claim by fixing the deformation to the ground truth and varying only the representation. Every method is deformed with the ground truth of the synthetic subject, so no rig error enters the comparison, and no method is fine-tuned per frame. Mani-GS~\citep{manigs} binds its splats to the ground-truth mesh, so they follow the ground-truth mesh sequence. Mani-GS (isotropic) binds its splats to the same mesh and constrains every splat to be isotropic. RigGS~\citep{yao2025riggs} skins its splats directly with linear blend skinning, with no mesh binding, and rotates each splat's covariance by its blended bone rotation. Because the deformation is given by the ground-truth geometry, RigGS's pose-dependent correction head is not needed and is omitted. On fox, RigGS and ours are skinned (Equation~\eqref{eq:lbs}) with the ground-truth rig; on T-Rex and Simpsons, which have no ground-truth skeleton, their bones are the faces of the ground-truth mesh sequence. Every splat-based method reorients each splat's covariance with the local deformation and uses more primitives than ours, from about two to fourteen times as many.

\ifdefined\gtidbasewidth\else\newlength{\gtidbasewidth}\fi
\begin{figure}[!htbp]
\centering
\setlength{\gtidbasewidth}{\linewidth}
\def\gtidleftpanelw{0.65\gtidbasewidth}
\def\gtidrightpanelw{0.34\gtidbasewidth}
\def\gtidimgwrexy{0.24\gtidbasewidth}
\def\gtidimgwfox{0.24\gtidbasewidth}
\def\gtidimgwsimp{0.125\gtidbasewidth}
\def\gtidimghrexy{0.1490\gtidbasewidth} %
\def\gtidimghfox{0.1378\gtidbasewidth}  %
\def\gtidimghsimp{0.1934\gtidbasewidth} %
\def\gtidlabelw{0.085\gtidbasewidth}
\def\gtidsubjectgap{8pt}
\footnotesize
\setlength{\tabcolsep}{0pt}
\renewcommand{\arraystretch}{0.4}
\newcommand{\gtidlab}[2]{\parbox[b][#1][c]{\gtidlabelw}{\raggedleft\scriptsize #2\hspace{2pt}}}
\newcommand{\gtidiso}{\shortstack[r]{Mani-GS\\(isotropic)}}

\begin{minipage}[t]{\gtidleftpanelw}
\centering
\begin{tabular}{@{}c@{\hspace{1pt}}cc@{}}
\multicolumn{3}{@{}l}{\textbf{T-Rex}} \\
\cmidrule(lr){2-3}
& {\scriptsize Binding} & {\scriptsize Deformed} \\
\gtidlab{\gtidimghrexy}{Mani-GS} &
\includegraphics[width=\gtidimgwrexy]{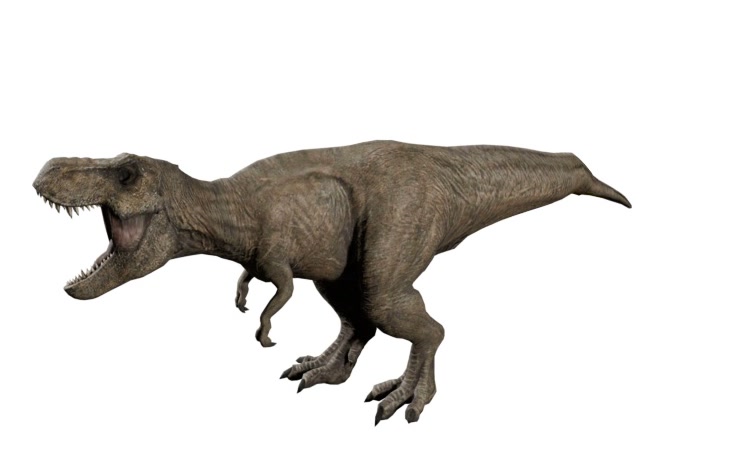} &
\includegraphics[width=\gtidimgwrexy]{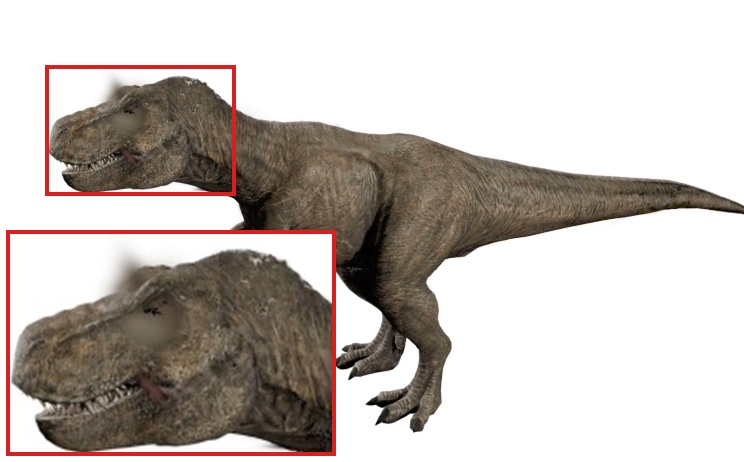} \\
\gtidlab{\gtidimghrexy}{\gtidiso} &
\includegraphics[width=\gtidimgwrexy]{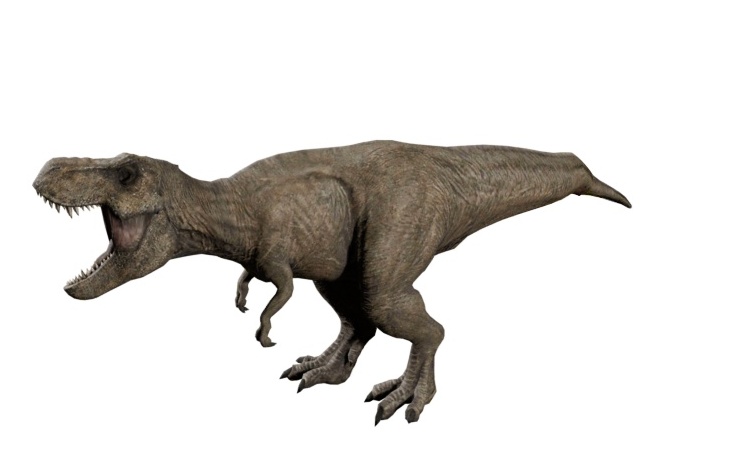} &
\includegraphics[width=\gtidimgwrexy]{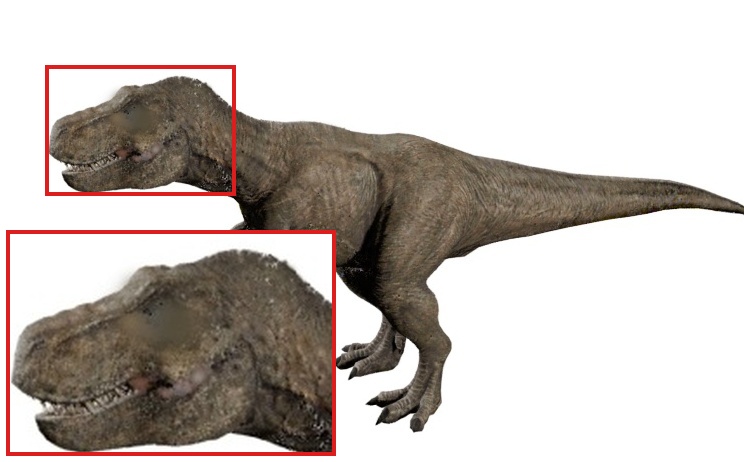} \\
\gtidlab{\gtidimghrexy}{RigGS} &
\includegraphics[width=\gtidimgwrexy]{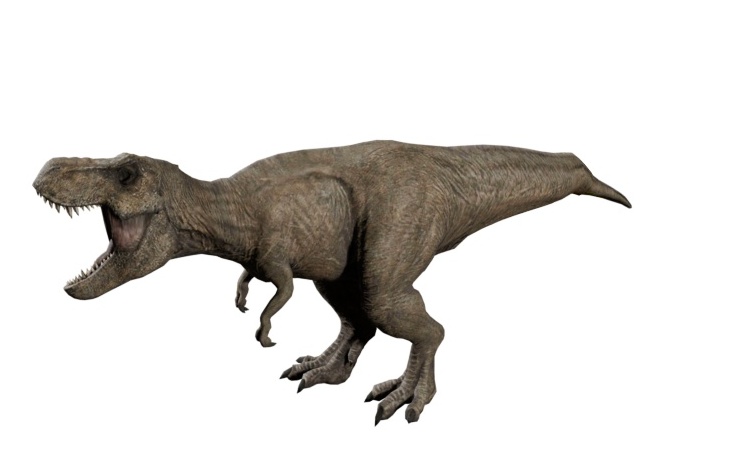} &
\includegraphics[width=\gtidimgwrexy]{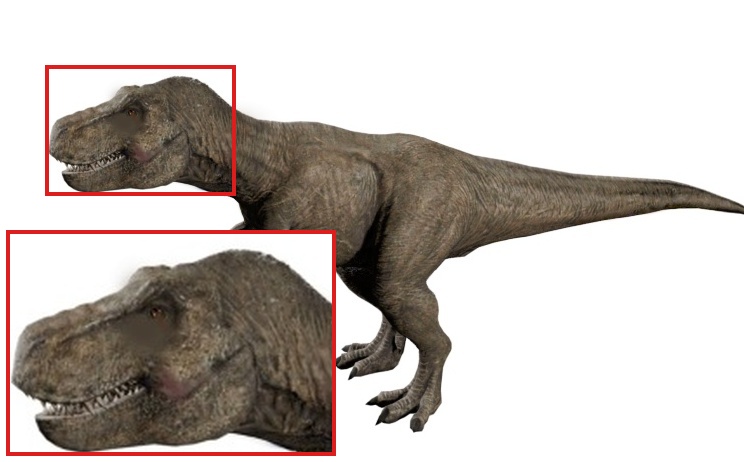} \\
\gtidlab{\gtidimghrexy}{Ours} &
\includegraphics[width=\gtidimgwrexy]{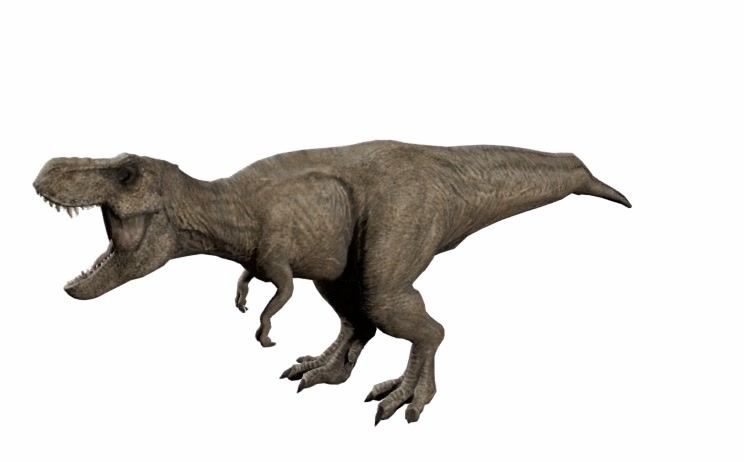} &
\includegraphics[width=\gtidimgwrexy]{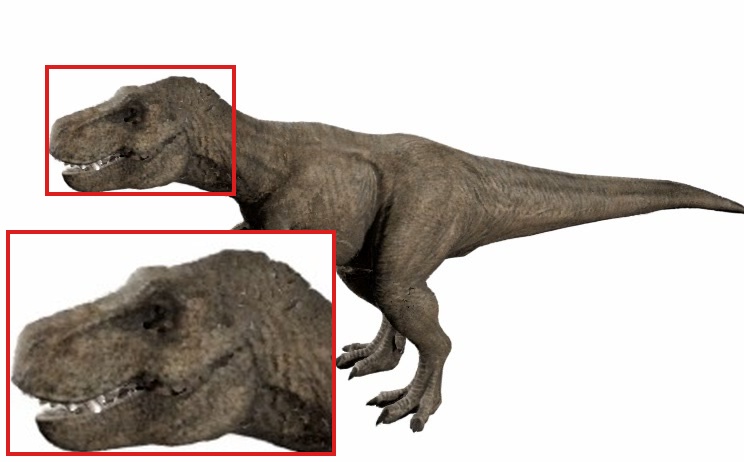} \\
\end{tabular}

\vspace{\gtidsubjectgap}

\begin{tabular}{@{}c@{\hspace{1pt}}cc@{}}
\multicolumn{3}{@{}l}{\textbf{Fox}} \\
\cmidrule(lr){2-3}
& {\scriptsize Binding} & {\scriptsize Deformed} \\
\gtidlab{\gtidimghfox}{Mani-GS} &
\includegraphics[width=\gtidimgwfox]{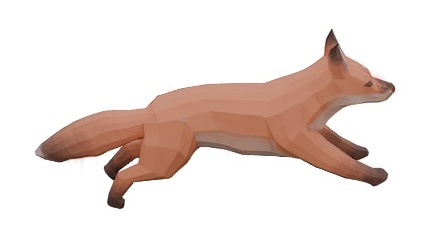} &
\includegraphics[width=\gtidimgwfox]{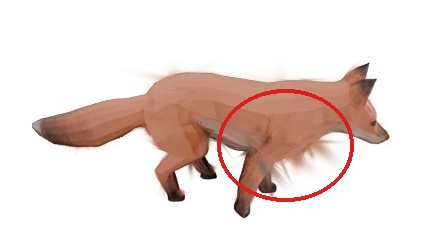} \\
\gtidlab{\gtidimghfox}{\gtidiso} &
\includegraphics[width=\gtidimgwfox]{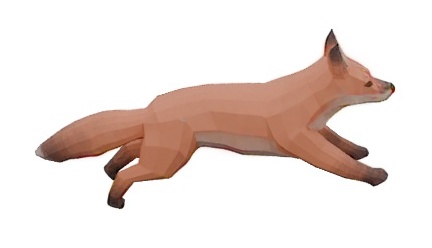} &
\includegraphics[width=\gtidimgwfox]{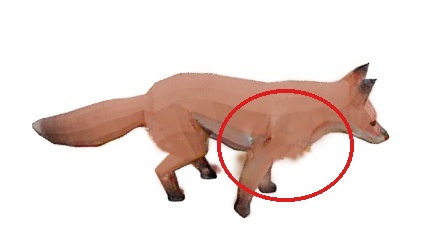} \\
\gtidlab{\gtidimghfox}{RigGS} &
\includegraphics[width=\gtidimgwfox]{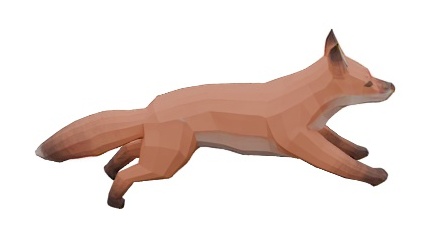} &
\includegraphics[width=\gtidimgwfox]{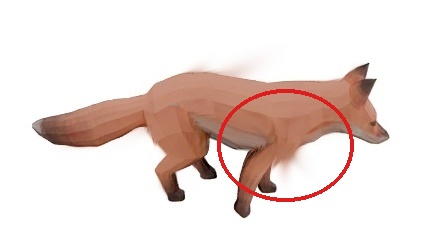} \\
\gtidlab{\gtidimghfox}{Ours} &
\includegraphics[width=\gtidimgwfox]{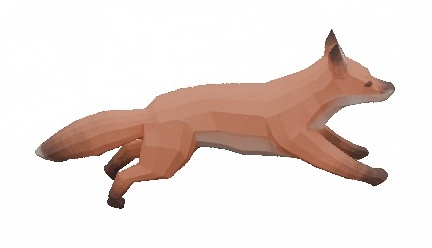} &
\includegraphics[width=\gtidimgwfox]{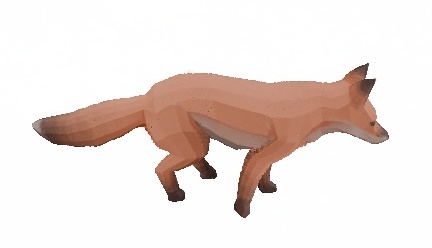} \\
\end{tabular}
\end{minipage}%
\hfill
\begin{minipage}[t]{\gtidrightpanelw}
\centering
\begin{tabular}{@{}c@{\hspace{1pt}}cc@{}}
\multicolumn{3}{@{}l}{\rule{0pt}{40pt}\textbf{Simpsons}} \\
\cmidrule(lr){2-3}
& {\scriptsize Binding} & {\scriptsize Deformed} \\
\gtidlab{\gtidimghsimp}{Mani-GS} &
\includegraphics[width=\gtidimgwsimp]{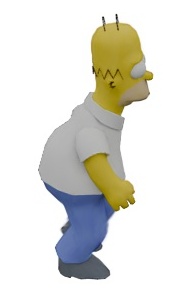} &
\includegraphics[width=\gtidimgwsimp]{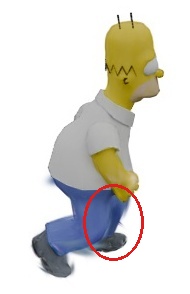} \\
\gtidlab{\gtidimghsimp}{\gtidiso} &
\includegraphics[width=\gtidimgwsimp]{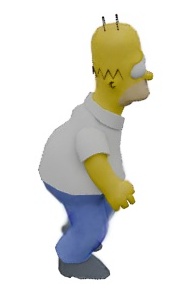} &
\includegraphics[width=\gtidimgwsimp]{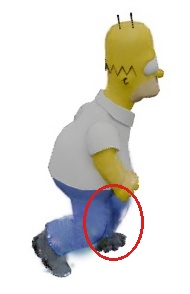} \\
\gtidlab{\gtidimghsimp}{RigGS} &
\includegraphics[width=\gtidimgwsimp]{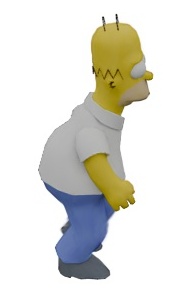} &
\includegraphics[width=\gtidimgwsimp]{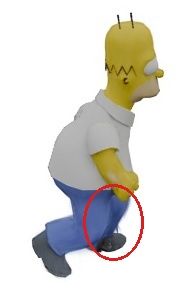} \\
\gtidlab{\gtidimghsimp}{Ours} &
\includegraphics[width=\gtidimgwsimp]{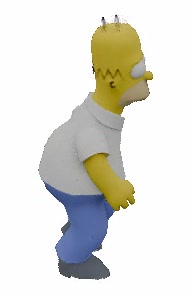} &
\includegraphics[width=\gtidimgwsimp]{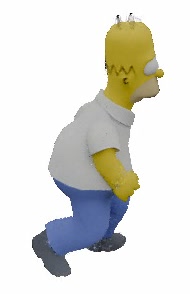} \\
\end{tabular}
\end{minipage}

\caption{\textbf{Splat variants under ground-truth deformation.} Every method is deformed with the ground truth, so no rig error enters the comparison. Rows are methods; columns are the canonical \emph{Binding} frame and one \emph{Deformed} frame. Every splat-based method shows joint-boundary artifacts at the deformed frame; ours does not. Best viewed zoomed in.}
\label{fig:gt-mesh-ablation}
\end{figure}

At the binding frame all methods render comparably (Figure~\ref{fig:gt-mesh-ablation}), so any difference at the deformed frame comes from the deformation. At the deformed frame Mani-GS shows artifacts at the joints, around the fox's front leg--body junction, the Simpsons' front shin and the T-Rex's eye and jaw, although the deformation is exact by construction. Isotropic splats do not remove them: the spikes round into blobs, as in Figure~\ref{fig:why-interpolation}. RigGS, which drops the mesh, shows them at the same joints. Ours, driven by the same ground-truth deformation, shows no such artifact. With the deformation exact, the artifact survives a change of binding and of covariance parameterisation and disappears only when the primitive changes.

\section{Sensitivity Analysis} \label{sec:supp-sensitivity}

We test how sensitive the two-phase optimization is to its loss weights by scaling one weight group at a time by $\times\frac{1}{3}$ and $\times 3$ around the defaults of Table~\ref{tab:loss-weights}, with all other settings fixed.
On T-Rex we vary all six groups; on fox we vary the four groups introduced in Phase~2.
Table~\ref{tab:sensitivity} reports novel-view PSNR.
The relative-depth weight is the most sensitive group: reducing it lowers PSNR, while increasing it raises PSNR.
The tracking weight has a moderate effect, and the remaining groups change PSNR by at most $0.19$~dB.

\begin{table}[h]
\centering
\caption{Loss-weight sensitivity. Novel-view PSNR (dB) when one weight group is scaled by $\times\frac{1}{3}$ or $\times 3$, with all other weights at their defaults (Table~\ref{tab:loss-weights}). The two regularizers of each paired group are scaled together. \emph{Spread} is the range over the three settings. Fox varies only the Phase~2 groups.}
\label{tab:sensitivity}
\footnotesize
\begin{tabular}{l|ccc|ccc}
\toprule
 & \multicolumn{3}{c|}{T-Rex (default 23.95)} & \multicolumn{3}{c}{Fox (default 24.13)} \\
Weight group & $\times\frac{1}{3}$ & $\times 3$ & Spread & $\times\frac{1}{3}$ & $\times 3$ & Spread \\
\midrule
$\mathcal{L}_{\Delta\text{depth}}$ (relative depth)            & 23.10 & 24.65 & 1.55 & --    & --    & --   \\
$\mathcal{L}_{\text{rot}}$ (rotation reg.)                     & 24.07 & 24.14 & 0.19 & --    & --    & --   \\
$\mathcal{L}_{\text{ARAP}}$                                    & 24.05 & 23.92 & 0.13 & 24.22 & 24.27 & 0.14 \\
$\mathcal{L}_{\text{track}}$ (2D tracking)                     & 24.23 & 23.69 & 0.54 & 23.80 & 24.07 & 0.33 \\
Skin-weight reg.\ (smoothness $+$ sparsity)                    & 23.98 & 24.10 & 0.15 & 24.27 & 24.14 & 0.14 \\
Correction reg.\ (magnitude $+$ temporal smoothness)           & 24.05 & 24.08 & 0.13 & 24.27 & 24.24 & 0.14 \\
\bottomrule
\end{tabular}
\end{table}

\section{Additional Implementation Details} \label{sec:supp-impl}

\subsection{Dataset Sequence Lengths} \label{sec:supp-data-lengths}

Table~\ref{tab:seq-lengths} lists the per-scene driving-video lengths referenced in Section~\ref{sec:setup}.

\begin{table}[h]
\centering
\caption{Per-scene driving-video length, in frames. Synthetic scenes are rendered at $512{\times}512$ and ZJU-MoCap frames are used at $512{\times}512$; the captured scenes and their generated driving videos are at $960{\times}540$.}
\label{tab:seq-lengths}
\small
\begin{tabular}{lc}
\toprule
Scene & Frames \\
\midrule
\multicolumn{2}{l}{\emph{Synthetic}} \\
T-Rex        & 40 \\
Simpsons    & 34 \\
Fox         & 24 \\
Wolf        & 18 \\
Spider      & 16 \\
\midrule
\multicolumn{2}{l}{\emph{ZJU-MoCap}} \\
313         & 25 \\
390         & 24 \\
\midrule
\multicolumn{2}{l}{\emph{Captured scenes}} \\
Robot       & 36 \\
Statue      & 18 \\
\bottomrule
\end{tabular}
\end{table}

\subsection{PAPR Backbone, Auto-Rig, and Conventions} \label{sec:supp-backbone}

\paragraph{PAPR backbone.} We use a U-Net-free variant of PAPR~\citep{zhang2023papr}: the rendering path is purely the proximity-attention aggregation over the top-$K$ nearest points, with no image-space decoder. This keeps the rendering pipeline transparent and ensures that no learned image-space network can compensate for deformation artifacts at render time. Backbone hyperparameters (number of points, attention layer widths, top-$K$, training schedule) follow the official PAPR release~\citep{zhang2023papr}; the renderer is frozen for all downstream optimization in this paper.

\paragraph{Auto-rig pipeline.} The auto-rig pipeline runs in two stages on the trained PAPR cloud, and the resulting skeleton and weights are used only as initialization for the two-phase optimization of Section~\ref{sec:multi-phase}. (i)~We extract a triangle-mesh proxy from the PAPR point cloud using the pre-trained, off-the-shelf meshing model OffsetOPT~\citep{offsetopt}, run with its public release defaults; the proxy serves only the next step. (ii)~We feed this proxy to Puppeteer~\citep{song2025puppeteer} with default settings, obtaining a skeleton of $B$ bones and per-vertex skinning weights on the proxy mesh. We then transfer the per-vertex skinning weights to each PAPR point by $k$-nearest-neighbour inverse-distance weighting (IDW) in the canonical pose, with $k=6$ by default, yielding $\mathbf{w}_i\in\mathbb{R}^B$. The proxy mesh is discarded after this transfer and participates in neither deformation nor rendering.

\paragraph{Driving camera.} The driving camera's intrinsics and extrinsics are known and held fixed throughout optimization. For synthetic scenes it is the rendering camera, and for ZJU-MoCap it is the provided calibration of the driving camera. For the captured scenes it is the calibrated camera of the canonical view used to generate the driving video, and we assume the generated video preserves this fixed viewpoint.

\paragraph{Rotation parametrization.} All bone rotations $\mathbf{R}_b^t$ in the optimization, both the Phase~1 frame-wise rotations and the Phase~2 per-joint corrections, are parametrized as axis-angle vectors $\boldsymbol{\omega}_b^t\in\mathbb{R}^3$ (direction giving the rotation axis, magnitude the rotation angle) and converted to $SO(3)$ on the fly via the Rodrigues exponential map $\mathbf{R}_b^t = \exp([\boldsymbol{\omega}_b^t]_\times)$, where $[\,\cdot\,]_\times$ denotes the skew-symmetric matrix of a 3-vector. No per-bone translation is optimized: only the root carries the global translation $\mathbf{t}^t\in\mathbb{R}^3$, and non-root bone offsets are fixed at bind time. Forward kinematics composes these rest-pose offsets with the bone rotations along the skeleton tree to yield the per-bone $SE(3)$ transforms $\mathbf{T}_b(\mathbf{R}^t)$ used in Equation~\eqref{eq:lbs}.

\paragraph{Skinning-weight parametrization.} The skinning weights $\mathbf{w}_i\in\mathbb{R}^B$ used in Equation~\eqref{eq:lbs} are not optimized directly; instead we store unconstrained per-point logits $\mathbf{l}_i\in\mathbb{R}^B$ and recover the active weights at every forward pass by a per-point softmax over bones, $\mathbf{w}_i = \mathrm{softmax}(\mathbf{l}_i)$. This guarantees $w_{i,b}\!\ge\!0$ and $\sum_{b=1}^{B} w_{i,b} = 1$ at every step without explicit projection. The logits are initialized from the auto-rig output as $\mathbf{l}_i = \log \tilde{\mathbf{w}}_i$, where $\tilde{\mathbf{w}}_i$ is the IDW-transferred Puppeteer weight clamped to a small floor for numerical stability so the logarithm is finite. Phase~1 keeps $\mathbf{l}_i$ frozen at this initialization; Phase~2 optimizes $\mathbf{l}_i$ jointly with the correction MLP, with the regularizers tabulated in Table~\ref{tab:loss-weights} acting on the resulting $\mathbf{w}_i$.

\subsection{Baseline Implementation} \label{sec:supp-baselines}

\paragraph{Mani-GS rig pipeline.} The Mani-GS baseline binds splats to a triangle-mesh proxy extracted from its trained first-stage 3DGS, following Mani-GS's own meshing procedure~\citep{manigs}. We auto-rig this proxy with Puppeteer~\citep{song2025puppeteer} (the same auto-rigger used in our method) to obtain a skeleton and per-vertex skinning weights, and the resulting skinning weights are refined as part of the shared Phase~2 schedule.

\paragraph{Mani-GS modifications.} The Mani-GS baseline runs through the same two-stage process as our method, with three additional modifications. First, we disable the per-triangle adaptive scaling throughout training; left on, it drives covariance blow-up that Phase~2 binding-rotation refinement only partially suppresses, leaving visibly noisier renderings (see Section~\ref{sec:supp-mani-scaling}). Second, in Phase~2 we additionally fine-tune each splat's binding rotation and offset against the driving-view loss, freeing the bindings from being fit only to the canonical frame. Third, we add a foreground-mask loss to its rendering objective for added silhouette supervision. Together these modifications yield the strongest Mani-GS baseline we could obtain for this fixed-camera setting.

\paragraph{RigGS.} RigGS~\citep{yao2025riggs} originally jointly reconstructs the canonical object and its rig from a monocular video in which the camera moves relative to the subject. Our setting instead begins with an existing canonical reconstruction and estimates its motion from a single fixed-viewpoint video. RigGS's paper notes that the method ``faces challenges with videos captured by a fixed camera'' and that improved results ``can be achieved when the camera is allowed to move.'' On ZJU-MoCap, it accordingly cycles through six cameras to simulate a monocular video with camera movement. Its complete reconstruction-and-rigging pipeline is therefore not directly applicable under our input assumptions.

We therefore adapt RigGS to our pipeline, following its released implementation. RigGS reconstructs the canonical splats and extracts a skeleton with skinning weights derived from point-to-bone distances, both from the driving video itself. In our setting these are available as inputs, so we provide them rather than have them inferred: the canonical splats and mesh reconstructed from the first frame, and the same canonical skeleton and skinning field the other methods receive, carried to each splat through its binding face. The deformation model is then RigGS's own. Each splat is skinned directly by linear blend skinning, taking its position from the blended bone transforms and its orientation from the blended joint quaternions, with scale left unchanged, and RigGS's pose-conditioned detail-deformation head adds a per-splat world-space translation after skinning. We use the released head at its published width, depth, and positional encoding, and its offset regularizer at the published weight including the canonical-frame term, and the skinning weights remain trainable as in RigGS. The canonical splats are not reinitialized, and the training schedule and evaluation match the other methods.

\subsection{Mani-GS Adaptive Scaling Ablation} \label{sec:supp-mani-scaling}

In Section~\ref{sec:supp-baselines} we noted that the Mani-GS baseline has its per-triangle adaptive scaling disabled throughout training. Mani-GS rescales each splat's scaling vector at every frame in proportion to its bound triangle's deformed edge lengths, so the splat tracks the triangle's shape change~\citep{manigs}. On thin or near-degenerate triangles, this rescaling drives the splat covariance to extreme values and produces visible artifacts under articulation. Phase~2 binding-rotation refinement reduces some of these artifacts by reorienting affected splats, but cannot address them as effectively as disabling adaptive scaling outright.

\begin{figure}[!htbp]
\centering
\footnotesize
\setlength{\tabcolsep}{0pt}
\renewcommand{\arraystretch}{0.4}

\def\mslabelw{0.10\linewidth}
\newcommand{\mslab}[1]{\parbox[b][\msimgh][c]{\mslabelw}{\centering\scriptsize #1}}

\def\msimgw{0.17\linewidth}
\def\msimgh{0.34\linewidth}
\begin{tabular}{@{}c@{\hspace{2pt}}ccc@{}}
\multicolumn{4}{@{}l}{\textbf{Simpsons}} \\
& {\scriptsize $t_0$ (canonical)} & {\scriptsize $t_1$} & {\scriptsize $t_2$} \\
\mslab{Mani-GS \\ (scaling on)} &
\includegraphics[width=\msimgw]{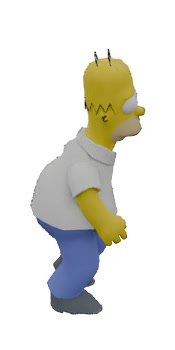} &
\includegraphics[width=\msimgw]{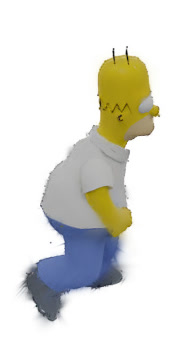} &
\includegraphics[width=\msimgw]{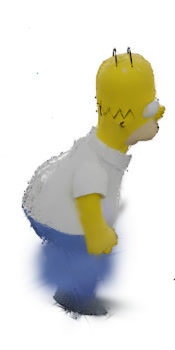} \\
\mslab{Mani-GS \\ (scaling off)} &
\includegraphics[width=\msimgw]{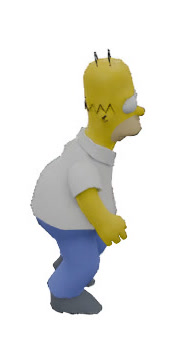} &
\includegraphics[width=\msimgw]{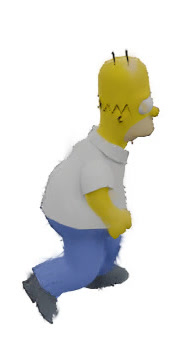} &
\includegraphics[width=\msimgw]{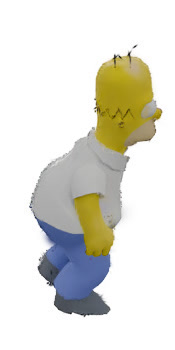} \\
\end{tabular}

\vspace{8pt}

\def\msimgw{0.29\linewidth}
\def\msimgh{0.194\linewidth}
\begin{tabular}{@{}c@{\hspace{2pt}}ccc@{}}
\multicolumn{4}{@{}l}{\textbf{Fox}} \\
& {\scriptsize $t_0$ (canonical)} & {\scriptsize $t_1$} & {\scriptsize $t_2$} \\
\mslab{Mani-GS \\ (scaling on)} &
\includegraphics[width=\msimgw]{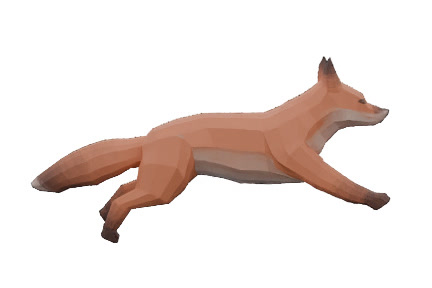} &
\includegraphics[width=\msimgw]{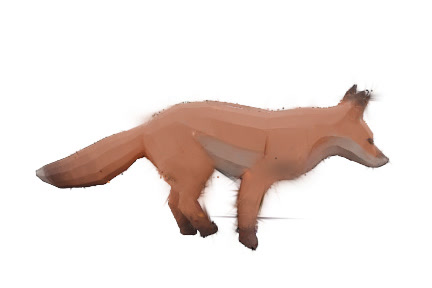} &
\includegraphics[width=\msimgw]{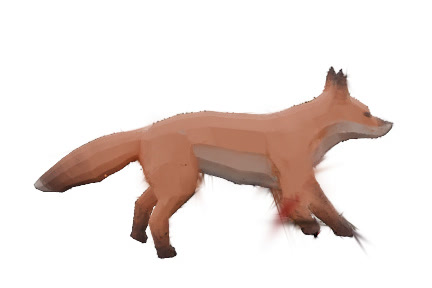} \\
\mslab{Mani-GS \\ (scaling off)} &
\includegraphics[width=\msimgw]{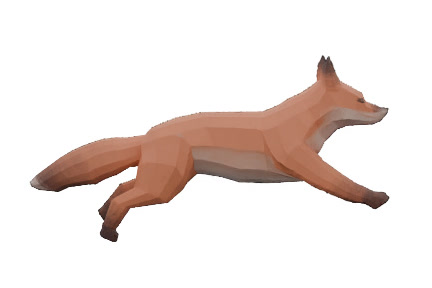} &
\includegraphics[width=\msimgw]{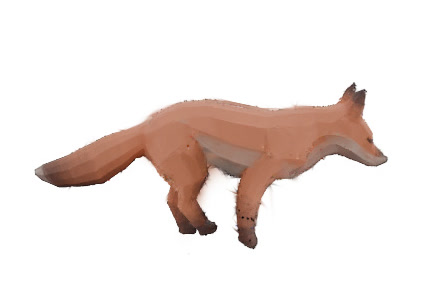} &
\includegraphics[width=\msimgw]{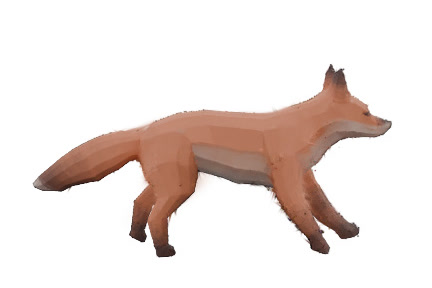} \\
\end{tabular}

\caption{\textbf{Mani-GS adaptive-scaling ablation.} Simpsons (top) and fox (bottom), each shown as two rows comparing the default Mani-GS configuration (per-triangle adaptive scaling \emph{on}) with the configuration used in the main paper (scaling \emph{off}). Both rows share our full two-phase optimization; only the scaling toggle differs. Columns are matched training-view frames, with $t_0$ the canonical (binding) frame and $t_1, t_2$ progressively articulated. Artifacts produced by the most degenerate splats remain visible with scaling enabled even after Phase~2.}
\label{fig:supp-mani-scaling}
\end{figure}

Figure~\ref{fig:supp-mani-scaling} compares the two configurations on Simpsons and fox. Both runs use our full two-phase optimization; only the adaptive-scaling toggle differs. With scaling enabled, articulated frames retain visible artifacts even after Phase~2; we therefore disable it to obtain the stronger Mani-GS baseline used throughout Section~\ref{sec:experiment}.

\subsection{Optimization Details} \label{sec:supp-optim}

This section gathers the loss weights, regularizers, and schedules used in the two-phase optimization of Section~\ref{sec:multi-phase}. Both phases use the rendering term $\mathcal{L}_{\text{rgb}} = 0.8 \cdot \mathrm{L1} + 0.2 \cdot (1-\mathrm{SSIM})$ on the rendered patch, augmented with an LPIPS term at weight $0.01$. The remaining weights are summarized in Table~\ref{tab:loss-weights}.

\begin{table}[h]
\centering
\caption{Loss weights for the two-phase optimization of Section~\ref{sec:multi-phase}. Entries marked ``decays $a \to 0$'' follow a schedule that decays from the listed value to zero over the course of Phase~2. Dashes indicate the term is not active in that phase. These are the default values; per-scene deviations are listed below the table.}
\label{tab:loss-weights}
\small
\begin{tabular}{l|cc}
\toprule
Term & Phase 1 & Phase 2 \\
\midrule
$\mathcal{L}_{\text{rgb}}$ ($0.8\cdot\mathrm{L1} + 0.2\cdot(1-\mathrm{SSIM})$) & $1.0$ & $1.0$ \\
LPIPS                                                                          & $0.01$ & $0.01$ \\
$\mathcal{L}_{\Delta\text{depth}}$ (relative depth)                            & $1.0$ & decays $1.0 \to 0$ \\
$\mathcal{L}_{\text{rot}}$ (descendant-weighted rotation reg.)                 & $0.1$ & --- \\
$\mathcal{L}_{\text{track}}$ (2D tracking)                                     & --- & decays $0.005 \to 0$ \\
$\mathcal{L}_{\text{ARAP}}$ ($k$-NN distance preservation, $k{=}8$)            & --- & $500$ \\
Composed-rotation magnitude reg.\ (Phase~1\,$+$\,correction)                   & --- & $0.01$ \\
Correction L2 magnitude                                                        & --- & $5\!\times\!10^{-4}$ \\
Correction temporal smoothness                                                 & --- & $0.1$ \\
Skin-weight smoothness ($k$-NN graph)                                          & --- & $0.03$ \\
Skin-weight sparsity                                                           & --- & $0.01$ \\
\bottomrule
\end{tabular}
\end{table}

On wolf, Phase~2 keeps the composed-rotation regularizer at its Phase~1 weight ($0.1$ instead of $0.01$). On ZJU-MoCap subject 390, the $\Delta$-depth weight is $2$ in Phase~1 and $8$ in Phase~2.

\paragraph{Optimizer and learning-rate schedule.} Both phases use Adam with cosine-annealing learning-rate schedules. In Phase~1, each frame is fit with a single learning rate over its joint rotations and root translation, annealed from $5\!\times\!10^{-3}$ to $1\!\times\!10^{-4}$. In Phase~2, the optimizer uses per-parameter-group learning rates: $3\!\times\!10^{-4}$ for the correction MLP and $1\!\times\!10^{-2}$ for the skinning-weight logits. By default, both groups are cosine-annealed to $1\!\times\!10^{-5}$ over $30{,}000$ Phase~2 iterations. On ZJU-MoCap, Phase~2 runs $20{,}000$ iterations with a skinning-weight learning rate of $1\!\times\!10^{-4}$; on wolf, the correction MLP uses a learning rate of $1\!\times\!10^{-4}$.

\paragraph{$\Delta$-depth supervision.} At each optimization step the relative-depth loss $\mathcal{L}_{\Delta\text{depth}}$ is computed on a random subsample of $4096$ canonical points. The same points are projected through the binding-frame and current-frame cameras to obtain camera-space depths $z^0_i, z^t_i$ and pixel coordinates $p^0_i, p^t_i$; binding-frame quantities are detached so that gradients flow only through the deformation. Monocular depths $\tilde m^0_i, \tilde m^t_i$ are bilinearly sampled from the Depth Anything~3 maps at $p^0_i$ and $p^t_i$ (we use the metric-depth variant of Depth Anything~3, so $\tilde m$ is in metric units rather than affine-invariant disparity, and the linear scale alignment below is well-posed), and we form the per-point deltas $\Delta z_i = z^t_i - z^0_i$ and $\Delta m_i = \tilde m^t_i - \tilde m^0_i$. We retain only points that lie in front of the camera and project inside the image at \emph{both} frames, and whose current-frame projection lands on a foreground-mask pixel; the foreground requirement restricts the depth signal to points that currently contribute to the rendered silhouette. The per-frame scale is computed in closed form on the retained samples, $s_t = \langle \Delta z, \Delta m \rangle / \|\Delta m\|^2$, under \texttt{no\_grad}, so it acts as a constant during backpropagation; it is recomputed fresh each step on the new subsample and never persists as state. The loss is then $\mathcal{L}_{\Delta\text{depth}} = \tfrac{1}{n}\sum_i (\Delta z_i - s_t \Delta m_i)^2$, where $n$ is the number of retained samples. Phase~2 uses the same construction with $\lambda_{\Delta\text{depth}}$ decayed to zero according to Table~\ref{tab:loss-weights}.

\paragraph{Correction network.} The Phase~2 correction $f_\theta(t)$ is a small temporal MLP that takes a positional encoding of the normalized frame index $t/T$ as input and outputs a per-joint axis-angle offset, added to the Phase~1 rotation of that joint before LBS. The positional encoding uses sine and cosine bases with frequency exponents $0$ through $10$ (eleven bands), giving an input of dimension $1+2\cdot 11=23$. The MLP has two hidden layers of width $128$ with ReLU activations and a linear output head producing $3B$ values reshaped to per-joint axis-angle vectors; the output layer is zero-initialized so $f_\theta$ produces zero correction at the start of training. The Phase~1 rotations and the PAPR network remain frozen throughout Phase~2.

\paragraph{Correction regularizers.} The correction outputs are regularized with an L2 penalty on their magnitude (weight $5\!\times\!10^{-4}$) and a temporal-smoothness penalty on consecutive-frame differences (weight $0.1$); together these keep the corrections small and slowly varying.

\paragraph{Skin-weight regularizers (Phase~2).} The Phase~2 skin-weight refinement is regularized by two terms acting on the active weights $\mathbf{w}_i = \mathrm{softmax}(\mathbf{l}_i)$. The smoothness term penalizes disagreement among $k$-nearest canonical neighbours,
\begin{equation}
\mathcal{L}_{\text{smooth}} \;=\; \frac{1}{N k}\sum_{i=1}^{N}\sum_{j\in\mathcal{N}(i)} \big\|\mathbf{w}_i-\mathbf{w}_j\big\|_1,
\label{eq:smooth}
\end{equation}
on the same canonical $k$-NN graph used by $\mathcal{L}_{\text{ARAP}}$ ($k=8$). The sparsity term is the per-coordinate negative binary entropy,
\begin{equation}
\mathcal{L}_{\text{sparse}} \;=\; -\frac{1}{NB}\sum_{i=1}^{N}\sum_{b=1}^{B} \Big[w_{i,b}\log w_{i,b} + (1-w_{i,b})\log(1-w_{i,b})\Big],
\label{eq:sparse}
\end{equation}
which pushes each $w_{i,b}$ toward $0$ or $1$ and so toward fewer effective bones per point.

\paragraph{Descendant-weighted rotation regularizer.} Joint rotations are penalized in proportion to how many child joints they propagate to, so that bones high in the kinematic tree (which carry their entire subtree with them) are encouraged to be quieter than leaf bones. Let $n_{\text{desc}}(b)$ denote the number of descendants of bone $b$ in the Puppeteer-returned kinematic tree (zero for leaves), and define
\begin{equation}
\alpha_b \;=\; \frac{1+n_{\text{desc}}(b)}{\,\overline{1+n_{\text{desc}}(\cdot)}\,},
\label{eq:desc-weight}
\end{equation}
normalized so $\overline{\alpha}=1$ across the $B$ bones. Phase~1 regularizes the per-frame axis-angle magnitudes,
\begin{equation}
\mathcal{L}_{\text{rot}}^{(1)} \;=\; \frac{1}{B}\sum_{b=1}^{B}\alpha_b\, \big\|\boldsymbol{\omega}_b^t\big\|_2^2.
\label{eq:rot-phase1}
\end{equation}
Phase~2 regularizes the \emph{composed} rotation (Phase~1 base $\mathbf{R}_b^t$ rotated by the per-frame correction $\mathbf{R}_b^{(c,t)}$ produced by $f_\theta$) via a Frobenius-norm proxy,
\begin{equation}
\mathcal{L}_{\text{rot}}^{(2)} \;=\; \frac{1}{T B}\sum_{t=1}^{T}\sum_{b=1}^{B}\alpha_b\, \big\|\mathbf{R}_b^t \mathbf{R}_b^{(c,t)} - \mathbf{I}\big\|_F^2,
\label{eq:rot-phase2}
\end{equation}
which is monotonically related to rotation magnitude and avoids a matrix logarithm.

\paragraph{ARAP distance-preservation term.} The as-rigid-as-possible term $\mathcal{L}_{\text{ARAP}}$ in Phase~2 acts on a fixed canonical $k$-nearest-neighbour graph over the PAPR points, with $k=8$ neighbours per point computed once at the binding pose. For each canonical edge $(i,j)$ we penalize the squared change in pairwise distance under the deformed configuration, $\sum_{(i,j)} \big(\|\mathbf{p}_i^t - \mathbf{p}_j^t\| - \|\mathbf{p}_i^0 - \mathbf{p}_j^0\|\big)^2$, weighted by the loss coefficient in Table~\ref{tab:loss-weights}. The graph is canonical (not recomputed per frame), so the term anchors local geometry to the binding-pose neighbourhood structure throughout the sequence.

\paragraph{Sequential frame fitting.} In Phase~1 we fit the per-frame rotations sequentially from the binding frame: each frame $t{+}1$ is initialized from the optimized rotations of frame $t$, exploiting temporal smoothness in the driving video so that no frame is optimized from scratch. Each frame is optimized for $2{,}000$ iterations.

\paragraph{Hardware and runtime.} All optimization runs on a single NVIDIA RTX~3090 GPU. For a $16$-frame sequence, Phase~1 takes approximately one hour, and Phase~2 takes another hour for its $30{,}000$ iterations.

\subsection{Attention-Guided Track Seeding} \label{sec:supp-tracks}

The 2D tracking loss $\mathcal{L}_{\text{track}}$ in Phase~2 requires a set of (point, query-pixel) pairs at the canonical frame: each selected canonical point is propagated forward in time by CoTracker from its query pixel, and the resulting trajectory supervises the projection of the LBS-deformed point. The construction below chooses which points to supervise and where to query them.

\paragraph{Off-ray issue with naive projection.} The simplest construction, projecting all canonical points through the camera and using those pixel positions as queries, is problematic for an attention-based renderer like PAPR. A point can contribute to a pixel without lying exactly on its ray, and conversely a point's geometric projection is not necessarily where it visually contributes. Projection-only queries therefore include points the renderer effectively ignores at the canonical view, points whose projection lands a pixel or two off the visible surface, and silently-occluded back-facing points. Each of these injects label noise into the tracking loss.

\paragraph{Attention-dominant selection.} We instead select queries from the renderer's own attention. We run PAPR over the canonical-frame image (tiled in patches for memory) to obtain, for each foreground pixel, the indices and attention weights of the top-$K$ points selected by proximity attention. For each foreground pixel we take the argmax over its top-$K$ attention weights to identify the dominant point: the single point that most explains that pixel. Pixels whose top-1 attention falls below a threshold (we use $0.05$) are discarded as smeared/low-confidence, and the remaining indices are deduplicated to yield the set of dominant points for the canonical view. Each selected point is then projected through the canonical camera; by construction this projection lies inside the rendered silhouette near a pixel the point dominates, and that 2D position is the CoTracker query for the point. Tracks and the corresponding 3D point indices are persisted together so that the optimizer knows which canonical point each trajectory supervises.

\paragraph{Mani-GS query seeding.} For Mani-GS we seed CoTracker queries by projecting the canonical mesh vertices through the same canonical view, restricting to those inside the foreground mask, and uniformly subsampling on the image plane by grid binning to a comparable query count. The CoTracker model, temporal window, canonical view, and query count are thus matched to ours; the seeding differs only in per-pixel point selection (mesh-vertex grid sampling vs.\ attention dominance).

\end{document}